\documentclass[10pt,twocolumn,letterpaper]{article}
\usepackage[pagenumbers]{cvpr}

\usepackage{cuted} 

\usepackage{currfile} 

\usepackage{caption} 

\usepackage[stretch=40,shrink=40,step=1]{microtype}

\makeatletter\renewcommand\paragraph{\@startsection{paragraph}{4}{\z@}{5pt \@plus 2pt \@minus 1pt}{-1em}{\normalfont\normalsize\bfseries}}\makeatother   

\makeatletter
\def\abstract{\iftoggle{cvprpagenumbers}{}{\thispagestyle{empty}}\centerline{\large\bf Abstract}\vspace*{0pt}\noindent\it\fontsize{10pt}{12.04pt}\selectfont\ignorespaces}

\makeatother
\AtBeginEnvironment{thebibliography}{\raggedright}

\usepackage{mathptmx}
\usepackage{booktabs}
\usepackage{pifont}
\usepackage{multirow}
\definecolor{cvprblue}{rgb}{0.21,0.49,0.74}
\usepackage[pagebackref,breaklinks,colorlinks,allcolors=cvprblue]{hyperref}

\def\confName{CVPR}
\def\confYear{2027}

\title{A Straight Line Is Not a Camera Pose:\\
When Aligned Metrics Let Input-Free Predictions Compete}

\author{Xinyi Shi\\Duke University}

\usepackage{longtable}
\newcommand{\mref}[1]{\ref{#1}}
\begin{document}
\twocolumn[{\renewcommand\twocolumn[1][]{#1}\maketitle\vspace{-2.2em}
\begin{center}
\includegraphics[width=\textwidth]{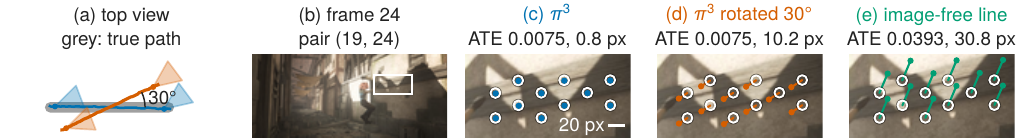}
\captionof{figure}{ATE does not see where the cameras look (Sintel alley\_2). (a)~With the orientations aligned, $\pi^3$ (blue) follows the true path (grey; wedges: viewing directions). Its centres rotated by 30$^\circ$ with the orientations kept (orange) move sideways to where the cameras look, yet ATE's position-only alignment maps them back onto $\pi^3$'s: both score 0.0075. (b)~A frame and the crop of (c--e), where each prediction places static points from 5 frames earlier (ring: true position; dot: predicted; 20\,px bar). Above (c--e): the clip's ATE and mean reprojection error (\cref{sec:consequence}).}\label{fig:teaser}
\end{center}\vspace{0.6em}}]
\begin{abstract}
Many vision benchmarks align a prediction to the ground truth before scoring it. If the alignment minimises the error, an input-free template scores the truth's distance from everything the alignment can reach, so the ground truth alone fixes how well a template can score, before any model runs.
Camera pose is our central case. The absolute trajectory error (ATE), the standard measure of camera motion from SLAM to feed-forward 3D reconstruction, aligns camera positions by a similarity transform. On the common Sintel protocol, an image-free constant-velocity line then has a lower mean ATE than 10 of 11 recent models and 85 of 110 published methods, and on Lightspeed than 5 of 9 models and 10 of 11 published methods. The straightness of the true paths separates these two from TUM, ScanNet, DyCheck and ADT, where every model beats the line. A residual signal-to-noise criterion with the alignment's first-order refit predicts when straightening a prediction lowers ATE in all but one of 4010 model--clip evaluations, including all 504 on three benchmarks registered in advance. The alignment also ignores where cameras look: rotating a trajectory by 30$^\circ$ against its orientations leaves ATE unchanged yet worsens the static-point reprojection of every model on Sintel, TUM and ScanNet, and the line reprojects worst on Sintel by geometric mean. Fitting the alignment's rotation to the predicted orientations gives pose-anchored ATE (PA-ATE), a one-function change at about ATE's cost (37 vs.\ 34\,\textmu s per clip, and 61 vs.\ 66\,\textmu s with the rotation error). It is never below ATE, equals the all-pairs translational relative pose error divided by $\sqrt2$ when orientations agree, ranks every image-free reference below every model on Sintel, tracks the reprojection error under controlled rotation and straightening, largely preserves the models' order, and exposes Lightspeed ground-truth orientations that the images contradict.
Beyond poses, templates beat some published methods on KITTI video depth (a mean scene layout), SSP-3D body shape (the mean shape) and NoW faces (the mean face), and in video depth this distance, computed from the ground truth alone, sets KITTI apart from Sintel and Bonn. PA-ATE's anchor follows the rule behind fixes in speech, face and video-depth evaluation: align only what the task may ignore.
\end{abstract}

\section{Introduction}
\label{sec:intro}

Many vision benchmarks align a prediction to the ground truth before scoring it, by a similarity transform for camera trajectories, a scale and shift for depth, or a similarity or size normalisation for body and face meshes.
The alignment removes what the input cannot determine, such as the world frame or the scale, and nothing it removes is scored.
When it minimises the error, an input-free template, one answer for every input, scores the ground truth's distance from all that the alignment can reach (\cref{sec:beyond}).
This distance is known before any model runs, and where it is small the template competes with trained models (\cref{tab:templates}).

Camera pose is our central case.
Camera poses link images to 3D space, and reconstruction, localisation and augmented reality all depend on their accuracy.
It is most often reported as the absolute trajectory error (ATE)~\cite{tum}: the predicted camera positions are aligned to the ground truth by a rigid or, for monocular methods, a similarity transform, and the remaining position RMSE is reported~\cite{zhang2018tutorial}.
SLAM and visual odometry have used ATE for over a decade, as have the feed-forward networks that map images or a video to camera poses and geometry in one pass~\cite{dust3r,monst3r,cut3r,vggt,pi3,da3,ttt3r,streamvggt,mapanything}.
For dynamic scenes, progress is tracked in a shared ATE table on short clips of Sintel~\cite{sintel}, TUM-dynamics~\cite{tum} and ScanNet~\cite{dai2017scannet}, following MonST3R~\cite{monst3r} (\cref{fig:progress}).

This paper asks what that number establishes.
On the standard 14-clip Sintel protocol, a camera moving along a line at constant speed, without images, reaches a mean ATE of 0.092, better than 10 of 11 models we re-ran, than 85 of 110 published methods we admit, and than MonST3R itself (0.118 in our run; \cref{sec:diagnosis}).
On Lightspeed~\cite{dynpose100k} it beats 5 of 9 models and 10 of 11 published methods, while on TUM, ScanNet, DyCheck~\cite{dycheck} and ADT~\cite{adt} every model beats it.

Two properties of the alignment explain the observation (\cref{sec:cause}).
First, the line is an input-free template. The alignment maps any moving constant-velocity prediction onto the best-fitting line through the true positions, so the line scores the true path's regression residual, and on Sintel this is often smaller than the models' errors.
The line is the end point of \emph{straightening}, which moves a prediction towards its own constant-velocity fit without images.
We derive a residual signal-to-noise criterion that tells, per clip, when straightening lowers ATE. It explains why straightening halfway lowers the mean Sintel ATE of all 11 models but raises that of every model on TUM, ScanNet and full-length Bonn and Wild-SLAM.
Second, the alignment is fitted to positions alone, so rotating the predicted positions against the orientations leaves ATE unchanged, though it changes where the cameras move relative to their view (\cref{fig:teaser}).

Reprojecting static scene points with the true depth tests the camera geometry directly (\cref{sec:consequence}). Neither ATE nor the rotational relative pose error (RPE$_r$) the protocol reports checks the heading, the direction of travel in each camera's axes, against the orientation, and we add it to ATE (\cref{sec:correction}).

\begin{table}[t]
\centering\small
\begin{tabular*}{\linewidth}{@{\extracolsep{\fill}}llcc@{}}
\toprule
Task & Benchmark & $R^2$ & Beats \\
\midrule
Camera poses & Sintel & 0.99 & 10 of 11 \\
& Lightspeed & 0.99 & 5 of 9 \\
& TUM, ScanNet & 0.51--0.67 & none \\
& DyCheck, ADT & 0.14--0.33 & none \\
\midrule
Video depth & KITTI & 0.75 & 1 of 5 \\
& Sintel, Bonn & 0.04--0.09 & none \\
\midrule
Body shape & SSP-3D & -- & 5 of 9 \\
\midrule
Faces & NoW & -- & 8 of 21 \\
\bottomrule
\end{tabular*}
\caption{Input-free templates under each benchmark's own alignment (constant-velocity line, mean layout of the other sequences, SMPL mean shape, FLAME mean face). $R^2$: median share of the true variance inside the template's orbit, from the ground truth alone (--: no least-squares fit). Beats: models we ran (poses) or methods of the source table with a higher error (depth: on both metrics).}
\label{tab:templates}
\end{table}

\smallskip\noindent\textbf{Contributions.}
\begin{itemize}[leftmargin=*,itemsep=0pt,topsep=1pt,parsep=0pt]
\item \emph{Principle (proved; \cref{sec:beyond}).} Under any alignment that minimises the error, an input-free template scores the truth's distance from the alignment's reach. Besides poses, templates beat some published methods on KITTI video depth, SSP-3D body shapes and NoW faces, and this distance, computed from the ground truth, singles out KITTI among three video-depth benchmarks (\cref{tab:templates}).
\item \emph{Audit (empirical; \cref{sec:diagnosis,sec:cause}).} Across ten datasets, the line beats 10 of 11 models and 85 of 110 published methods on Sintel and 5 of 9 models on Lightspeed.
\item \emph{Theory (proved; \cref{sec:line,sec:cause,sec:correction}).} ATE's alignment scores any constant-velocity prediction with nonzero fitted speed at the true path's regression residual and is invariant to rotating positions against orientations. A residual SNR condition decides exactly, at fixed alignment, when straightening lowers ATE. PA-ATE equals the all-pairs translational RPE divided by $\sqrt2$ when the anchored orientations agree, and differs from it within a stated bound otherwise.
\item \emph{Validation (empirical; \cref{sec:cause}).} With the refit's first-order term the condition accounts for 4009 of 4010 model--clip outcomes (95 to 98\% of each pooled set without the term), among them every held-out and ScanNet one and all 504 on three benchmarks registered in advance, and for every pair at three other straightening strengths. The straightness of the true path separates Sintel and Lightspeed from the other four pose benchmarks of \cref{tab:templates}.
\item \emph{Geometry check (empirical; \cref{sec:consequence}).} Rotating positions by 30$^\circ$ leaves ATE unchanged but worsens the static-point reprojection of all 11 models on Sintel, TUM and ScanNet (8 to 13\,px on Sintel), and the line, 4th by geometric-mean ATE on Sintel, places points worst (33 vs.\ 6 to 26\,px).
\item \emph{A practical score (\cref{sec:correction,sec:recs}).} \emph{Pose-anchored ATE} (PA-ATE) fits ATE's rotation to the orientations, a one-function change that re-scores saved predictions at about ATE's cost (37 against 34\,\textmu s per clip, 61 against 66\,\textmu s with the rotation error). It ranks every image-free reference below every model on Sintel, tracks the reprojection error under controlled rotation and straightening (Spearman 0.61 vs.\ 0.32 on Sintel) and largely preserves the models' order.
\item \emph{Findings on the benchmarks (\cref{sec:benchchange}).} PA-ATE shows that Bonn's ground truth gives the markers' orientations, not the camera's, and that 20 of 36 Lightspeed clips store orientations reflected against the models', where image-only rotations side with the models on 15 of the 19 testable clips. ATE detects neither.
\item \emph{Release.} Code, predictions and a number-checking script.
\end{itemize}

\section{Related work}
\label{sec:related}
Each line of work below ends with what we add.
\paragraph{Feed-forward reconstruction and its benchmarks.}
Since DUSt3R and MASt3R~\cite{dust3r,mast3r}, pointmap regression has been extended to dynamic scenes~\cite{monst3r,align3r,easi3r,d4rt}, to streams~\cite{spann3r,cut3r,ttt3r,streamvggt,point3r,stream3r,wint3r,mast3rslam}, and to many views~\cite{fast3r,flare,vggt,pi3,da3,mapanything,vggtomega}, with optimisation-based references such as MegaSaM~\cite{megasam}.
Their poses are compared on short Sintel, TUM-dynamics and ScanNet clips following MonST3R~\cite{monst3r}, and E3D-Bench~\cite{e3dbench} and OmniWorld~\cite{omniworld} widen these evaluations.
None asks what this shared score separates from an image-free motion prior, which is what we audit.

\paragraph{Pose metrics.}
ATE and RPE~\cite{tum,kummerle2009} are standard, and the alignment should remove exactly the degrees of freedom a method cannot observe~\cite{zhang2018tutorial}.
Work on the alignment has mainly asked what it absorbs. Common tools fit its rotation to the positions or to one reference pose~\cite{evo,zhang2018tutorial}, and global alignment can hide drift~\cite{daum2026curvature,zhang2026rtk}.
Scores that use the orientations serve other goals: robustness to outlying poses (the discernible trajectory error, DTE~\cite{lee2022ate}), joint or separate fits of positions and orientations~\cite{salas2015alignment,lee2024alignment}, flow or reprojection checks~\cite{princeton365,ore2023,uavff3d}, pairwise angular accuracy~\cite{posediffusion,vggtomega} and, concurrently, ATE relative to a static camera~\cite{noctif3r}.
None characterises what ATE itself rewards on short clips. We prove two exact properties of its position-only alignment that decide rankings there: a constant-velocity prediction scores the true path's regression residual, and rotating positions against orientations leaves the score unchanged.
Sturm et al.~\cite{tum} observed that the RPE over all intervals is at least ATE and equals it without rotational error. We prove a different, pair-pooled identity: PA-ATE, which adds the missing check to ATE itself, equals the translational RPE over all frame pairs divided by $\sqrt2$ when the orientations agree, the error that relation-based evaluation computes with every pair as a relation~\cite{kummerle2009}. Concurrently, OC-GS~\cite{ocgs} aligns cameras in the same order to evaluate novel views. DTE caps and normalises its position errors, and a joint fit weighs positions against rotations. PA-ATE fixes the rotation from the orientations alone, needs no such weight and stays in ATE's units.

\paragraph{Near-linear data.}
Benchmarks have long guarded against trajectories that are too simple.
KITTI scores odometry on 100--800\,m segments~\cite{kitti,kittiodo}, ParticleSfM~\cite{particlesfm} and Lightspeed~\cite{dynpose100k} drop static or straight camera paths, TrajNet++~\cite{trajnetpp} excludes static trajectories, and ORBIT++~\cite{orbitpp} notes that pose benchmarks have simple trajectories.
These choices rest on heuristics or fixed thresholds. Our residual SNR turns them into a measured quantity: per model and clip, it gives a closed-form condition for when straightening lowers ATE, which held out of sample.

\paragraph{Trivial baselines and audits.}
Simple baselines have recalibrated progress in motion forecasting~\cite{scholler2020cvm,martinez2017motion}, time-series forecasting~\cite{dlinear}, language and vision~\cite{vqa,gururangan2018artifacts,poliak2018hypothesis,pham2021order}, absolute pose regression~\cite{sattler2019limitations} and single-view 3D reconstruction~\cite{tatarchenko2019}. Evaluation audits have traced anomalous scores to what a metric misses~\cite{otani2019summaries,ge2024fvd,brachmann2021limits}.
Our line adds an exact account: its score equals a regression residual of the ground truth, so it is known on any clip before a model runs. The line is a diagnostic, not a certificate~\cite{feng2019partial}.

Bootstrap ranking stability~\cite{efron1979,maierhein2018,musgrave2020reality,dehghani2021lottery} and benchmark-level signal-to-noise analyses~\cite{signalnoise2025,leaderboardlottery2026} measure how reliably a whole benchmark ranks methods, whereas our residual SNR works per method and clip.

\section{A straight line on Sintel}
\label{sec:diagnosis}
We first set out the observation the paper explains.
\subsection{Protocol}
We follow the dynamic-scene pose protocol of MonST3R~\cite{monst3r}, used by CUT3R and its successors~\cite{cut3r,ttt3r}.
On Sintel, it uses the 14 \emph{final}-pass training clips of 20--50 frames.
On TUM-dynamics, it uses 90 frames taken at stride 3 from the start of 8 sequences (about 9\,s each), and the same on ScanNet~\cite{dai2017scannet}, a registered test, for the 94 of 100 test scenes with valid poses (appendix).
Each predicted trajectory is aligned to the ground truth by the similarity transform (scale, rotation and translation; Sim(3)) minimising position error (the closed-form least-squares fit of Umeyama~\cite{umeyama1991}, as implemented in the evo toolkit~\cite{evo}).
We report the RMSE of the aligned positions (ATE, in the ground truth's units: scene units on Sintel, metres on TUM and ScanNet) and the relative pose error between consecutive frames (RPE; RPE$_t$ in the ground truth's units, RPE$_r$ in degrees).
Our ATE matches the official CUT3R/TTT3R evaluator to within $5\times10^{-7}$.

\subsection{Straightening and the line}
\label{sec:line}
Let $\mathrm{cv}(\mathbf{P})$ be the least-squares constant-velocity fit of a predicted trajectory $\mathbf{P}\in\mathbb{R}^{n\times3}$ with camera centres $\mathbf{p}_i$ at times $t_i$, that is, the constant-speed straight line closest to the centres in least squares.
For $\alpha\in[0,1]$,
\begin{equation}
\mathbf{P}_\alpha=\mathrm{cv}(\mathbf{P})+(1-\alpha)\big(\mathbf{P}-\mathrm{cv}(\mathbf{P})\big)
\label{eq:straighten}
\end{equation}
keeps the constant-velocity fit of the prediction and shrinks its deviation from it by a factor $1-\alpha$, without images or ground truth. We call this \emph{straightening}: half straightening for $\alpha=0.5$, full straightening for $\alpha=1$.

Full straightening leaves a prediction $\mathbf{p}_i=\mathbf{a}+t_i\mathbf{d}$. Let $\mathbf{d}\neq 0$ and $\mathbf{g}_1,\dots,\mathbf{g}_n\in\mathbb{R}^3$ be the ground-truth centres.
Since a Sim(3) transform maps a line traversed at constant speed to another such line, the aligned prediction is the least-squares fit of $\mathbf{g}_i$ on $(1,t_i)$, and
\begin{equation}
\mathrm{ATE}_{\text{line}}=\Big(\tfrac{1}{n}\textstyle\sum_i\big\|\mathbf{g}_i-\hat{\mathbf{a}}-t_i\hat{\mathbf{d}}\big\|^2\Big)^{1/2},
\label{eq:line}
\end{equation}
where $\hat{\mathbf{a}}$ and $\hat{\mathbf{d}}$ are the least-squares coefficients. Thus \cref{eq:line} is the RMS time-regression residual for nonconstant timestamps (proof, and the case in which the fitted scale would be zero, in the appendix).
Every fully straightened prediction with nonzero fitted speed thus has the same ATE, whatever the model. We call this baseline \emph{the line}. It uses only the frame times: no images, no parameters, no ground truth. It is aligned and scored like a prediction, and the alignment supplies its direction, speed and offset (appendix).

We also use the share of motion a line explains,
\begin{equation}
R^2_{\text{line}} = 1-\mathrm{ATE}_{\text{line}}^2 \big/ \tfrac1n\textstyle\sum_i\|\mathbf{g}_i-\bar{\mathbf{g}}\|^2 ,
\end{equation}
with $\bar{\mathbf{g}}$ the mean true centre, as a property of the benchmark itself, computed from ground truth alone.

\subsection{Results}
\begin{figure}[b]
\centering
\includegraphics[width=\linewidth]{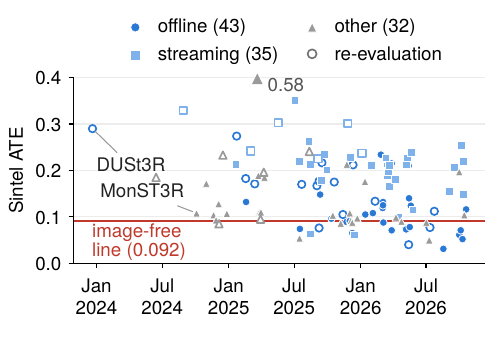}
\caption{Published Sintel ATE values admitted by our inclusion rule (appendix), one per method paper (its best variant), by the method's first arXiv date, from DUSt3R on; open markers: 29 later re-evaluations. No admitted value published before July 2025 lies below the line (MegaSaM's open marker is a Dec 2025 re-evaluation; appendix). Offline, streaming: feed-forward models; other: optimisation and SLAM/VO. Red: the line, 0.0915.}
\label{fig:progress}
\end{figure}

\Cref{tab:sintel} re-evaluates 11 recent models with their official inference code.
Eight are offline models: VGGT~\cite{vggt}, $\pi^3$ and $\pi^3$X~\cite{pi3}, Depth Anything 3 (DA3) Small, Base, Large and Giant~\cite{da3}, and MapAnything~\cite{mapanything}.
Three are streaming models: CUT3R~\cite{cut3r}, TTT3R~\cite{ttt3r} and StreamVGGT~\cite{streamvggt}.
On Sintel, the line ranks second of twelve by mean ATE: 10 of the 11 models score worse, and in 77\% of clip-bootstrap resamples (clips drawn with replacement) the line still beats at least 8 of them (appendix).
Every model beats the line on at least 6 of 8 TUM and 89 of 94 ScanNet clips.
Of the 110 methods with a published Sintel ATE that our inclusion rule admits (a value counts unless its source or code shows another setting; 81 of 101 under a stricter rule), 85 lie above the line at their reported precision (\cref{fig:progress}). This says what the score separates, not what methods achieve (appendix).

Since every fully straightened prediction scores the line's 0.092, half straightening shows how much of this a motion prior explains.
It lowers the mean Sintel ATE of all 11 models (\cref{tab:sintel}), by a median of 22\%, but not in the same way.
Seven models gain on 11 to 13 of the 14 clips and by the geometric mean over clips. The other four, among them the three with the lowest ATE, gain on 5 to 8 clips and by the geometric mean not at all: $\pi^3$ gains where it fails (cave\_2 and temple\_3), on two paths that are lines to $R^2_{\text{line}}>0.9998$ and on ambush\_5, all five with residual SNR (\cref{sec:cause}) below $\sqrt3$, and its mean falls only from 0.082 to 0.081.

\begin{table}[t]
\caption{Rows: the 11 models we re-ran with their official code, by Sintel ATE, and the image-free line. Below: five methods added later in official environments ($^\dagger$), not in \cref{tab:deletion,tab:reproj} or \cref{sec:h2h} (published values~\cite{monst3r,uniquery4r,laser,anycam,spatrackerv2,d4rt,track4world} in the appendix). Columns: errors under the official protocol, lower is better, on Sintel, TUM (8 clips) and ScanNet (94). Half: ATE after half straightening. Cl.: Sintel clips (of 14) on which it lowers the ATE. PA-ATE: \cref{sec:correction}; for the line, constant orientation along the camera axis, of six, with the lowest mean. Our CUT3R values (0.211; ScanNet 0.097) are close to the published 0.213 and 0.099~\cite{cut3r}.}
\label{tab:sintel}
\centering\small
\setlength{\tabcolsep}{1.3pt}
\begin{tabular*}{\linewidth}{@{\extracolsep{\fill}}lccccccc@{}}
\toprule
& \multicolumn{5}{c}{Sintel (14 clips)} & TUM & ScanNet \\
\cmidrule(lr){2-6}\cmidrule(lr){7-7}\cmidrule(l){8-8}
Method & ATE & half & cl. & PA-ATE & RPE$_r$ & ATE & ATE \\
\midrule
$\pi^3$                & 0.082 & 0.081 & 5 & 0.131 & 0.284 & 0.015 & 0.030 \\
Line (no img.)         & 0.092 & -- & -- & 0.540 & 0.964 & 0.148 & 0.232 \\
DA3-Giant              & 0.112 & 0.092 & 7 & 0.163 & 0.347 & 0.012 & 0.032 \\
$\pi^3$X               & 0.113 & 0.097 & 5 & 0.171 & 0.265 & 0.011 & 0.030 \\
DA3-Base               & 0.133 & 0.103 & 11 & 0.193 & 0.936 & 0.029 & 0.059 \\
DA3-Large              & 0.142 & 0.111 & 8 & 0.203 & 0.469 & 0.015 & 0.035 \\
VGGT                   & 0.171 & 0.146 & 11 & 0.204 & 0.471 & 0.012 & 0.035 \\
DA3-Small              & 0.180 & 0.121 & 12 & 0.241 & 3.779 & 0.032 & 0.073 \\
TTT3R                  & 0.208 & 0.161 & 11 & 0.249 & 0.726 & 0.029 & 0.064 \\
CUT3R                  & 0.211 & 0.163 & 12 & 0.262 & 0.630 & 0.044 & 0.097 \\
StreamVGGT             & 0.220 & 0.149 & 13 & 0.256 & 1.060 & 0.026 & 0.048 \\
MapAnything            & 0.226 & 0.179 & 13 & 0.248 & 0.641 & 0.029 & 0.053 \\
\midrule
UniQuery4R$^\dagger$   & 0.033 & 0.054 & 3 & 0.047 & 0.197 & 0.014 & 0.034 \\
LASER$^\dagger$        & 0.064 & 0.066 & 7 & 0.113 & 0.248 & 0.017 & 0.031 \\
AnyCam$^\dagger$       & 0.093 & 0.089 & 5 & 0.229 & 0.454 & 0.043 & 0.106 \\
SpatialTrackerV2$^\dagger$ & 0.115 & 0.087 & 8 & 0.173 & 0.636 & 0.018 & 0.053 \\
MonST3R$^\dagger$      & 0.118 & 0.093 & 11 & 0.232 & 0.795 & 0.106 & 0.057 \\
\bottomrule
\end{tabular*}
\end{table}

By the geometric mean over clips, which damps failures such as $\pi^3$X's on cave\_2 and temple\_3 ($\pi^3$X beats the line on 10 of 14 clips), the line still beats 8 of the 11 (appendix).
The line has no rotation, and 9 of the 11 models have a lower RPE$_r$ than its constant orientation: its advantage lies in the translation, which is what ATE scores.
Of the methods added later ($^\dagger$), UniQuery4R and LASER are below the line and keep their lead under PA-ATE (0.047 and 0.113 vs.\ 0.540). AnyCam's PA-ATE is 2.5 times its ATE, because its heading is more than 45$^\circ$ off on 3 clips.

\section{What the alignment leaves to score}
\label{sec:cause}
Two exact properties of ATE's alignment explain the observation and say when straightening pays.

\subsection{What the alignment supplies}
\label{sec:supplies}
Let $\mathbf{g}_i$ and $\mathbf{p}_i$ be the true and predicted camera centres and $\mathbf{R}_i$, $\hat{\mathbf{R}}_i$ the true and predicted camera-to-world rotations, $i=1,\dots,n$.
ATE minimises the position error over a similarity transform with scale $s$, rotation $\mathbf{Q}$ and translation $\mathbf{b}$, fitted to the positions alone~\cite{zhang2018tutorial}.
Two properties follow directly.
First, for any rotation $\mathbf{Q}'$ and translation $\mathbf{b}'$, centres $\mathbf{Q}'\mathbf{p}_i+\mathbf{b}'$ with unchanged orientations $\hat{\mathbf{R}}_i$ have exactly the same ATE, because the alignment absorbs this transform (proof in the appendix).
Rotating the positions alone changes how the cameras move relative to where they look, and ATE does not score that relationship.
Second, as shown in \cref{sec:line}, a constant-velocity prediction scores the true path's regression residual, whatever its direction and speed.

How much is left depends on the clip.
On the 14 Sintel clips the median $R^2_{\text{line}}$ is 0.988 (\cref{tab:deletion}): a constant velocity explains 98.8\% of the positions' variance.
We call \emph{heading} the direction of the constant-velocity fit of the centres, expressed in each frame's camera axes, and measure its error as the median over frames of the angle between predicted and true heading.
The models do recover it: the median error over model--clip pairs is 6.6$^\circ$ on Sintel, 4.1$^\circ$ on TUM, 4.7$^\circ$ on ScanNet and 1.2$^\circ$ on Wild-SLAM~\cite{wildslam}.
ATE gives no credit for it, and no penalty when it is wrong: on 26 of the 154 Sintel model--clip pairs the heading is more than 45$^\circ$ off. On temple\_3, ATE ranks $\pi^3$X first despite a 104$^\circ$ heading error, and PA-ATE ranks first DA3-Giant, 13$^\circ$ off (appendix).

\subsection{Where straightening pays}
\label{sec:pays}
On Sintel, applied only to the weaker model of a pair, half straightening reverses 22 of the 55 pairwise comparisons: DA3-Small, for example, moves from 0.180 to 0.121 and overtakes VGGT (0.171).
On the official TUM and ScanNet clips and on full-length Bonn~\cite{bonn} and Wild-SLAM sequences, the same step raises every model's mean ATE (\cref{tab:deletion}).

\begin{table}[t]
\caption{Half straightening ($\alpha=0.5$ in \cref{eq:straighten}) on eight benchmarks (rows): five for the 11 models and, below, three registered before 9 models ran on them (\cref{sec:newbench}). $R^2_{\text{line}}$: median over clips, from ground truth. Models gaining: of the 11 (below: of the 9), those whose mean ATE decreases. Predicted: by \cref{eq:threshold} from each pair's $S$, $N$ and $c$ before straightening. Observed: after it (clip-bootstrap 95\% interval on Sintel 5--11). Ratio: median over models of ATE after\,/\,before. SNR: median over model--clip pairs of the residual SNR (\cref{eq:snr}). Bonn: camera poses via its calibration.}
\label{tab:deletion}
\centering\small
\setlength{\tabcolsep}{2pt}
\begin{tabular*}{\linewidth}{@{\extracolsep{\fill}}lcccccc@{}}
\toprule
& & & \multicolumn{2}{c}{models gaining} & & median \\
\cmidrule(lr){4-5}
Benchmark & clips & $R^2_{\text{line}}$ & predicted & observed & ratio & SNR \\
\midrule
Sintel (official) & 14 & 0.99 & 10 & 11 & 0.78 & 0.86 \\
TUM (official)    & 8  & 0.51 & 0  & 0  & 2.65 & 6.1 \\
ScanNet (official) & 94 & 0.67 & 0 & 0  & 2.25 & 5.7 \\
Bonn, full        & 5  & 0.36 & 0  & 0  & 1.45 & 2.7 \\
Wild-SLAM, full   & 10 & 0.23 & 0  & 0  & 6.66 & 27 \\
\midrule
Lightspeed        & 36 & 0.99 & 5  & 5  & 0.78 & 1.3 \\
DyCheck           & 12 & 0.14 & 0  & 0  & 3.88 & 14 \\
ADT               & 8  & 0.33 & 1  & 2  & 1.74 & 6.7 \\
\bottomrule
\end{tabular*}
\end{table}

\paragraph{A residual signal-to-noise ratio.}
Write $\mathrm{r}(\cdot)=(\cdot)-\mathrm{cv}(\cdot)$ for the part of a trajectory a constant velocity does not explain, $\mathrm{rms}(\mathbf{X})=\|\mathbf{X}\|_F/\sqrt{n}$, with $\|\cdot\|_F$ the Frobenius norm, for the root of the mean over frames of the squared Euclidean norm, $\mathbf{G}$ for the true trajectory and $\mathbf{E}=\mathbf{G}-\hat{\mathbf{P}}$ for the error of the aligned prediction $\hat{\mathbf{P}}$, and define
\begin{equation}
\mathrm{SNR}=\frac{S}{N}=\frac{\mathrm{rms}\,\mathrm{r}(\mathbf{G})}{\mathrm{rms}\,\mathrm{r}(\mathbf{E})}.
\label{eq:snr}
\end{equation}
$S$ depends on the ground truth only and equals the line's ATE. $N$ also needs it, so the SNR is a diagnostic, not a test-time correction.
With the alignment held fixed, straightening changes the residual error to \mbox{$\alpha\,\mathrm{r}(\mathbf{G})+(1-\alpha)\,\mathrm{r}(\mathbf{E})$} while the constant-velocity error stays the same. Define $c=\langle\mathrm{r}(\mathbf{G}),\mathrm{r}(\mathbf{E})\rangle_F/(nSN)$, with $\langle\cdot,\cdot\rangle_F$ the Frobenius inner product, so that $c$ is the cosine between the two residuals, taking $c=0$ when $S=0<N$, and $\beta=(1-\alpha)/\alpha$. For $\alpha\in(0,1]$ and $N>0$, the fixed-alignment error decreases exactly when
\begin{equation}
\mathrm{SNR}<-\beta c+\sqrt{\beta^2c^2+(2-\alpha)/\alpha},
\label{eq:threshold}
\end{equation}
which for uncorrelated errors ($c=0$) becomes
\begin{equation*}
\mathrm{SNR}<\sqrt{(2-\alpha)/\alpha}\,,\qquad\text{that is }\sqrt3\text{ for }\alpha=0.5.
\end{equation*}
At the fixed alignment, full straightening ($\alpha=1$) helps exactly when the model's departure from constant velocity is further from the true departure than no departure is ($N>S$). As in shrinkage~\cite{jamesstein1961}, half straightening already helps, for uncorrelated errors, once $N$ exceeds $S/\sqrt3$, 58\% of the line's ATE, so a gain does not require $N>S$.

\begin{figure}[t]
\centering
\includegraphics[width=\linewidth]{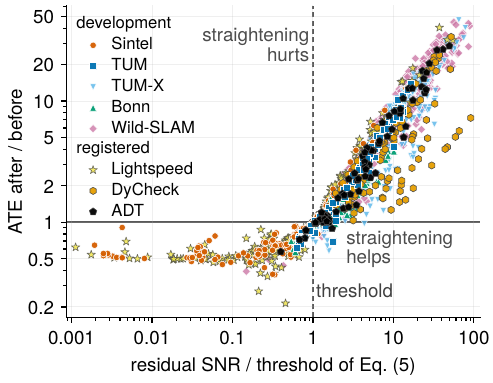}
\caption{Effect of half straightening (\cref{eq:straighten}) against residual SNR (\cref{eq:snr}), for every development pair and every pair of the three benchmarks registered later (outlined markers, \cref{sec:newbench}). The SNR is divided by the threshold of \cref{eq:threshold} with each pair's measured $c$, so straightening lowers the ATE below 1 (dashed).}
\label{fig:snr}
\end{figure}

\Cref{eq:threshold} is exact at fixed alignment and sufficient after the alignment is refitted (proof in the appendix). The outcome thus depends on the clip's nonlinear motion $S$, a property of the benchmark, the model's residual error $N$, and their correlation $c$, and the tests check whether these, not straightness or error alone, decide it.
We developed the criterion on 1056 model--clip pairs of Sintel, TUM, ten more hand-held TUM sequences (TUM-X), and Bonn and Wild-SLAM, at full length and in 9- and 15-s windows (\cref{fig:snr}), and tested it on 2- and 10-s windows of KITTI~\cite{kitti} and 2- and 5-s windows of 7-Scenes~\cite{sevenscenes} and on ScanNet, with the windows and a pass mark for the threshold with $c=0$ registered before any model ran (appendix). \Cref{tab:survey} collects every test.
With the measured $c$, the outcome after refitting is the one the criterion gives for 95\% of the development pairs, 98\% of the 792 held-out pairs, 97\% of the 624 pairs of the added methods and 97\% of the 1034 ScanNet pairs. Every miss is a gain that appears only after refitting, and the refit's first-order term (a first-order approximation of the part of the straightened error that refitting the alignment absorbs) recovers all but one, a 0.4\% gain (\cref{tab:survey}, +refit). On ScanNet gains are rare (5\%), so predicting none already gives 95\%. On the held-out windows, where 38\% of pairs gain, thresholds on straightness $R^2_{\text{line}}$ alone or on the error $N$ alone, fitted on the development pairs, give 61\% and 62\%, as does always predicting no gain. With $c=0$, the fixed threshold $\sqrt3$ is right for 90 to 95\% of each pooled set (pass mark 85\%).

\begin{table}[t]
\caption{Every test of the straightening criterion, per model--clip pair. Rows: development pairs, KITTI and 7-Scenes windows registered before any model ran, the registered ScanNet test, methods added later (MonST3R and the four of \cref{tab:sintel}, LASER and AnyCam in two configurations each), the criterion applied to PA-ATE on Sintel, TUM and Wild-SLAM, and three benchmarks registered in advance, with 9 models (\cref{sec:newbench}). SNR: median residual SNR (\cref{eq:snr}). Gain: \% of pairs whose ATE falls under half straightening. Correct: \% of pairs whose outcome after refitting is given by a threshold fitted on the development pairs on straightness alone or on the model's error $N$ (relative to the true centres' RMS spread) alone, by \cref{eq:threshold} with $c=0$ ($\sqrt3$) or the measured $c$ (criterion), and by the criterion with the refit's first-order term (+refit). Details in the appendix.}
\label{tab:survey}
\centering\small
\setlength{\tabcolsep}{1.5pt}
\begin{tabular*}{\linewidth}{@{\extracolsep{\fill}}lcccccccc@{}}
\toprule
& & & & \multicolumn{5}{c}{correct (\%)} \\
\cmidrule(l){5-9}
Set & pairs & SNR & gain & $R^2_{\text{line}}$ & error & $c=0$ & criterion & +refit \\
\midrule
Development & 1056 & 7.4 & 20 & 84 & 88 & 94 & 95 & 99.9 \\
Held-out & 792 & 2.6 & 38 & 61 & 62 & 90 & 98 & 100 \\
\midrule
ScanNet & 1034 & 5.7 & 5 & 94 & 94 & 95 & 97 & 100 \\
Added methods & 624 & 4.5 & 14 & 89 & 84 & 90 & 97 & 100 \\
PA-ATE & 352 & 3.3 & 32 & 80 & 77 & 92 & 97 & 100 \\
\midrule
Lightspeed & 324 & 1.3 & 55 & 63 & 62 & 95 & 99.7 & 100 \\
DyCheck & 108 & 14 & 1 & 93 & 98 & 99 & 100 & 100 \\
ADT & 72 & 6.7 & 17 & 83 & 81 & 92 & 90 & 100 \\
New, all & 504 & 3.0 & 38 & 72 & 73 & 95 & 98 & 100 \\
\bottomrule
\end{tabular*}
\end{table}

\subsection{Three benchmarks registered in advance}
\label{sec:newbench}
Lightspeed~\cite{dynpose100k}, DyCheck~\cite{dycheck} and ADT~\cite{adt,tapvid3d} also score dynamic-scene poses by ATE (synthetic clips under DynPose-100K's protocol, MegaSaM's copy~\cite{megasam} of the phone videos at Easi3R's stride~\cite{easi3r}, and Easi3R's egocentric clips; appendix). Before any model ran, we registered the protocols, the pass mark and, from the ground truth alone, that Lightspeed, as straight as Sintel ($R^2_{\text{line}}$ 0.99) although its clips were chosen to exclude trivially linear paths~\cite{dynpose100k}, would side with Sintel, and that on DyCheck and ADT, less straight than TUM (0.14 and 0.33), no model would gain and every model would beat the line. Over the 504 pairs of the eight offline models and VGGT-$\Omega$~\cite{vggtomega}, the criterion gives the outcome for 99.7, 100 and 90\% of the Lightspeed, DyCheck and ADT pairs, and with the refit term for all (\cref{tab:deletion,tab:survey}).
As registered, Lightspeed sides with Sintel. Half straightening lowers the mean ATE of 5 of the 9 models, by a median ratio of 0.78, and the line (0.099) beats the same 5, VGGT among them (0.223, published 0.226~\cite{spatrackerv2}), and 10 of the 11 methods in the two published tables of this protocol~\cite{dynpose100k,spatrackerv2}\nocite{dynapo}, MegaSaM (0.105) included. On DyCheck and ADT every model beats the line, as registered, and none gains on DyCheck, though on ADT the two weakest gain where the ground truth alone predicted none and the criterion with each model's error, refit included, gives both.

\section{What low ATE does not guarantee}
\label{sec:consequence}

If a motion prior can lower ATE, does a lower ATE still mean better camera geometry? ATE is a proxy.
What a downstream user of the poses needs is that a static scene point seen in one frame lands where it is seen in another.
Reprojection already scores poses in relocalisation~\cite{wald2020rio10,arnold2022mapfree,brachmann2021limits} and in other work~\cite{princeton365,ore2023}. It checks ATE's conclusions with the true depth of Sintel, TUM and ScanNet.

\paragraph{Protocol.}
For each source frame $i$ we take a fixed $40\times30$ grid of pixels with valid ground-truth depth and back-project them with that frame's true intrinsics (appendix).
A point is kept as static and visible in a target frame $j$ if, projected with the true poses into every frame from $i$ to $j$, it stays in the image, agrees with the ground-truth depth there within 2\% (Sintel) or 3\% (TUM; ScanNet, at the target frame), and is not marked as moving.
Moving points are found by ground-truth flow and occlusion on Sintel~\cite{wulff2017rigid} and person masks~\cite{maskrcnn} on TUM. ScanNet's scenes are static.
The prediction places the point at $\hat{\mathbf{R}}_j^{\top}\big(\hat{\mathbf{R}}_i\mathbf{X}_i+\lambda(\mathbf{p}_i-\mathbf{p}_j)\big)$, projected with the true intrinsics, where $\mathbf{X}_i$ is the point in camera $i$ and $\lambda>0$ is a single scale per clip and frame gap, the median ratio of true to predicted camera displacement over the evaluated pairs. No rotation or per-frame quantity is fitted, so of the alignment's degrees of freedom the check discounts only scale. The line keeps a constant orientation and moves along the optical axis.
Frame pairs with fewer than 20 static points are skipped, and points placed behind the camera are penalised (both rare; appendix).
The error of a frame pair is the median over its points. We average it over the pairs $(i,i+5)$ of a clip and aggregate clips by arithmetic and geometric means (appendix).

\begin{table}[t]
\caption{Rows: three scores, arithmetic means over clips as in the protocol. PA-ATE: \cref{sec:correction}. Reprojection error: of static points, pairs 5 frames apart. Columns: three tests on each benchmark. Rank: the image-free line's rank among itself and the 11 models (of 12, 12 is last). Rot.: of the 11 models, those whose score worsens when their centres are rotated by 30$^\circ$ about their centroid and the $y$-axis of the world frame, orientations kept. Str.: of the 11 models, those whose score improves when their positions are replaced by their constant-velocity fit ($\alpha=1$). Under geometric means the line ranks 4th by ATE but last under both other scores on Sintel, and none of the 11 models reprojects better when straightened. Ranges over 26 settings of gap, filter and scale (4 on ScanNet) in the appendix.}
\label{tab:reproj}
\centering\small
\setlength{\tabcolsep}{2pt}
\begin{tabular*}{\linewidth}{@{\extracolsep{\fill}}lccccccccc@{}}
\toprule
& \multicolumn{3}{c}{Sintel (14 clips)} & \multicolumn{3}{c}{TUM (8)} & \multicolumn{3}{c}{ScanNet (94)} \\
\cmidrule(lr){2-4}\cmidrule(lr){5-7}\cmidrule(l){8-10}
Score & rank & rot. & str. & rank & rot. & str. & rank & rot. & str. \\
\midrule
ATE                  & 2 & 0 & 10 & 12 & 0 & 0 & 12 & 0 & 0 \\
\textbf{PA-ATE}      & \textbf{12} & \textbf{11} & \textbf{6} & \textbf{12} & \textbf{11} & \textbf{0} & \textbf{12} & \textbf{11} & \textbf{0} \\
reprojection error   & 11 & 11 & 3 & 12 & 11 & 0 & 12 & 11 & 0 \\
\bottomrule
\end{tabular*}
\end{table}

\paragraph{Results.}
\Cref{tab:reproj} compares the three scores on the same clips (plots of ATE against reprojection error in the appendix).
On Sintel the line ranks fourth by ATE and last by reprojection under geometric means, 33\,px vs.\ 6 to 26\,px (worst in 77\% of clip resamples), and second and eleventh under arithmetic means (appendix, also for an angular error).
On TUM the line is at 40\,px against 2 to 6\,px, and on ScanNet (appendix) at 81\,px vs.\ 3 to 11\,px. Part of the line's error comes from its constant orientation, which RPE$_r$ already penalises. With the true relative rotations and its own camera-frame displacements, it still reprojects worst on all three datasets (12.5\,px vs.\ 2.8 to 11.8\,px on Sintel; appendix).
Rotating each trajectory by 30$^\circ$ against its orientations leaves ATE unchanged and raises the reprojection error of every model on all three datasets, by 8 to 13\,px on Sintel, 0.8 to 2.4\,px on TUM and 7 to 11\,px on ScanNet.
The cost lies in the translation: with the true rotations, the rotated trajectories are again worse for all 11 models (appendix).
Full straightening lowers the Sintel ATE of 10 models but the reprojection error of only 3 (geometric means: 8 and 0).

On Sintel, PA-ATE agrees with the reprojection ranking better than ATE does: under geometric means, Kendall $\tau$ (rank correlation) 0.82 vs.\ 0.48 with the line included (paired bootstrap interval of the difference 0.15 to 0.42) and 0.78 vs.\ 0.67 among the 11 models ($-0.04$ to 0.22). On TUM and ScanNet all three scores agree (0.78 and 0.93 among the models). Per clip, both scores track the reprojection of unmodified predictions (Spearman 0.59 to 0.86). With rotated and straightened predictions, PA-ATE still does (0.61 on Sintel, 0.82 on ScanNet), ATE less (0.32, 0.68; difference intervals 0.16 to 0.45, 0.10 to 0.17), tying on TUM (0.77, 0.76).

\section{Scoring translation relative to orientation}
\label{sec:correction}
Scoring cannot add information the clips lack, but one change to the alignment stops it ignoring the orientations.

\subsection{What can repair the omission}
\label{sec:repair}
Three existing scores use the orientations: a rotation error computed under the same position alignment; the translational relative pose error RPE$_t$, which expresses each frame pair's displacement in the predicted camera frame~\cite{tum,zhang2018tutorial}; and DTE, which robustly estimates the alignment's centre, rotation and scale from positions and orientations~\cite{lee2022ate}.

We make one change and keep the rest of ATE. The alignment's rotation is fitted to the predicted orientations rather than to the positions,
\begin{equation}
\mathbf{Q}^*\in\arg\min_{\mathbf{Q}\in SO(3)}\sum_i\big\|\mathbf{R}_i-\mathbf{Q}\hat{\mathbf{R}}_i\big\|_F^2,
\label{eq:anchor}
\end{equation}
which is the chordal mean~\cite{hartley2013rotation} of the per-frame rotation offsets (the anchor), in closed form from the SVD of $\sum_i\mathbf{R}_i\hat{\mathbf{R}}_i^{\top}$. Scale and translation are then fitted to the positions as before,
\begin{equation}
(s^*,\mathbf{b}^*)\in\arg\min_{s\ge0,\;\mathbf{b}}\;\sum_i\big\|\mathbf{g}_i-s\,\mathbf{Q}^*\mathbf{p}_i-\mathbf{b}\big\|^2,
\label{eq:anchor_fit}
\end{equation}
and the RMS of $\mathbf{g}_i-s^*\mathbf{Q}^*\mathbf{p}_i-\mathbf{b}^*$ is the \emph{pose-anchored ATE} (PA-ATE). We report it with the rotation error under the same $\mathbf{Q}^*$, the RMS over frames of the angle of $\mathbf{R}_i^{\top}\mathbf{Q}^*\hat{\mathbf{R}}_i$.
It still removes the seven degrees of freedom a monocular method cannot observe~\cite{zhang2018tutorial}. It is never below ATE, which minimises over the rotation, and the gap is the cost of taking the rotation from the orientations (0.131 against 0.082 for $\pi^3$).
Because $s^*\ge0$, a prediction whose anchored positions are anti-correlated with the true ones (10 of the 154 Sintel model--clip pairs) is scored as a static camera, at the true centres' RMS distance from their mean (appendix).
Units and aggregation are those of ATE, at about its cost (37 against 34\,\textmu s per clip), so saved predictions can be re-scored.

PA-ATE is also a relative error. Let $\boldsymbol\varepsilon_{ij}$ be the error of the displacement from camera $i$ to camera $j$ in camera $i$'s frame, the translational RPE of the pair with the anchored scale. PA-ATE equals the RMS of $\boldsymbol\varepsilon_{ij}$ over all $n^2$ ordered pairs, the pairs $(i,i)$ included with zero error, divided by $\sqrt2$ whenever the anchored orientations agree with the true ones, and differs from it by at most $2\sin(\theta/2)$ RMS radii for the largest rotation error $\theta$ (appendix).

\subsection{Head-to-head comparison}
\label{sec:h2h}
\Cref{tab:compare} applies six scores to the same tests, and all five orientation-aware scores detect the rotated trajectory.
Consecutive-frame RPE$_t$, the version this protocol reports, is dominated by frame-to-frame jitter: a static camera beats 8 of the 11 models.
RPE$_t$ over 10 frames, DTE and PA-ATE rank every image-free reference below all 11 models (ATE: below 1), and fewer orders flip when the weaker model is straightened. Only PA-ATE does so in ATE's form, and with its rotation error it costs no more than the others.

\begin{table}[t]
\caption{Six scores on the Sintel tests, 11 models. Detects rotation: true centres rotated 20$^\circ$ against the orientations score worse. ATE's form: RMS position error, one global alignment. \textmu s per clip: one CPU thread, alone and with the rotation error. $>$ref: models beating every image-free reference. Flips: of 55 pairs, when the weaker model is half straightened. $\tau_{\text{ref}}$: Kendall $\tau$ of the Sintel ranking with the ATE rank on TUM and Wild-SLAM. --: undefined.}
\label{tab:compare}
\centering\small
\setlength{\tabcolsep}{1.5pt}
\begin{tabular*}{\linewidth}{@{\extracolsep{\fill}}lccccccc@{}}
\toprule
& detects & ATE's & \multicolumn{2}{c}{\textmu s per clip} & & & \\
\cmidrule(lr){4-5}
Score & rotation & form & alone & +rot. & $>$ref & flips & $\tau_{\text{ref}}$ \\
\midrule
ATE (positions)            & $\times$     & $\checkmark$ & 34 & 66 & 1   & 22 & .54 \\
ATE's rotation error       & $\checkmark$ & $\times$     & 66 & --   & --  & 0  & .54 \\
RPE$_t$, 1 frame           & $\checkmark$ & $\times$     & 40 & 68 & 3   & 18 & .69 \\
RPE$_t$, 10 frames         & $\checkmark$ & $\times$     & 38 & 66 & 11  & 9  & .65 \\
DTE~\cite{lee2022ate}      & $\checkmark$ & $\times$     & $1.3{\times}10^5$ & --   & 11  & 9  & .65 \\
\textbf{PA-ATE} & $\pmb{\checkmark}$ & $\pmb{\checkmark}$ & \textbf{37} & \textbf{61} & \textbf{11} & \textbf{10} & \textbf{.69} \\
\bottomrule
\end{tabular*}
\end{table}

RPE$_t$ over 10 frames scores the frame pairs at one gap and PA-ATE all of them: on ScanNet, the former follows the reprojection more closely for frames 5 apart and PA-ATE at 6\,s (appendix). We therefore recommend ATE for comparability with published values, PA-ATE with its rotation error for global pose, and RPE$_t$ at stated gaps for local motion.

\subsection{What changes for the benchmarks}
\label{sec:benchchange}
PA-ATE largely preserves the order of the models.
The two scores order the 11 models similarly on Sintel, TUM, Wild-SLAM, ScanNet and 7-Scenes (Kendall $\tau$ 0.85, 0.85, 1.00, 1.00 and 0.78 to 0.96), and on Sintel no pair of models is ordered oppositely with bootstrap support under the two. Its heading check also caught that Bonn's true orientations are those of the markers: with the published calibration applied, the median heading error of 8 of the 11 models falls from about 120$^\circ$ to 1 to 9$^\circ$.
On Lightspeed, in 20 of 36 clips the stored orientations mirror the models'. On the 19 with enough matches, rotations from the frames lie a median 1.6$^\circ$ from the four best models and 4.9$^\circ$ from the stored poses, further by over 0.5$^\circ$ on 15 clips (the models on one; appendix).

\section{Beyond camera poses}
\label{sec:beyond}
Both properties extend to any score that minimises an error over an alignment (appendix): nothing the alignment absorbs is scored, and an input-free template whose orbit closes to an affine subspace scores the ground truth's distance to it (\cref{tab:templates}). On KITTI the mean layout beats NVDS~\cite{nvds} under DepthCrafter's own code~\cite{depthcrafter}, NVDS and CasualSAM~\cite{casualsam} under MonST3R's~\cite{monst3r}, and MiDaS~\cite{midas} aligned per image as in Marigold~\cite{marigold}. On SSP-3D~\cite{straps} the SMPL mean shape beats the five general pose-and-shape regressors that HuManiFlow compares~\cite{humaniflow}, a class that lacks training data with varied body shapes~\cite{straps}, but not the four methods trained for shape. An estimate whose departure from the mean does not follow the truth's scores worse than the mean, just as straightening lowers ATE on Sintel. MICA and OptiFaces noted the mean FLAME face's strength on NoW~\cite{mica,optifaces,ringnet,nowleaderboard}. SSP-3D normalises and NoW refines by ICP, so neither has the closed form, but there too a template's score depends on the ground truth alone. PA-ATE's anchor follows the rule behind SI-SDR~\cite{leroux2019sdr}, world-space human errors~\cite{drazic2024procrustes}, NoW-Metric~\cite{mica} and per-video depth alignment~\cite{depthcrafter,videodepthanything}: align only what the task may ignore.

\section{Recommendations and limitations}
\label{sec:recs}
We close with what to report and where our analysis stops.

\paragraph{Reporting pose accuracy on short clips.}
(1) Next to ATE, report PA-ATE (\cref{eq:anchor_fit}) and its rotation error per clip and as arithmetic and geometric means. (2) Report the benchmark's $R^2_{\text{line}}$, the straightening test (\cref{eq:straighten}) and the residual SNR (\cref{eq:snr}) to show when a motion prior alone gains. (3) Claim a difference only when clip resampling supports it, and with true depth check poses by reprojection (\cref{sec:consequence}). (4) For any score that aligns before it scores, report an input-free template under the same alignment. \emph{Limitations:} PA-ATE needs predicted camera orientations. Our templates are simple: one that wins shows that a benchmark is exposed, one that loses does not show it is safe, and beyond poses we test no fix.

\section{Conclusion}
When a score aligns a prediction before scoring it, an input-free template scores the ground truth's distance from the alignment's reach, known before any model runs. Camera pose is the central case: an image-free constant-velocity line beats 10 of 11 recent models on Sintel clips, and the residual SNR with the alignment's first-order refit accounts for all but one of 4010 straightening outcomes. Because the alignment ignores where cameras look, rotating a trajectory against its orientations leaves ATE unchanged but worsens every model's reprojection. PA-ATE closes this gap with a one-function change at about ATE's cost. It ranks every model above every image-free reference on Sintel, tracks the reprojection error under controlled rotation and straightening, largely keeps the models' order where ATE was sound, and exposed Lightspeed orientations that the images contradict. Templates likewise compete on KITTI video depth, SSP-3D body shape and NoW faces. Reported next to ATE with the benchmark's straightness, PA-ATE lets models that recover the camera receive the ranking their geometry earns.

\section*{LLM Usage Statement}
Claude (Anthropic) and GPT (OpenAI) were used as agents, under the authors' direction, for writing, related work, code, experiments, figures, data collection and ideation. Gemini (Google) gave reviewer-style feedback on the manuscript. A script checks the reported numbers against the saved results, which we will release with the code. The authors take full responsibility for the content of the paper.

{
    \small
    \nocite{leapvo,mona,rodynrf,gcut3r,ss3d,vlrc,sam2,murphy1988,benchdepth,wu2025deptheval,leroy2020smply,freihand}
    \bibliographystyle{ieeenat_fullname}
    \bibliography{main}
}
\clearpage
\onecolumn
\appendix
\setcounter{section}{0}\setcounter{table}{0}\setcounter{figure}{0}\setcounter{equation}{0}
\renewcommand{\thesection}{S\arabic{section}}\renewcommand{\thetable}{S\arabic{table}}\renewcommand{\thefigure}{S\arabic{figure}}\renewcommand{\theequation}{S\arabic{equation}}
\captionsetup[table]{position=bottom,skip=4pt}
\begin{center}{\Large\bf Appendix}\end{center}
\noindent Sections, equations, tables and figures numbered with the prefix S are in this appendix, the others in the main paper. Every table and almost every number below is generated from saved predictions and source records by the scripts named in each paragraph (the exceptions are console outputs of the named scripts). File paths refer to the released code, whose tools/verify\_numbers.py checks numerical claims against the saved results.

\section{Proofs}
\label{sec:s_proofs}
The definitions used here are those of the main paper: the constant-velocity line (Eq.~(\mref{eq:line})), straightening (Eq.~(\mref{eq:straighten})), the residual SNR (Eq.~(\mref{eq:snr})), the threshold (Eq.~(\mref{eq:threshold})) and the orientation anchor (Eq.~(\mref{eq:anchor})) of pose-anchored ATE (PA-ATE).
Propositions 1 to 4 were checked numerically on every saved model--clip pair of the development benchmarks (tools/verify\_propositions.py; results/proposition\_checks) and independently against brute-force optimisation on random and degenerate inputs (tools/check\_props.py), and Proposition~\ref{prop:allpairs} is checked after its proof.
{\let\origsection\section\renewcommand{\section}[1]{}
\newcounter{proofprop}
\crefname{proofprop}{Proposition}{Propositions}
\Crefname{proofprop}{Proposition}{Propositions}
\section{Proofs}
\label{sec:proofs}


\paragraph{Notation and alignment convention.}
Let $n\ge2$, let $t_1,\ldots,t_n\in\mathbb R$ be timestamps that are not all equal, and let
$\mathbf X\in\mathbb R^{n\times3}$ denote a trajectory whose $i$th row is
$\mathbf x_i^\top$, with $\mathbf x_i\in\mathbb R^3$ a column vector. $\mathbf P$ and $\mathbf G$ denote the predicted and true trajectories in this form.
For two such arrays define
\[
 \langle\mathbf X,\mathbf Y\rangle_F
 =\sum_{i=1}^n\mathbf x_i^\top\mathbf y_i,
 \qquad
 \|\mathbf X\|_F^2=\langle\mathbf X,\mathbf X\rangle_F,
 \qquad
 \operatorname{rms}(\mathbf X)=\frac{\|\mathbf X\|_F}{\sqrt n}.
\]
The RMS averages squared Euclidean errors over frames, not over the $3n$
individual coordinates.  Write $\mathbf1\in\mathbb R^n$ for the all-ones vector,
$\bar{\mathbf x}=n^{-1}\sum_i\mathbf x_i$, and
$\mathbf X_c=\mathbf X-\mathbf1\bar{\mathbf x}^{\top}$.
Let $SO(3)=\{\mathbf Q\in\mathbb R^{3\times3}:
\mathbf Q^\top\mathbf Q=\mathbf I,\ \det\mathbf Q=1\}$.
The pointwise map $A_{s,\mathbf Q,\mathbf b}(\mathbf x)=s\mathbf Q\mathbf x+\mathbf b$
acts on an array as
$A_{s,\mathbf Q,\mathbf b}(\mathbf X)=s\mathbf X\mathbf Q^\top+\mathbf1\mathbf b^\top$.
We use the closed, nonnegative-scale convention
\begin{equation}
 \operatorname{ATE}(\mathbf P;\mathbf G)
 =\min_{s\ge0,\,\mathbf Q\in SO(3),\,\mathbf b\in\mathbb R^3}
   \operatorname{rms}\bigl(\mathbf G-A_{s,\mathbf Q,\mathbf b}(\mathbf P)\bigr).
 \label{eq:proof_ate_definition}
\end{equation}
At $s=0$ this permits a collapsed, constant trajectory. This boundary map is
not an invertible similarity.  If strictly positive scales are required, write
\[
 \operatorname{ATE}_{>0}(\mathbf P;\mathbf G)
 =\inf_{s>0,\,\mathbf Q\in SO(3),\,\mathbf b\in\mathbb R^3}
   \operatorname{rms}\bigl(\mathbf G-A_{s,\mathbf Q,\mathbf b}(\mathbf P)\bigr).
\]
These two values always agree, but the latter infimum need not be attained.

To justify both assertions, for fixed $s,\mathbf Q$ expand the squared error as
\[
 \|\mathbf G-s\mathbf P\mathbf Q^\top-\mathbf1\mathbf b^\top\|_F^2
 =\|\mathbf G_c-s\mathbf P_c\mathbf Q^\top\|_F^2
   +n\|\bar{\mathbf g}-s\mathbf Q\bar{\mathbf p}-\mathbf b\|^2.
\]
The cross term vanishes because the centred arrays have zero row sum.
Thus the optimal offset is
$\mathbf b=\bar{\mathbf g}-s\mathbf Q\bar{\mathbf p}$.
If $\mathbf P_c=\mathbf0$, the remaining error is independent of $s,\mathbf Q$
and its minimum is attained even with $s>0$.
Otherwise, the reverse triangle inequality gives
\[
 \|\mathbf G_c-s\mathbf P_c\mathbf Q^\top\|_F
 \ge \bigl|s\|\mathbf P_c\|_F-\|\mathbf G_c\|_F\bigr|.
\]
Since $s=0$ achieves error $\|\mathbf G_c\|_F$, a minimiser can be sought with
$0\le s\le2\|\mathbf G_c\|_F/\|\mathbf P_c\|_F$.
This interval and $SO(3)$ are compact, and the objective is continuous, so
the nonnegative-scale minimum exists.  The positive-scale infimum is at
least this minimum.  If a minimiser has positive scale, the two are equal.
If a minimiser has zero scale, replacing its scale by $\varepsilon>0$ and
using the corresponding optimal offset gives errors converging to that
minimum as $\varepsilon\downarrow0$, proving equality in the remaining case.
This argument establishes equality of the values, not existence of a
positive-scale minimiser.

\paragraph{The constant-velocity projection.}
Define the design matrix
$\mathbf B=[\mathbf1\ \mathbf t]\in\mathbb R^{n\times2}$,
where $\mathbf t=(t_1,\ldots,t_n)^\top$.
The timestamp assumption makes its two columns linearly independent.
Its Moore--Penrose inverse is therefore
$\mathbf B^+=(\mathbf B^\top\mathbf B)^{-1}\mathbf B^\top$.
In particular $\mathbf B^+\mathbf B=\mathbf I_2$.
Set
\[
 \mathbf H=\mathbf B\mathbf B^+,
 \qquad
 \operatorname{cv}(\mathbf X)=\mathbf H\mathbf X,
 \qquad
 \operatorname{r}(\mathbf X)=(\mathbf I_n-\mathbf H)\mathbf X.
\]
This $\operatorname{cv}$ is the least-squares constant-velocity fit: differentiating $\|\mathbf X-\mathbf B\mathbf C\|_F^2$ with respect
to the $2\times3$ coefficient matrix $\mathbf C$ gives the normal equation
$\mathbf B^\top\mathbf B\mathbf C=\mathbf B^\top\mathbf X$, with the unique
solution $\mathbf C=\mathbf B^+\mathbf X$.
Direct multiplication gives
$\mathbf H^\top=\mathbf H$, $\mathbf H^2=\mathbf H$, and
$\mathbf H\mathbf1=\mathbf1$.
Consequently, for all trajectories $\mathbf X,\mathbf Y$,
\[
 \langle\mathbf H\mathbf X,(\mathbf I_n-\mathbf H)\mathbf Y\rangle_F
 =\operatorname{tr}\!\left(\mathbf X^\top\mathbf H
                    (\mathbf I_n-\mathbf H)\mathbf Y\right)=0.
\]
Thus $\operatorname{cv}$ is the orthogonal projection onto the trajectories
$\mathbf x_i=\mathbf a+t_i\mathbf d$.

\refstepcounter{proofprop}
\paragraph{Proposition 1 (position-only alignment is invariant to a position-only rigid transform).}
\label{prop:ate_position_invariance}
\emph{Let $\mathbf W\in SO(3)$ and $\mathbf h\in\mathbb R^3$, and set
$\mathbf p'_i=\mathbf W\mathbf p_i+\mathbf h$ while leaving the predicted
camera-to-world orientations $\hat{\mathbf R}_i\in SO(3)$ unchanged.
Then
$\operatorname{ATE}(\mathbf P';\mathbf G)=\operatorname{ATE}(\mathbf P;\mathbf G)$.
The same equality holds for $\operatorname{ATE}_{>0}$.
Camera-frame displacements, translational relative pose error, and relative
static-point reprojection can nevertheless change, but they are not required
to change for every $\mathbf W,\mathbf h$ or every trajectory.}

\emph{Proof.}
For any admissible $s,\mathbf Q,\mathbf b$,
\[
 s\mathbf Q\mathbf p'_i+\mathbf b
 =s(\mathbf Q\mathbf W)\mathbf p_i+s\mathbf Q\mathbf h+\mathbf b.
\]
For each fixed $s$, the parameter map
\[
 (\mathbf Q,\mathbf b)\longmapsto
 (\widetilde{\mathbf Q},\widetilde{\mathbf b})
 =(\mathbf Q\mathbf W,\ s\mathbf Q\mathbf h+\mathbf b)
\]
is a bijection of $SO(3)\times\mathbb R^3$ onto itself: its inverse is
$\mathbf Q=\widetilde{\mathbf Q}\mathbf W^\top$ and
$\mathbf b=\widetilde{\mathbf b}
 -s\widetilde{\mathbf Q}\mathbf W^\top\mathbf h$.
It preserves $s$, including $s=0$ if allowed.  The two optimisation domains
therefore generate exactly the same aligned trajectories, proving equality
of their minima, or of their positive-scale infima.
The orientations never enter either objective.

For frames $i,j$, put $\Delta\mathbf p_{ij}=\mathbf p_j-\mathbf p_i$.
Their displacement expressed in predicted camera $i$ changes from
\[
 \mathbf u_{ij}=\hat{\mathbf R}_i^\top\Delta\mathbf p_{ij}
 \quad\hbox{to}\quad
 \mathbf u'_{ij}=\hat{\mathbf R}_i^\top\mathbf W\Delta\mathbf p_{ij}.
\]
Thus it is unchanged exactly when
$(\mathbf W-\mathbf I)\Delta\mathbf p_{ij}=\mathbf0$.
For a fixed fitted scale $s$, the corresponding translational relative
pose error is the norm of
$s\mathbf u_{ij}-\mathbf R_i^\top(\mathbf g_j-\mathbf g_i)$,
where $\mathbf R_i$ is the true camera-to-world orientation.
Even if $\mathbf u_{ij}$ changes, this scalar norm need not change.

For relative reprojection, take a static point with source-camera coordinates
$\mathbf y_i$.  The predicted target-camera coordinates obtained by
backprojecting it in camera $i$ and transporting it to camera $j$ are
\[
 \mathbf y_j=\hat{\mathbf R}_j^\top
 \bigl(\hat{\mathbf R}_i\mathbf y_i+\mathbf p_i-\mathbf p_j\bigr).
\]
Keeping $\mathbf y_i$ and the orientations fixed while transforming the
predicted centres gives
\[
 \mathbf y'_j-\mathbf y_j
 =\hat{\mathbf R}_j^\top(\mathbf W-\mathbf I)
                  (\mathbf p_i-\mathbf p_j).
\]
Their image projections can consequently differ.  For example, for unit
intrinsics define $\pi(x,y,z)=(x/z,y/z)$ when $z>0$.
Take two frames with true and predicted centres
$\mathbf p_1=\mathbf g_1=(0,0,0)^\top$ and
$\mathbf p_2=\mathbf g_2=(1,0,0)^\top$, and let all orientations be
$\mathbf I$.  With
\[
 \mathbf W=\begin{pmatrix}0&-1&0\\1&0&0\\0&0&1\end{pmatrix},
 \qquad \mathbf h=\mathbf0,
\]
ATE is zero before and after the change, because the latter prediction can
be aligned by $s=1$, $\mathbf Q=\mathbf W^\top$, $\mathbf b=\mathbf0$.
The camera-frame displacement changes from $(1,0,0)^\top$ to
$(0,1,0)^\top$, and the translational relative pose error at scale one changes
from $0$ to $\sqrt2$.
For $\mathbf y_1=(0,0,2)^\top$, the target-camera coordinates are initially
$(-1,0,2)^\top$ and afterwards $(0,-1,2)^\top$, giving image points
$(-1/2,0)$ and $(0,-1/2)$.
This proves that invariant ATE does not imply invariant relative motion
or reprojection.

A common translation alone ($\mathbf W=\mathbf I$), or in this example a
rotation about the $x$-axis, on which both centres lie, leaves them unchanged.
\hfill$\square$

\refstepcounter{proofprop}
\paragraph{Proposition 2 (the constant-velocity line and the scale boundary).}
\label{prop:constant_velocity_ate}
\emph{Suppose $\mathbf p_i=\mathbf a+t_i\mathbf d$ with
$\mathbf d\ne\mathbf0$ and timestamps not all equal.  Then
\[
 \operatorname{ATE}(\mathbf P;\mathbf G)
 =\operatorname{ATE}_{>0}(\mathbf P;\mathbf G)
 =\operatorname{rms}\bigl(\operatorname{r}(\mathbf G)\bigr),
\]
which is \cref{eq:line} and is independent of the prediction's offset,
direction and nonzero speed.  Under the strictly positive-scale convention,
this value is attained if and only if the least-squares regression slope
of $\mathbf G$ on time is nonzero.}

\emph{Proof.}
An aligned prediction has the form
\[
 s\mathbf Q(\mathbf a+t_i\mathbf d)+\mathbf b
 =\mathbf a'+t_i\mathbf d',
 \qquad
 \mathbf a'=s\mathbf Q\mathbf a+\mathbf b,
 \quad \mathbf d'=s\mathbf Q\mathbf d.
\]
Every nonzero target slope $\mathbf d'$ is attainable: choose
$s=\|\mathbf d'\|/\|\mathbf d\|$ and a proper rotation taking
$\mathbf d/\|\mathbf d\|$ to $\mathbf d'/\|\mathbf d'\|$.
Such a rotation exists explicitly by completing each of these unit vectors
to a right-handed orthonormal basis and multiplying the second basis matrix
by the transpose of the first.  Any desired intercept $\mathbf a'$ is then
obtained with $\mathbf b=\mathbf a'-s\mathbf Q\mathbf a$.
The zero target slope is attainable precisely when $s=0$ is permitted.
In that case, choose $s=0$ and $\mathbf b=\mathbf a'$.
The nonnegative-scale domain therefore gives every constant-velocity
trajectory, whereas the positive-scale domain gives exactly those with
nonzero slope.

For completeness, put $\bar t=n^{-1}\sum_i t_i$,
$\tau_i=t_i-\bar t$, and $D_t=\sum_i\tau_i^2>0$.
The unique regression coefficients are
\[
 \mathbf d_*=
 \frac{\sum_i\tau_i(\mathbf g_i-\bar{\mathbf g})}{D_t},
 \qquad
 \mathbf a_*=\bar{\mathbf g}-\bar t\,\mathbf d_*.
\]
Writing $\mathbf e_i=\mathbf g_i-\mathbf a_*-t_i\mathbf d_*$,
these formulas give $\sum_i\mathbf e_i=\mathbf0$ and
$\sum_i\tau_i\mathbf e_i=\mathbf0$.
For any coefficients $\mathbf a',\mathbf d'$ we have
\[
 \mathbf g_i-\mathbf a'-t_i\mathbf d'
 =\mathbf e_i
  -\bigl(\mathbf a'+\bar t\,\mathbf d'-\bar{\mathbf g}\bigr)
  -\tau_i(\mathbf d'-\mathbf d_*).
\]
Squaring and summing, the cross terms vanish by the two identities just
given and by $\sum_i\tau_i=0$, yielding
\begin{align*}
 \frac1n\sum_i\|\mathbf g_i-\mathbf a'-t_i\mathbf d'\|^2
 &=\operatorname{rms}\bigl(\operatorname{r}(\mathbf G)\bigr)^2
   +\|\mathbf a'+\bar t\,\mathbf d'-\bar{\mathbf g}\|^2
   +\frac{D_t}{n}\|\mathbf d'-\mathbf d_*\|^2.
\end{align*}
The last two terms vanish at, and only at, the stated regression
coefficients.  This proves the nonnegative-scale minimum and the
positive-scale attainment assertion when $\mathbf d_*\ne\mathbf0$.
If $\mathbf d_*=\mathbf0$, any admissible positive-scale prediction has
$\mathbf d'\ne\mathbf0$ and hence strictly greater squared error.
Nevertheless, for any unit vector $\mathbf v$, choosing
$\mathbf d'=\varepsilon\mathbf v$ and
$\mathbf a'=\bar{\mathbf g}-\bar t\,\varepsilon\mathbf v$ gives squared
error equal to the regression residual plus $\varepsilon^2D_t/n$.
Letting $\varepsilon\downarrow0$ proves the same infimum and its
nonattainment. \hfill$\square$

If $\|\mathbf G_c\|_F>0$, the associated coefficient of determination is
$R^2_{\rm line}=1-\|\operatorname{r}(\mathbf G)\|_F^2/\|\mathbf G_c\|_F^2$.
It is undefined when the true centres are all equal, although the line's
nonnegative-scale ATE remains well defined and equals zero.

\refstepcounter{proofprop}
\paragraph{Proposition 3 (exact fixed-alignment straightening threshold).}
\label{prop:straightening_threshold}
\emph{Fix any alignment $A(\mathbf x)=s\mathbf Q\mathbf x+\mathbf b$
in the chosen admissible alignment class, and put
$\hat{\mathbf P}=A(\mathbf P)$ and $\mathbf E=\mathbf G-\hat{\mathbf P}$.
Use the RMS quantities of \cref{eq:snr},
\[
 S=\operatorname{rms}\bigl(\operatorname{r}(\mathbf G)\bigr),
 \qquad
 N=\operatorname{rms}\bigl(\operatorname{r}(\mathbf E)\bigr).
\]
For $N>0$ define
\[
 c=\frac{\langle\operatorname{r}(\mathbf G),
                   \operatorname{r}(\mathbf E)\rangle_F}{nSN}
 \quad\text{if }S>0,
 \qquad c=0\quad\text{if }S=0.
\]
For $\alpha\in(0,1]$, set $\beta=(1-\alpha)/\alpha$ and form the raw,
unaligned straightening of \cref{eq:straighten},
$\mathbf P_\alpha=\operatorname{cv}(\mathbf P)
 +(1-\alpha)\operatorname{r}(\mathbf P)$.
If $N>0$, the error under the same fixed map $A$ strictly decreases if and only if
\begin{equation}
 \frac SN< -\beta c+
       \sqrt{\beta^2c^2+\frac{2-\alpha}{\alpha}}.
 \label{eq:proof_straightening_threshold}
\end{equation}
If $N=0$, the fixed-alignment squared error increases by $\alpha^2S^2$,
so it never strictly decreases.  Finally, if $A$ is an attained minimiser
for the original prediction, the condition in \cref{eq:proof_straightening_threshold}
is sufficient for a strict decrease of ATE after refitting the alignment.}

\emph{Proof.}
We first establish the commutation needed to use the same map on the raw
straightened trajectory.  For any fixed affine map
$F(\mathbf x)=\mathbf L\mathbf x+\mathbf b$,
with arbitrary $\mathbf L\in\mathbb R^{3\times3}$, not necessarily
invertible, its array action is
$F(\mathbf X)=\mathbf X\mathbf L^\top+\mathbf1\mathbf b^\top$.
Using $\mathbf H\mathbf1=\mathbf1$ gives
\begin{align*}
 \operatorname{cv}\bigl(F(\mathbf X)\bigr)
 &=\mathbf H\mathbf X\mathbf L^\top+\mathbf1\mathbf b^\top
   =F\bigl(\operatorname{cv}(\mathbf X)\bigr),\\
 \operatorname{r}\bigl(F(\mathbf X)\bigr)
 &=(\mathbf I_n-\mathbf H)\mathbf X\mathbf L^\top
   =\operatorname{r}(\mathbf X)\mathbf L^\top.
\end{align*}
Consequently,
\begin{align*}
 F\bigl(\operatorname{cv}(\mathbf X)+(1-\alpha)\operatorname{r}(\mathbf X)\bigr)
 &=\mathbf H\mathbf X\mathbf L^\top+\mathbf1\mathbf b^\top
        +(1-\alpha)(\mathbf I_n-\mathbf H)\mathbf X\mathbf L^\top\\
 &=\operatorname{cv}\bigl(F(\mathbf X)\bigr)
       +(1-\alpha)\operatorname{r}\bigl(F(\mathbf X)\bigr).
\end{align*}
This proves commutation for every fixed affine map and, in particular,
for the old alignment $A$, including a zero-scale alignment.
There is no assertion that a newly fitted alignment equals the old one.

Since $\operatorname{r}(\hat{\mathbf P})
=\operatorname{r}(\mathbf G)-\operatorname{r}(\mathbf E)$, the new error
under $A$ is exactly
\begin{align*}
 \mathbf E_\alpha
 &=\mathbf G-A(\mathbf P_\alpha)\\
 &=\operatorname{cv}(\mathbf G)-\operatorname{cv}(\hat{\mathbf P})
      +\operatorname{r}(\mathbf G)-(1-\alpha)\operatorname{r}(\hat{\mathbf P})\\
 &=\operatorname{cv}(\mathbf E)
      +\alpha\operatorname{r}(\mathbf G)
      +(1-\alpha)\operatorname{r}(\mathbf E).
\end{align*}
The constant-velocity part is orthogonal to both residual terms.  Therefore,
writing $e_0=\operatorname{rms}(\mathbf E)$ and
$e_\alpha=\operatorname{rms}(\mathbf E_\alpha)$, we obtain
\begin{align*}
 e_\alpha^2-e_0^2
 &=\frac1n\left(
   \|\alpha\operatorname{r}(\mathbf G)+(1-\alpha)\operatorname{r}(\mathbf E)\|_F^2
    -\|\operatorname{r}(\mathbf E)\|_F^2\right)\\
 &=\alpha^2S^2
   +\frac{2\alpha(1-\alpha)}n
        \langle\operatorname{r}(\mathbf G),\operatorname{r}(\mathbf E)\rangle_F
   -\alpha(2-\alpha)N^2.\tag{*}
\end{align*}
When $N>0,S>0$, the definition of $c$ changes the middle term to
$2\alpha(1-\alpha)cSN$.  The same expression remains valid when $S=0$
because then $\operatorname{r}(\mathbf G)=\mathbf0$ and the inner product
vanishes.  For nonzero residuals, $|c|\le1$: apply
$0\le\|\mathbf U-\langle\mathbf U,\mathbf V\rangle_F
\mathbf V/\|\mathbf V\|_F^2\|_F^2
=\|\mathbf U\|_F^2-\langle\mathbf U,\mathbf V\rangle_F^2/\|\mathbf V\|_F^2$
to $\mathbf U=\operatorname{r}(\mathbf G)$ and
$\mathbf V=\operatorname{r}(\mathbf E)$.

For $N>0$, dividing (*) by $\alpha^2N^2>0$ and setting $x=S/N\ge0$
shows that $e_\alpha<e_0$ is equivalent to
\[
 x^2+2\beta cx-\frac{2-\alpha}{\alpha}<0.
\]
The roots of this quadratic are
$x_\pm=-\beta c\pm\sqrt{\beta^2c^2+(2-\alpha)/\alpha}$.
Because $(2-\alpha)/\alpha>0$, the square root is strictly greater than
$|\beta c|$, so $x_-<0<x_+$.  The quadratic factors as
$(x-x_-)(x-x_+)$. For $x\ge0$ its first factor is positive, and the product
is negative exactly when $x<x_+$, proving the stated equivalence.
Equality at $x=x_+$ means equality of the fixed-alignment errors.

The degenerate cases also follow directly from (*).
If $S=0<N$, then
$e_\alpha^2-e_0^2=-\alpha(2-\alpha)N^2<0$.
The threshold with $c=0$ agrees with this conclusion.
If $N=0$, then $\operatorname{r}(\mathbf E)=\mathbf0$ and (*) is
$\alpha^2S^2$, strictly positive for $S>0$ and zero for $S=0$.
In this case neither $S/N$ nor a residual correlation needs to be defined.
If $c=0$ and $N>0$, the threshold reduces to
$S/N<\sqrt{(2-\alpha)/\alpha}$. At $\alpha=1$ it is $S<N$, and at
$\alpha=1/2$ it is $S/N<\sqrt3$.

For the refitted score, now assume additionally that $A$ minimises the
original objective, so $e_0=\operatorname{ATE}(\mathbf P;\mathbf G)$.
The same map $A$ is admissible when scoring $\mathbf P_\alpha$.
If the fixed-alignment condition holds, minimisation over all new
alignments gives
\[
 \operatorname{ATE}(\mathbf P_\alpha;\mathbf G)
 \le \operatorname{rms}\bigl(\mathbf G-A(\mathbf P_\alpha)\bigr)
 = e_\alpha<e_0=\operatorname{ATE}(\mathbf P;\mathbf G).
\]
The same argument applies to the positive-scale infimum when the original
infimum is attained by the chosen positive-scale $A$. It also applies
numerically whenever the stated old map is genuinely optimal.
For an arbitrary nonoptimal old map the final equality is unavailable,
so decreasing its fixed error alone does not prove a decrease in ATE.
When the threshold fails, refitting may still decrease ATE, and the
fixed-alignment result gives no converse for that refitted score.
\hfill$\square$

\paragraph{What position alignment does not remove.}
It is not true in general that only the non-constant-velocity component
can contribute to aligned error.  The exact decomposition is
\[
 \operatorname{rms}(\mathbf E)^2
 =\operatorname{rms}\bigl(\operatorname{cv}(\mathbf E)\bigr)^2
  +\operatorname{rms}\bigl(\operatorname{r}(\mathbf E)\bigr)^2,
\]
and the first term need not vanish at an optimal position alignment.
For an explicit example let $t=(-1,0,1)$,
$\mathbf g_i=(t_i,0,0)^\top$, and
$\mathbf p_i=(t_i,t_i^2-2/3,0)^\top$.
Both trajectories are centred, $\|\mathbf G\|_F^2=2$,
$\|\mathbf P\|_F^2=8/3$, and
$\langle\mathbf G,\mathbf P\mathbf Q^\top\rangle_F=2Q_{11}\le2$.
Hence their squared alignment error is
$2+(8/3)s^2-4sQ_{11}$, minimised over $s\ge0,\mathbf Q\in SO(3)$
by $s=3/4$ and any $\mathbf Q$ with $Q_{11}=1$, e.g.\ $\mathbf Q=\mathbf I$.
For every such minimiser,
$\operatorname{cv}(\mathbf E)_i=(t_i/4,0,0)^\top$, whose squared RMS is
$1/24>0$. The residual-error squared RMS is $1/8$ and the total is $1/6$.
Position alignment removes a common spatial rotation, scale and offset,
but it does not independently fit the prediction's temporal trend and
its residual.  Proposition~2 gives a residual-only expression for the
constant-velocity baseline, not for arbitrary predictions.

\refstepcounter{proofprop}
\paragraph{Proposition 4 (orientation anchor: uniqueness, stability and position error).}
\label{prop:pose_anchored_properties}
Let $n\ge1$, let $\mathbf{p}_i,\mathbf{g}_i\in\mathbb R^3$ be predicted and true camera centres, and let $\hat{\mathbf R}_i,\mathbf R_i\in SO(3)$ be their camera-to-world rotations.
For matrices, write $\langle\mathbf A,\mathbf B\rangle_F=\operatorname{tr}(\mathbf A^{\top}\mathbf B)$ and $\|\mathbf A\|_F^2=\langle\mathbf A,\mathbf A\rangle_F$.
Here a product $\mathbf Q\mathbf X$ denotes a pointwise rotation. In the row-array convention, it is $\mathbf X\mathbf Q^{\top}$.
For trajectories, write $\langle\mathbf X,\mathbf Y\rangle=\sum_i\mathbf x_i^{\top}\mathbf y_i$ and $\|\mathbf X\|^2=\sum_i\|\mathbf x_i\|^2$, with rotations acting on each point as a column vector.
Define
\[
 \mathbf M=\sum_{i=1}^n\mathbf R_i\hat{\mathbf R}_i^{\top},
 \qquad
 \mathcal A(\mathbf M)=\arg\min_{\mathbf Q\in SO(3)}
       \sum_{i=1}^n\|\mathbf R_i-\mathbf Q\hat{\mathbf R}_i\|_F^2.
\]
Choose any ordered singular value decomposition
$\mathbf M=\mathbf U\operatorname{diag}(\sigma_1,\sigma_2,\sigma_3)\mathbf V^{\top}$,
where $\sigma_1\ge\sigma_2\ge\sigma_3\ge0$ and $\mathbf U,\mathbf V$ are orthogonal, and put
\[
 d=\det(\mathbf U\mathbf V^{\top})\in\{-1,1\},\quad
 \mathbf D=\operatorname{diag}(1,1,d),\quad
 \mathbf Q^*=\mathbf U\mathbf D\mathbf V^{\top},\quad
 \gamma=\sigma_2+d\sigma_3.
\]
For a specified anchor $\mathbf Q$, define the position score
\[
 e_{\mathbf Q}(\mathbf P,\mathbf G)
 =\min_{s\ge0,\,\mathbf b\in\mathbb R^3}
   \left(\frac1n\sum_i
      \|\mathbf g_i-s\mathbf Q\mathbf p_i-\mathbf b\|^2\right)^{1/2}.
\]
Thus the PA-ATE is $e_{\mathbf Q^*}$ when the anchor is unique.
When it is not unique, a reported scalar additionally requires a rule for selecting or rejecting an anchor.
The constraint $s\ge0$ includes the collapsed map $s=0$, which is not an invertible similarity transform.

\begin{enumerate}
\item[(a)] \emph{Optimality and uniqueness.}
$\mathbf Q^*\in\mathcal A(\mathbf M)$, and the anchor is unique if and only if $\gamma>0$.
The number $\gamma$ is independent of the chosen SVD, including when $\mathbf M$ is singular.
For the above sum of $n$ rotations, $0\le\gamma/n\le2$.

\item[(b)] \emph{A global stability bound.}
For every $\mathbf Q\in SO(3)$,
\begin{equation}
 \langle\mathbf Q^*,\mathbf M\rangle_F
 -\langle\mathbf Q,\mathbf M\rangle_F
 \ \ge\ \frac{\gamma}{4}\|\mathbf Q-\mathbf Q^*\|_F^2.
 \label{eq:anchor_trace_gap}
\end{equation}
If $\gamma>0$, $\boldsymbol\Delta\in\mathbb R^{3\times3}$, and
$\widetilde{\mathbf Q}$ is \emph{any} maximiser of
$\langle\mathbf Q,\mathbf M+\boldsymbol\Delta\rangle_F$ over $SO(3)$, then
\begin{equation}
 \|\widetilde{\mathbf Q}-\mathbf Q^*\|_F
 \ \le\ \min\!\left\{2\sqrt2,\,
                 \frac{4\|\boldsymbol\Delta\|_F}{\gamma}\right\}.
 \label{eq:anchor_stability_bound}
\end{equation}
No uniqueness assumption on the perturbed maximiser is needed.
Equivalently, the bound uses $\|\boldsymbol\Delta/n\|_F$ divided by $\gamma/n$.

\item[(c)] \emph{Coordinate equivariance.}
Let $\mathbf W_p,\mathbf W_g\in SO(3)$, $a>0$, and
$\mathbf b_p,\mathbf b_g\in\mathbb R^3$.
Change the prediction and truth by
\[
 \begin{aligned}
 \mathbf p_i'&=a\mathbf W_p\mathbf p_i+\mathbf b_p,
 &\hat{\mathbf R}_i'&=\mathbf W_p\hat{\mathbf R}_i,\\
 \mathbf g_i'&=\mathbf W_g\mathbf g_i+\mathbf b_g,
 &\mathbf R_i'&=\mathbf W_g\mathbf R_i.
 \end{aligned}
\]
If the original anchor is unique, the new one is
$\mathbf Q^{*\prime}=\mathbf W_g\mathbf Q^*\mathbf W_p^{\top}$,
and the PA-ATE is unchanged.
Without uniqueness, the \emph{set} of optimal anchors obeys the same equivariance, but an arbitrary SVD tie-breaking rule need not select corresponding anchors or preserve the scalar score.

\item[(d)] \emph{Position fit and zero scale.}
Write $\bar{\mathbf p}=n^{-1}\sum_i\mathbf p_i$,
$\bar{\mathbf g}=n^{-1}\sum_i\mathbf g_i$,
$\mathbf P_c=(\mathbf p_i-\bar{\mathbf p})_i$,
$\mathbf G_c=(\mathbf g_i-\bar{\mathbf g})_i$, and
$r_G=\|\mathbf G_c\|/\sqrt n$ (the RMS radius of the true centres; not to be confused with the residual operator $\operatorname r(\cdot)$).
For a fixed anchor $\mathbf Q$, if $v=\|\mathbf P_c\|^2>0$, set
$h=\langle\mathbf Q\mathbf P_c,\mathbf G_c\rangle$ and
$h_+=\max\{h,0\}$.
The unique optimal scale and its translation are
\begin{equation}
 s^*=\frac{h_+}{v},\qquad
 \mathbf b^*=\bar{\mathbf g}-s^*\mathbf Q\bar{\mathbf p},\qquad
 e_{\mathbf Q}^2
 =\frac1n\left(\|\mathbf G_c\|^2-\frac{h_+^2}{v}\right).
 \label{eq:anchored_position_closed_form}
\end{equation}
In particular, $h\le0$ gives $s^*=0$ and $e_{\mathbf Q}=r_G$:
all aligned centres collapse to $\bar{\mathbf g}$.
This is the same \emph{position score} as a static predicted camera. It does not equate the predicted orientations or their companion rotation errors.
If $v=0$, every $s\ge0$ can be absorbed in the translation, and $e_{\mathbf Q}=r_G$. One may choose $s=0$ by convention.
For every input, $0\le e_{\mathbf Q}\le r_G$.

\item[(e)] \emph{Rotating true positions while keeping true orientations.}
Let $\hat{\mathbf R}_i=\mathbf R_i$ and
$\mathbf p_i=\bar{\mathbf g}+\mathbf Z(\mathbf g_i-\bar{\mathbf g})$
for $\mathbf Z\in SO(3)$.
The anchor is uniquely $\mathbf I$.
If $r_G>0$, define
\[
 \rho=\frac{\langle\mathbf Z\mathbf G_c,\mathbf G_c\rangle}
                  {\|\mathbf G_c\|^2}\in[-1,1].
\]
Then
\begin{equation}
 s^*=\max\{\rho,0\},\qquad
 e_{\mathbf I}=\begin{cases}
      r_G\sqrt{1-\rho^2},&\rho>0,\\
      r_G,&\rho\le0.
     \end{cases}
 \label{eq:anchored_rotated_centres}
\end{equation}
At $\rho=0$, the optimal scale is zero and the two expressions agree.
If $r_G=0$, $\rho$ is undefined and is not used: both trajectories are constant, and the score is zero for every $\mathbf Z$.
Position-aligned ATE is zero in all these cases, since the inverse rotation recovers the true centres.

More specifically, suppose $\mathbf Z$ has principal angle
$\theta\in[0,\pi]$ and unit rotation axis $\mathbf u$, and $r_G>0$.
Define the fraction of centred position energy parallel to that axis by
\[
 \eta=\frac{\sum_i\big(\mathbf u^{\top}
                    (\mathbf g_i-\bar{\mathbf g})\big)^2}
                 {\|\mathbf G_c\|^2}\in[0,1].
\]
Then
\begin{equation}
 \rho=\eta+(1-\eta)\cos\theta.
 \label{eq:anchored_axis_energy}
\end{equation}
If $\eta=0$, the centred trajectory lies in the plane perpendicular to the axis, and the score is $r_G\sin\theta$ on $[0,\pi/2]$ and $r_G$ on $[\pi/2,\pi]$.
In general, for $0\le\eta<1/2$ the zero-scale threshold is
$\theta_{\rm sat}=\arccos[-\eta/(1-\eta)]$, and the score equals $r_G$ for $\theta\in[\theta_{\rm sat},\pi]$.
For $\eta=1/2$, zero scale occurs only at $\theta=\pi$. For $\eta>1/2$, it never occurs on this interval.
In particular, $\eta=1$ gives a zero score for every angle because the rotation leaves every centred position unchanged.

\item[(f)] \emph{The score is Lipschitz in the anchor.}
For any prediction and any $\mathbf Q_1,\mathbf Q_2\in SO(3)$ whose relative rotation $\mathbf Q_1^{\top}\mathbf Q_2$ has angle $\theta\in[0,\pi]$,
\begin{equation}
 \big|e_{\mathbf Q_1}-e_{\mathbf Q_2}\big|\ \le\ 2\sin(\theta/2)\,r_G .
 \label{eq:anchor_lipschitz}
\end{equation}
In particular, if the predicted positions are exact, an anchor that is off by $\theta$ gives a PA-ATE of at most $2\sin(\theta/2)\,r_G$.
\end{enumerate}

\emph{Proof.}
\emph{(a) Reduction to a trace maximisation.}
Since rotation matrices have squared Frobenius norm $3$,
\[
 \sum_i\|\mathbf R_i-\mathbf Q\hat{\mathbf R}_i\|_F^2
   =6n-2\operatorname{tr}(\mathbf Q^{\top}\mathbf M).
\]
Put $\mathbf U'=\mathbf U\mathbf D$ and
$\boldsymbol\Lambda=\operatorname{diag}(\sigma_1,\sigma_2,d\sigma_3)$.
Then $\mathbf M=\mathbf U'\boldsymbol\Lambda\mathbf V^{\top}$,
$\det(\mathbf U'\mathbf V^{\top})=1$, and every
$\mathbf Q\in SO(3)$ has the unique representation
$\mathbf Q=\mathbf U'\boldsymbol\Omega\mathbf V^{\top}$ with
$\boldsymbol\Omega\in SO(3)$.
In these coordinates $\mathbf Q^*$ corresponds to $\boldsymbol\Omega=\mathbf I$ and
\[
 \operatorname{tr}(\mathbf Q^{\top}\mathbf M)
   =\operatorname{tr}(\boldsymbol\Omega^{\top}\boldsymbol\Lambda).
\]
To justify the axis--angle representation used next, a real orthogonal
$3\times3$ matrix has eigenvalues of modulus one, with nonreal eigenvalues
in conjugate pairs.  Since $\det\boldsymbol\Omega=1$, it has a real eigenvalue $+1$.
Choose a corresponding unit eigenvector $\mathbf w$.
Its orthogonal complement is invariant, and the restriction there is a
$2\times2$ orthogonal matrix of determinant one, hence has the form
$\left(\begin{smallmatrix}\cos\varphi&-\sin\varphi\\
\sin\varphi&\cos\varphi\end{smallmatrix}\right)$ in an oriented basis.
Choosing the sign of the axis permits $\varphi\in[0,\pi]$.
Thus every $\boldsymbol\Omega\in SO(3)$ is a rotation through an angle
$\varphi\in[0,\pi]$ about some unit axis $\mathbf w$ (the axis is arbitrary when $\boldsymbol\Omega=\mathbf I$).
Its diagonal entries are
$\Omega_{kk}=\cos\varphi+(1-\cos\varphi)w_k^2$.
Consequently the trace gap is exactly
\begin{align}
 \operatorname{tr}(\boldsymbol\Lambda)
 -\operatorname{tr}(\boldsymbol\Omega^{\top}\boldsymbol\Lambda)
 &= (1-\cos\varphi)
       \left(\operatorname{tr}(\boldsymbol\Lambda)
                    -\mathbf w^{\top}\boldsymbol\Lambda\mathbf w\right)
 \nonumber\\
 &\ge (1-\cos\varphi)(\sigma_2+d\sigma_3).
 \label{eq:anchor_axis_trace_gap}
\end{align}
Here the largest diagonal entry of $\boldsymbol\Lambda$ is $\sigma_1$, and
$\gamma=\sigma_2+d\sigma_3\ge0$ by the ordering of the singular values.
Thus $\mathbf Q^*$ is optimal.
If $\gamma>0$, every $\varphi>0$ has a strictly positive gap, so it is the unique optimum.
If $\gamma=0$, choose $\mathbf w=(1,0,0)^{\top}$. Every nonzero rotation angle about this axis attains gap zero, giving distinct optima.
This proves the equivalence, including all zero-singular-value cases.

If $\mathbf M$ is nonsingular, $d=\operatorname{sign}\det\mathbf M$ is independent of its SVD.
If it is singular, $\sigma_3=0$, so even if null-space choices change $d$, they do not change $\gamma=\sigma_2$.
In particular, rank two gives a unique anchor, whereas rank zero or one does not.
For a nonsingular negative-determinant matrix the only nonunique case is $\sigma_2=\sigma_3$.
Whenever the optimum is unique, all SVD choices therefore give the same $\mathbf Q^*$.
Finally, each summand of $\mathbf M$ is a rotation and has operator norm one, so $\sigma_k\le n$ for every $k$. Hence $0\le\gamma\le2n$.
The value $2n$ is attained when all relative rotations $\mathbf R_i\hat{\mathbf R}_i^{\top}$ coincide.

\emph{(b) Stability.}
For a three-dimensional rotation through $\varphi$,
\[
 \|\boldsymbol\Omega-\mathbf I\|_F^2
   =6-2\operatorname{tr}(\boldsymbol\Omega)
   =4(1-\cos\varphi).
\]
Orthogonal invariance of the Frobenius norm gives
$\|\mathbf Q-\mathbf Q^*\|_F=\|\boldsymbol\Omega-\mathbf I\|_F$.
Substitution in \cref{eq:anchor_axis_trace_gap} proves
\cref{eq:anchor_trace_gap}.
Existence of a maximiser of the perturbed objective follows from continuity on compact $SO(3)$.
Its optimality gives
\[
 \langle\widetilde{\mathbf Q},\mathbf M+\boldsymbol\Delta\rangle_F
 \ge \langle\mathbf Q^*,\mathbf M+\boldsymbol\Delta\rangle_F.
\]
Therefore, with $\delta=\|\widetilde{\mathbf Q}-\mathbf Q^*\|_F$,
\[
 \frac{\gamma}{4}\delta^2
 \le \langle\mathbf Q^*-\widetilde{\mathbf Q},\mathbf M\rangle_F
 \le \langle\widetilde{\mathbf Q}-\mathbf Q^*,\boldsymbol\Delta\rangle_F
 \le \delta\|\boldsymbol\Delta\|_F.
\]
For $\delta=0$ the bound is immediate. Otherwise, division by
$\gamma\delta/4>0$ gives $\delta\le4\|\boldsymbol\Delta\|_F/\gamma$.
The same rotation-distance identity gives the universal bound
$\delta\le2\sqrt2$.

\emph{(c) Coordinate changes.}
The new orientation matrix is
$\mathbf M'=\mathbf W_g\mathbf M\mathbf W_p^{\top}$.
The bijection
$\mathbf Q\mapsto\mathbf Q'=\mathbf W_g\mathbf Q\mathbf W_p^{\top}$
preserves the trace objective, which proves equivariance of its optimiser set and, when unique, its optimiser.
For any original scale and translation $(s,\mathbf b)$, set
\[
 s'=s/a,\qquad
 \mathbf b'=\mathbf W_g\mathbf b+\mathbf b_g
                   -(s/a)\mathbf Q'\mathbf b_p.
\]
Direct substitution gives
\[
 \mathbf g_i'-s'\mathbf Q'\mathbf p_i'-\mathbf b'
   =\mathbf W_g(\mathbf g_i-s\mathbf Q\mathbf p_i-\mathbf b).
\]
The parameter map is bijective for $a>0$, and $\mathbf W_g$ preserves norms, so the two position minima agree for corresponding anchors.
If the optimiser is not unique, selecting noncorresponding anchors does not have this guarantee.
For example, take true orientations $\mathbf I$, predicted orientations
$\{\mathbf I,\operatorname{diag}(1,-1,-1),\operatorname{diag}(-1,1,-1),\operatorname{diag}(-1,-1,1)\}$,
and identical true and predicted centres at the four vertices
$(1,1,1),(1,-1,-1),(-1,1,-1),(-1,-1,1)$.
Then $\mathbf M=\mathbf0$ and every anchor is optimal, but the identity anchor gives zero position error whereas a $90^\circ$ anchor about the third axis gives $\sqrt{8/3}$.
Thus the position scalar is genuinely ambiguous, not just its matrix representation.

\emph{(d) Fitting scale and translation.}
For fixed $s$ and $\mathbf Q$, completing the square in $\mathbf b$ gives
\[
 \sum_i\|\mathbf g_i-s\mathbf Q\mathbf p_i-\mathbf b\|^2
 =\|\mathbf G_c-s\mathbf Q\mathbf P_c\|^2
  +n\|\bar{\mathbf g}-s\mathbf Q\bar{\mathbf p}-\mathbf b\|^2.
\]
The second term is minimised uniquely at
$\mathbf b=\bar{\mathbf g}-s\mathbf Q\bar{\mathbf p}$.
The first is $\|\mathbf G_c\|^2-2sh+s^2v$.
For $v>0$ this is a strictly convex quadratic, whose constrained minimiser is $s^*=h_+/v$, and substitution proves \cref{eq:anchored_position_closed_form}.
For $v=0$ the centred prediction is zero and this expression is $\|\mathbf G_c\|^2$ independently of $s$.
The upper bound $e_{\mathbf Q}\le r_G$ follows by choosing $s=0$, and the lower bound follows because the objective is a squared norm.
If the true trajectory is static ($r_G=0$), the score is consequently zero for every predicted position trajectory under the allowed zero scale. This is a scale-identifiability boundary, not evidence of accurate predicted motion.

\emph{(e) Rotated positions.}
True predicted orientations give $\mathbf M=n\mathbf I$, so (a) gives the unique anchor $\mathbf I$.
The centred prediction is $\mathbf Z\mathbf G_c$.
If $r_G>0$, its squared norm is $\|\mathbf G_c\|^2>0$, and its inner product with $\mathbf G_c$ is $\rho\|\mathbf G_c\|^2$.
The bound $|\rho|\le1$ follows from Cauchy--Schwarz.
Part (d) now gives \cref{eq:anchored_rotated_centres}, including $\rho=0$.
If $r_G=0$, both centred trajectories vanish and (d) gives zero directly.
For standard position alignment, the transformation
$\mathbf p\mapsto\mathbf Z^{\top}(\mathbf p-\bar{\mathbf g})+\bar{\mathbf g}$
recovers the true centres exactly, proving its zero ATE without any assumption about $r_G$.

Finally, decompose each centred true position as
$\mathbf g_{c,i}=\mathbf g_{\parallel,i}+\mathbf g_{\perp,i}$,
where $\mathbf g_{\parallel,i}=(\mathbf u^{\top}\mathbf g_{c,i})\mathbf u$.
A rotation about $\mathbf u$ fixes the parallel component and rotates the perpendicular component through $\theta$, and orthogonality of the two components gives
\[
 \mathbf g_{c,i}^{\top}\mathbf Z\mathbf g_{c,i}
 =\|\mathbf g_{\parallel,i}\|^2
     +\cos\theta\,\|\mathbf g_{\perp,i}\|^2.
\]
Sum and divide by $\|\mathbf G_c\|^2$ to obtain
\cref{eq:anchored_axis_energy}.
Solving $\eta+(1-\eta)\cos\theta\le0$ on $[0,\pi]$
gives the stated saturation cases.
Thus a $90^\circ$ onset of saturation is exact for perpendicular planar motion, not a general consequence of the metric.

\emph{(f) Lipschitz bound.}
By (d), $e_{\mathbf Q}=\min_{s\ge0}\|\mathbf G_c-s\mathbf Q\mathbf P_c\|/\sqrt n$. If $v=0$, both scores equal $r_G$. Otherwise let $s_1$ be the optimal scale for $\mathbf Q_1$. Then
\[
 \sqrt n\,e_{\mathbf Q_2}\le\|\mathbf G_c-s_1\mathbf Q_2\mathbf P_c\|
 \le\|\mathbf G_c-s_1\mathbf Q_1\mathbf P_c\|+s_1\|(\mathbf Q_1-\mathbf Q_2)\mathbf P_c\|
 \le\sqrt n\,e_{\mathbf Q_1}+\|\mathbf Q_1-\mathbf Q_2\|_2\,s_1\|\mathbf P_c\|.
\]
Since $s_1=h_+/v$ and $|h|\le\|\mathbf P_c\|\,\|\mathbf G_c\|$ by Cauchy--Schwarz, $s_1\|\mathbf P_c\|\le\|\mathbf G_c\|=\sqrt n\,r_G$.
The spectral norm $\|\mathbf Q_1-\mathbf Q_2\|_2=\|\mathbf I-\mathbf Q_1^{\top}\mathbf Q_2\|_2$ equals $|1-e^{\mathrm i\theta}|=2\sin(\theta/2)$, the largest modulus among the eigenvalues $0$ and $1-e^{\pm\mathrm i\theta}$ of the normal matrix $\mathbf I-\mathbf Q_1^{\top}\mathbf Q_2$.
Exchanging $\mathbf Q_1$ and $\mathbf Q_2$ gives \cref{eq:anchor_lipschitz}. For exact positions, $e_{\mathbf I}=0$ with $s=1$, and the second statement follows with $\mathbf Q_1=\mathbf I$.
\hfill$\square$

\refstepcounter{proofprop}
\paragraph{Proposition 5 (PA-ATE is the relative position error over all frame pairs).}
\label{prop:allpairs}
\emph{In the setting of Proposition~\ref{prop:pose_anchored_properties}, let $s^*\ge0$ and $\mathbf b^*$ be the scale and translation of Eq.~(\mref{eq:anchor_fit}) for the anchor $\mathbf Q^*$, and $\mathbf e_i=\mathbf g_i-s^*\mathbf Q^*\mathbf p_i-\mathbf b^*$, so that PA-ATE $=\operatorname{rms}(\mathbf E)$. For every ordered frame pair $(i,j)$ let
\[
 \boldsymbol\varepsilon_{ij}=\mathbf R_i^{\top}(\mathbf g_j-\mathbf g_i)-s^*\hat{\mathbf R}_i^{\top}(\mathbf p_j-\mathbf p_i)
\]
be the translational relative pose error of the pair, the error of the displacement to camera $j$ as camera $i$ sees it, with the anchored scale, and
\[
 \mathrm{RPE}_{\mathrm{all}}=\Big(\frac1{2n^2}\sum_{i,j=1}^n\|\boldsymbol\varepsilon_{ij}\|^2\Big)^{1/2}.
\]
Let $\theta_i$ be the angle of $\mathbf R_i^{\top}\mathbf Q^*\hat{\mathbf R}_i$, the rotation error of frame $i$ under the anchor, and $\theta_{\max}=\max_i\theta_i$. Then
\begin{equation}
 \bigl|\mathrm{RPE}_{\mathrm{all}}-\text{PA-ATE}\bigr|\le2\sin(\theta_{\max}/2)\,s^*\operatorname{rms}(\mathbf P_c)\le2\sin(\theta_{\max}/2)\,r_G.
 \label{eq:allpairs_bound}
\end{equation}
In particular $\mathrm{RPE}_{\mathrm{all}}=\text{PA-ATE}$ whenever every anchored orientation $\mathbf Q^*\hat{\mathbf R}_i$ equals the true $\mathbf R_i$. ATE satisfies the analogous identity only with the displacements expressed in the world frame.}

\emph{Proof.}
Write $\boldsymbol\Delta_i=\mathbf R_i^{\top}\mathbf Q^*\hat{\mathbf R}_i\in SO(3)$, so that $\hat{\mathbf R}_i^{\top}=\boldsymbol\Delta_i^{\top}\mathbf R_i^{\top}\mathbf Q^*$, and $\mathbf w_{ij}=s^*\mathbf R_i^{\top}\mathbf Q^*(\mathbf p_j-\mathbf p_i)$. Since $\mathbf g_j-\mathbf g_i=s^*\mathbf Q^*(\mathbf p_j-\mathbf p_i)+\mathbf e_j-\mathbf e_i$,
\[
 \boldsymbol\varepsilon_{ij}=\mathbf a_{ij}+\mathbf c_{ij},\qquad
 \mathbf a_{ij}=\mathbf R_i^{\top}(\mathbf e_j-\mathbf e_i),\qquad
 \mathbf c_{ij}=(\mathbf I-\boldsymbol\Delta_i^{\top})\,\mathbf w_{ij}.
\]
Because $\mathbf R_i$ is orthogonal, $\|\mathbf a_{ij}\|=\|\mathbf e_j-\mathbf e_i\|$. The translation $\mathbf b^*$ is an unconstrained least-squares fit, so $\bar{\mathbf e}=\mathbf 0$, and
\[
 \frac1{n^2}\sum_{i,j}\|\mathbf e_j-\mathbf e_i\|^2=\frac2n\sum_i\|\mathbf e_i\|^2-2\|\bar{\mathbf e}\|^2=2\,\text{PA-ATE}^2 .
\]
Treat $\{\boldsymbol\varepsilon_{ij}\}$, $\{\mathbf a_{ij}\}$ and $\{\mathbf c_{ij}\}$ as vectors in $\mathbb R^{3n^2}$ with the norm $(\frac1{2n^2}\sum_{i,j}\|\cdot\|^2)^{1/2}$. Then the norm of $\{\mathbf a_{ij}\}$ is PA-ATE, and the triangle inequality gives $|\mathrm{RPE}_{\mathrm{all}}-\text{PA-ATE}|$ at most the norm of $\{\mathbf c_{ij}\}$. As in Proposition~\ref{prop:pose_anchored_properties}(f), $\|\mathbf I-\boldsymbol\Delta_i^{\top}\|_2=2\sin(\theta_i/2)\le2\sin(\theta_{\max}/2)$, and $\|\mathbf w_{ij}\|=s^*\|\mathbf p_j-\mathbf p_i\|$, so
\[
 \frac1{2n^2}\sum_{i,j}\|\mathbf c_{ij}\|^2\le\bigl(2\sin(\theta_{\max}/2)\bigr)^2\frac{s^{*2}}{2n^2}\sum_{i,j}\|\mathbf p_j-\mathbf p_i\|^2=\bigl(2\sin(\theta_{\max}/2)\,s^*\operatorname{rms}(\mathbf P_c)\bigr)^2,
\]
using $\frac1{n^2}\sum_{i,j}\|\mathbf p_j-\mathbf p_i\|^2=2\operatorname{rms}(\mathbf P_c)^2$ as above. This is the first inequality of \cref{eq:allpairs_bound}. The second follows from $s^*\|\mathbf P_c\|\le\|\mathbf G_c\|$, shown in the proof of Proposition~\ref{prop:pose_anchored_properties}(f). If every $\boldsymbol\Delta_i=\mathbf I$, all $\mathbf c_{ij}$ vanish and the identity is exact.
For ATE, with its alignment $(s,\mathbf Q,\mathbf b)$ and residuals $\mathbf e_i$ of zero mean, the same computation gives $\frac1{2n^2}\sum_{i,j}\|(\mathbf g_j-\mathbf g_i)-s\mathbf Q(\mathbf p_j-\mathbf p_i)\|^2=\mathrm{ATE}^2$, the errors of the displacements in the world frame. Expressing them in the cameras' frames adds the terms $\mathbf c_{ij}$ with the rotation errors under ATE's alignment, which the position fit does not control: rotating the predicted centres against the orientations leaves ATE unchanged (Proposition~\ref{prop:ate_position_invariance}) but can change the displacement errors $\boldsymbol\varepsilon_{ij}$.
\hfill$\square$

\emph{Numerical check.} On every saved prediction of the 11 models on Sintel, TUM and ScanNet, unmodified, with the centres rotated by 30$^\circ$, and half and fully straightened, and on the six camera-axis lines (5,800 cases; tools/allpairs\_identity.py, results/package1/allpairs\_identity.csv), \cref{eq:allpairs_bound} holds in every case to numerical precision (relative tolerance $10^{-5}$, as the stored rotations are orthonormal to about $10^{-7}$). On the unmodified predictions, $\mathrm{RPE}_{\mathrm{all}}$ and PA-ATE differ by a median of 0.6, 0.3 and 1.3\% (Sintel, TUM, ScanNet; 90th percentile 4.3, 2.1 and 6.8\%), whereas ATE falls short of it by a median of 20, 11 and 18\%. On Bonn, Wild-SLAM and TUM-X (full sequences and windows) and the KITTI and 7-Scenes windows (4,890 further cases; tools/allpairs\_identity\_more.py), the bound also holds, except for 84 image-free lines on 7-Scenes, whose fitted scale is zero and which miss it by at most $3\times10^{-4}$ of the score, as the dataset's rotations are orthonormal only to about $10^{-4}$. There the unmodified predictions agree with PA-ATE to a median of 0.2 to 3.7\% per dataset, and ATE falls 5 to 35\% short. With the centres rotated, ATE departs from $\mathrm{RPE}_{\mathrm{all}}$ by a median factor of 2.9, 3.0 and 4.4 and PA-ATE by at most 0.4\% at the median. RPE$_t$ over a gap of $\delta$ frames is the restriction of $\mathrm{RPE}_{\mathrm{all}}$ to the pairs $(i,i+\delta)$, up to the factor $\sqrt2$ and the scale estimate, so its 10-frame version scores a subset of the same pairs, 3 to 5\% of the distinct pairs on clips of 20 to 50 frames. The all-interval RPE of the TUM benchmark~\cite{tum} averages the RMS error of each interval over the intervals instead of pooling the pairs.
}

\paragraph{Numerical check of \cref{eq:threshold}.} At fixed alignment, the condition classified the outcome of straightening exactly in all 5,280 checks: the 1056 model--clip pairs of the development benchmarks (\cref{tab:s_rule_dev}) at five values of $\alpha$ (results/proposition\_checks).

\paragraph{The refit term.} After straightening, the alignment is refitted, which can only lower the error further, so every miss of \cref{eq:threshold} is a gain that the refit adds (Proposition~\ref{prop:straightening_threshold}). To first order, one Gauss--Newton step from the current alignment removes the projection of the straightened error $\mathbf E_\alpha$ onto the tangent space of the alignment at the straightened, aligned prediction $\hat{\mathbf P}_\alpha$, spanned by the three translations, the three infinitesimal rotations $\boldsymbol\omega\times(\hat{\mathbf p}_{\alpha,i}-\bar{\hat{\mathbf p}}_\alpha)$ and the scale direction $\hat{\mathbf p}_{\alpha,i}-\bar{\hat{\mathbf p}}_\alpha$. For PA-ATE the rotation is fixed by the orientations, which straightening leaves unchanged, so only translation and scale are refitted: the refit is a linear least-squares problem, and its term is exact (with the scale clipped at zero as in Eq.~(\mref{eq:anchor_fit})). With $A$ the decrease of the mean squared residual under this linearised refit ($A=\|\Pi\mathbf E_\alpha\|^2/n$ for the projection $\Pi$ onto the tangent space, and for PA-ATE the decrease at the constrained minimum over translation and scale increments with $s+\delta s\ge0$), the expansion in the proof of Proposition~\ref{prop:straightening_threshold} gives that the error after refitting decreases, to first order, exactly when
\[
 \alpha^2S^2+(1-\alpha)^2N^2+2\alpha(1-\alpha)cSN-N^2<A ,
\]
of which \cref{eq:threshold} is the case $A=0$. This condition gives the outcome after refitting for 1055 of the 1056 development pairs and for all 792 held-out, 1034 ScanNet, 624 added-method, 352 PA-ATE and 504 newly registered pairs (+refit in \cref{tab:s_rule_dev} and Tab.~\mref{tab:survey}; tools/refit\_term.py, results/straighten\_test/refit\_term.json). The exception, $\pi^3$ on Bonn synchronous, gains 0.4\%, and the first-order term falls just short of the fixed-alignment loss (0.069 against $0.086\,N^2$). These are 4362 scorings of 4010 distinct model--clip evaluations (LASER and AnyCam in two configurations each): the 352 PA-ATE pairs re-score development predictions. At $\alpha=0.25$, 0.75 and 1 the condition gives the outcome for every one of the 4362 scorings (tools/refit\_term\_alpha.py). Without the term, \cref{eq:threshold} certifies 77\% of the development gains, 96\% of the held-out ones and 28 of the 56 on ScanNet (\cref{tab:s_rule_dev}).
\begin{table}[h]
\centering\small
\setlength{\tabcolsep}{3pt}
\begin{tabular*}{\linewidth}{@{\extracolsep{\fill}}lccccccccc@{}}
\toprule
& & & \multicolumn{5}{c}{correct (\%)} & & \\
\cmidrule(lr){4-8}
Set & pairs & SNR & $R^2_{\text{line}}$ & error & $c=0$ & criterion & +refit & gains & certified \\
\midrule
\multicolumn{10}{@{}l}{\emph{Development}} \\
Sintel & 154 & 0.86 & 69 & 63 & 98 & 97 & 100 & 108 & 104 \\
TUM & 88 & 6.1 & 81 & 92 & 93 & 95 & 100 & 17 & 13 \\
TUM-X & 330 & 11 & 85 & 94 & 94 & 91 & 100 & 37 & 8 \\
Bonn & 154 & 2.6 & 75 & 73 & 86 & 95 & 99.4 & 38 & 30 \\
Wild-SLAM & 330 & 20 & 94 & 98 & 98 & 99 & 100 & 14 & 10 \\
all & 1056 & 7.4 & 84 & 88 & 94 & 95 & 99.9 & 214 & 165 \\
\midrule
\multicolumn{10}{@{}l}{\emph{Held-out}} \\
KITTI, 2\,s & 242 & 1.2 & 59 & 41 & 91 & 100 & 100 & 143 & 143 \\
KITTI, 10\,s & 242 & 2.1 & 62 & 59 & 93 & 99.6 & 100 & 109 & 108 \\
7-Scenes, 2\,s & 154 & 3.1 & 66 & 76 & 87 & 100 & 100 & 26 & 26 \\
7-Scenes, 5\,s & 154 & 4.7 & 57 & 86 & 86 & 92 & 100 & 21 & 9 \\
all & 792 & 2.6 & 61 & 62 & 90 & 98 & 100 & 299 & 286 \\
\midrule
ScanNet, registered & 1034 & 5.7 & 94 & 94 & 95 & 97 & 100 & 56 & 28 \\
\midrule
Added methods & 624 & 4.5 & 89 & 84 & 90 & 97 & 100 & 85 & 68 \\
PA-ATE & 352 & 3.3 & 80 & 77 & 92 & 97 & 100 & 114 & 104 \\
\midrule
\multicolumn{10}{@{}l}{\emph{Registered new benchmarks}} \\
Lightspeed & 324 & 1.3 & 63 & 62 & 95 & 99.7 & 100 & 178 & 177 \\
DyCheck & 108 & 14 & 93 & 98 & 99 & 100 & 100 & 1 & 1 \\
ADT & 72 & 6.7 & 83 & 81 & 92 & 90 & 100 & 12 & 5 \\
all & 504 & 3.0 & 72 & 73 & 95 & 98 & 100 & 191 & 183 \\
\bottomrule
\end{tabular*}
\caption{All rows of Tab.~\mref{tab:survey}, with the development benchmarks by dataset (all windows). Correct (\%): straightness alone (gain if $R^2_{\text{line}}\ge0.9513$), error alone (gain if $N/r\ge0.3165$, $r$ the RMS radius of the true path), each threshold the one with the highest accuracy on the development pairs (smallest such value on ties); \cref{eq:threshold} with $c=0$ and with the measured $c$, and the criterion with the refit's first-order term (+refit, the refit term in \cref{sec:s_proofs}). Gains: pairs whose ATE falls under half straightening; certified: those that \cref{eq:threshold} with the measured $c$ identifies. The condition is sufficient, not necessary, after refitting, so a pair it does not certify is not predicted to lose. Missed gains are typically smaller: a median ATE decrease of 6.2\%, against 36.4\% for certified gains, on the development and held-out pairs (ScanNet: 13.3\% against 23.2\%).}
\label{tab:s_rule_dev}
\end{table}

\section{Literature survey}
\label{sec:s_lit}
We retained 384 Sintel ATE values from 84 papers under the dynamic-scene protocol of MonST3R, from the sources we found up to 25 October 2026. Candidates came from 38 arXiv queries (13,836 papers since October 2024) and a full-text screen, followed by a reading of every paper whose text puts ``Sintel'' near ``ATE'' or ``absolute trajectory''. Each value was copied from its source table, and a second, independent check of every new value below the line found each one printed as recorded (results/lit\_values\_v2.csv; notes/lit\_resweep\_2026-10-25.md; notes/lit\_verify\_below\_line\_2026-10-25.md; tools/lit\_points.py). Papers that are not on arXiv or match none of the queries can still be missing.
A value is excluded when its source states a setting that differs from the protocol (other clips or frames, ground-truth intrinsics or depth as input, no scale alignment, another statistic), or when the method's released evaluation code shows one. An incomplete description is not a reason for exclusion. This excludes 88 values from 35 sources (\cref{tab:s_lit_excluded}). Two exclusions rest on released code: AnyCam's 0.078 is a mean, not an RMS error (anycam/scripts/evaluate\_trajectories.py, l.~252, reproduced in Sec.~\ref{sec:s_impl}), and LEAP-VO's~\cite{leapvo} released Sintel evaluation reads the ground-truth intrinsics of every frame and passes them to the model (github.com/chiaki530/leapvo, main/stream.py l.~57--69, main/eval.py l.~54). One exclusion is an inference, marked as such in \cref{tab:s_lit_excluded}: MONA~\cite{mona} (January 2025, 0.029) runs inside LEAP-VO but states no setting and has released no code. Counted, its own value would make 111 points with 85 above the line, and it would be the one value below the line published before July 2025 within the window of Fig.~\mref{fig:progress}.
A method is represented by the median of its own values or, if its paper reports none, by that of the re-evaluations of its released model, and each method paper by its best variant, which favours the methods rather than our claim. This gives the 110 points of Fig.~\mref{fig:progress}, 29 of them from re-evaluations and 85 above the line. One is printed as 0.092, so its order relative to the line's 0.0915 rests on rounding.
All 35 included values published before July 2025 lie above the line. Lower values first appear in July 2025 (SpatialTrackerV2, $\pi^3$), and of the methods first released since then, 24 report one, two of them streaming (LASER, G-CUT3R~\cite{gcut3r}). MegaSaM, released in December 2024, lies below the line through re-evaluations from December 2025 on. Before DUSt3R, outside the window of Fig.~\mref{fig:progress}, RoDynRF~\cite{rodynrf} (January 2023) reported 0.089 on the same clips. The term was added after the development, held-out and ScanNet results and before the three benchmarks of Sec.~\mref{sec:newbench} were registered, so their 504 outcomes are out of sample for it.

\paragraph{Setting sensitivity.} Relative differences below use the larger value as denominator. Sorted by value, the 25 admitted values below the line leave 24 gaps between consecutive values, 22 of them under 12\% (at most 11.5\%). One method under reproducible changes differs by more: AnyCam's evaluation script reports a mean error, 0.078, 16\% below its RMS ATE of 0.093 (Sec.~\ref{sec:s_impl}). LASER's two released configurations score 0.064 and 0.085, 25\% apart. The GPU type changes the mean Sintel ATE of the four DA3 models by 3 to 14\% and of the other seven by at most 1.2\% (results/repro/compare\_summary.csv). The first is a difference in the statistic reported, the second a configuration choice that may also change the quality of the method, and the third a sensitivity to hardware. Clip-bootstrap intervals hold these conditions fixed, so stating them is a separate requirement (tools/lit\_records.py; results/lit\_gaps.json).
Eight sources are kept by the rule although their values disagree with all other sources or leave the setting open (notes a--g in \cref{tab:s_literature}). We read each of these sources in full: every value is printed as recorded, none states a setting that differs from the protocol, and none states its setting completely, so the rule keeps them. The discrepancies in the notes are facts about the sources' tables, not evidence that they used another protocol. Their released code does not settle the setting either (notes/low\_values\_code\_check\_2026-10-25.md): G-CUT3R, SS3D~\cite{ss3d} and SS3D+VLRC~\cite{vlrc} have released no code, and SpatialTrackerV2 and UniQuery4R have released inference code but no Sintel evaluation script. For SpatialTrackerV2 and UniQuery4R we therefore ran the released models ourselves (0.115 and 0.033, Sec.~\ref{sec:s_impl}). Dropping all values of these eight sources and recomputing every point gives 101 points, 81 above the line. The count is therefore a statement about the published record under this inclusion rule, not about methods evaluated under an identical, fully documented protocol.
Few sources document the protocol fully: of the 94 points with a single source, 7 state both the 14 clips and a similarity alignment.
All 28 papers of our first survey report ATE with translational and rotational RPE, and none reports an orientation-aware global alignment (notes/lit\_pose\_scores\_28papers.md).

{\scriptsize\setlength{\tabcolsep}{3pt}
\begin{longtable}{@{}lccccccc@{}}
\caption{The 110 points of Fig.~\mref{fig:progress}, one per method paper (its best variant), sorted by ATE; values to four decimals (medians where several). The line scores 0.0915, so a published 0.092 lies above it, and 85 points lie above it. Method date: first arXiv version of the method. Value: the method\textquoteright s own value (own) or the median of re-evaluations (re-eval.). Source: the paper that reported the value, or the number of re-evaluations whose median is used when the method\textquoteright s own paper reports none. 14 and Sim(3): whether the source states the 14 clips and a similarity (scale) alignment (for a median: in how many of its sources, the number in the source column). Notes: a: the source's CUT3R row (0.086) lies below all 26 other CUT3R values (0.134--0.235); b: the source does not state its Sintel clips for pose, and uses 18 clips for depth; c: the source compares with AnyCam's mean error and may report a mean; d: the source's VGGT row (0.100) lies far below the median of 30 other VGGT values (0.171), and our run of UniQuery4R's released model scores 0.033; e: the source states a ``slightly revised'' setup without details; f: the source's VGGT4D row equals VGGT4D's subset-clip value; g: the source's VGGT row (0.082) lies far below the median of 30 other VGGT values (0.171), and our run of SpatialTrackerV2's released full pipeline scores 0.115, two re-evaluations 0.126 and 0.133. Excluded sources: \cref{tab:s_lit_excluded}.}\label{tab:s_literature}\\
\toprule
Method & ATE & method date & source & value & 14 & Sim(3) & note \\
\midrule
\endfirsthead
\toprule
Method & ATE & method date & source & value & 14 & Sim(3) & note \\
\midrule
\endhead
\bottomrule
\endfoot
UniQuery4R & 0.0310 & 2026-08 & 2608.17283 & own & -- & -- & d \\
VGGT-Omega & 0.0400 & 2026-05 & median of 5 & re-eval. & 0 & 1 & d \\
GenCeption (Specialist - L) & 0.0500 & 2026-07 & 2607.09024 & own & -- & -- &  \\
PURE (DA3) & 0.0520 & 2026-10 & 2610.14698 & own & -- & yes &  \\
SpatialTrackerV2 & 0.0540 & 2025-07 & 2507.12462 & own & -- & yes & g \\
LASER (Pi3) & 0.0610 & 2025-12 & 2512.13680 & own & -- & yes &  \\
ARROW & 0.0613 & 2026-10 & 2610.01314 & own & -- & -- &  \\
G-CUT3R (512, no priors) & 0.0630 & 2025-08 & 2508.11379 & own & -- & yes & a \\
D4RT & 0.0650 & 2025-12 & 2512.08924 & own & yes & yes &  \\
HD-VGGT & 0.0710 & 2026-03 & 2603.27222 & own & -- & -- &  \\
MGE & 0.0710 & 2026-10 & 2610.06813 & own & -- & yes &  \\
Spark3R+Pi3 & 0.0730 & 2026-05 & 2605.06270 & own & -- & yes &  \\
Pi3 & 0.0740 & 2025-07 & 2507.13347 & own & -- & yes &  \\
FasterVGGT (on Pi3, 25\%) & 0.0760 & 2025-09 & 2512.02541 & re-eval. & -- & -- &  \\
Syn4D: 4RC re-trained & 0.0760 & 2026-05 & 2605.05207 & own & -- & -- &  \\
OmniX & 0.0770 & 2026-07 & 2610.01314 & re-eval. & -- & -- &  \\
LongDPM & 0.0780 & 2026-05 & 2605.17303 & own & -- & -- &  \\
MegaSaM & 0.0850 & 2024-12 & median of 2 & re-eval. & 1 & 2 & b \\
4D3R & 0.0860 & 2025-11 & 2511.05229 & own & yes & yes &  \\
MoRe (FA) & 0.0877 & 2026-03 & 2603.05078 & own & -- & yes &  \\
SS3D + VLRC & 0.0880 & 2026-07 & 2607.02707 & own & -- & -- & c \\
TrackingWorld (DELTA) & 0.0880 & 2025-12 & 2512.08358 & own & -- & -- &  \\
4D-VGGT & 0.0890 & 2025-11 & 2511.18416 & own & -- & -- &  \\
SS3D (w/o ref.) & 0.0900 & 2026-04 & 2604.22686 & own & -- & -- & c \\
A\_pi3(2) (AVGGT on pi3) & 0.0910 & 2025-12 & 2512.02541 & own & -- & -- &  \\
Easi3R + BA w/ Dynapo mask & 0.0920 & 2025-12 & 2512.05398 & own & -- & yes & b \\
RoMo: TAPIR + TheiaSfM + RoMo masks & 0.0930 & 2024-11 & 2411.18650 & own & yes & -- &  \\
AnyCam & 0.0950 & 2025-03 & 2512.05398 & re-eval. & -- & yes & b \\
WorldMirror & 0.0960 & 2025-10 & 2510.10726 & own & -- & -- &  \\
SirenPose & 0.0970 & 2025-12 & 2512.20531 & own & -- & -- &  \\
SceneScribe-1M: MonST3R fine-tuned & 0.0990 & 2026-04 & 2604.07990 & own & -- & -- &  \\
LingBot-Map & 0.1000 & 2026-04 & 2604.14141 & own & -- & yes &  \\
C4D-M & 0.1030 & 2025-10 & 2510.14960 & own & -- & -- &  \\
DATAP-SfM (full set) & 0.1040 & 2024-11 & 2411.13291 & own & yes & -- &  \\
Gen4P-II & 0.1040 & 2026-10 & 2610.22371 & own & yes & yes &  \\
DA3 & 0.1050 & 2025-11 & median of 8 & re-eval. & 2 & 4 & de \\
V-DPM & 0.1050 & 2026-01 & 2601.09499 & own & -- & -- &  \\
DAS3R & 0.1070 & 2024-12 & 2412.19584 & own & -- & -- &  \\
DynamicGen & 0.1080 & 2025-12 & 2512.03000 & own & -- & -- &  \\
MonST3R & 0.1080 & 2024-10 & 2410.03825 & own & yes & yes &  \\
TrajVG & 0.1080 & 2026-02 & 2602.04439 & own & -- & -- &  \\
Easi3R (MonST3R backbone) & 0.1100 & 2025-03 & 2503.24391 & own & -- & yes &  \\
Uni4D & 0.1100 & 2025-03 & 2503.21761 & own & -- & -- &  \\
SM4RT & 0.1120 & 2026-07 & 2610.01314 & re-eval. & -- & -- &  \\
R3 (streaming) & 0.1150 & 2026-05 & 2605.26519 & own & -- & -- &  \\
Causal State Evolution & 0.1160 & 2026-10 & 2610.27594 & own & -- & -- &  \\
Track4World & 0.1190 & 2026-03 & 2603.02573 & own & yes & -- &  \\
UniT (G = N) & 0.1240 & 2026-05 & 2605.21131 & own & -- & yes & e \\
ZipMap (w/ ref) & 0.1250 & 2026-03 & 2603.04385 & own & -- & -- &  \\
Align3R (Depth Pro) & 0.1280 & 2024-12 & 2412.03079 & own & yes & -- &  \\
DAGE (252px) & 0.1320 & 2026-03 & 2603.03744 & own & yes & yes &  \\
L4P (e2e FT) & 0.1320 & 2025-02 & 2502.13078 & own & -- & -- &  \\
4RC & 0.1336 & 2026-02 & median of 4 & re-eval. & 1 & 1 & d \\
IVGT & 0.1400 & 2026-05 & 2605.16258 & own & -- & yes &  \\
CameraVLM & 0.1480 & 2026-10 & 2610.17242 & own & -- & -- &  \\
SpatialVID: VGGT fine-tuned & 0.1480 & 2025-09 & 2509.09676 & own & -- & -- &  \\
Scal3R Symmetric (VPT) & 0.1540 & 2026-09 & 2609.04201 & own & -- & yes &  \\
RayMap3R & 0.1660 & 2026-03 & 2603.20588 & own & -- & yes &  \\
FastVGGT & 0.1670 & 2025-09 & median of 3 & re-eval. & 0 & 1 &  \\
VGGT-Long & 0.1700 & 2025-07 & 2604.14141 & re-eval. & -- & yes &  \\
VGGT & 0.1710 & 2025-03 & median of 31 & re-eval. & 4 & 17 & bdeg \\
DELTA & 0.1720 & 2024-10 & 2410.24211 & own & yes & yes &  \\
PAGE-4D & 0.1750 & 2025-10 & 2610.27594 & re-eval. & -- & -- &  \\
CogniMap3D & 0.1760 & 2026-01 & 2601.08175 & own & -- & -- &  \\
SparseVGGT & 0.1770 & 2025-09 & 2602.23361 & re-eval. & -- & -- &  \\
OmniWorld: CUT3R fine-tuned & 0.1780 & 2025-09 & 2509.12201 & own & -- & -- &  \\
Mem3R & 0.1800 & 2026-04 & 2604.07279 & own & -- & -- &  \\
FLARE & 0.1825 & 2025-02 & median of 2 & re-eval. & 0 & 2 &  \\
Geo4D & 0.1850 & 2025-04 & 2504.07961 & own & yes & -- &  \\
MASt3R-GA & 0.1850 & 2024-06 & median of 7 & re-eval. & 1 & 5 &  \\
Aether & 0.1890 & 2025-03 & 2503.18945 & own & -- & yes &  \\
CUT3R + MeMix & 0.1900 & 2026-03 & 2603.15330 & own & -- & -- &  \\
POMATO & 0.1960 & 2025-04 & median of 2 & re-eval. & 1 & 0 &  \\
Dyna3 & 0.1970 & 2026-10 & 2610.01286 & own & -- & -- & f \\
STream3R-STAC & 0.1980 & 2026-03 & 2603.20284 & own & -- & yes &  \\
TTT3R & 0.2010 & 2025-09 & 2509.26645 & own & -- & yes &  \\
Adaptive World Memory model & 0.2070 & 2026-09 & 2609.21502 & own & -- & -- &  \\
SSR & 0.2090 & 2026-03 & 2603.14765 & own & -- & -- &  \\
TTSA3R & 0.2100 & 2026-01 & 2601.22615 & own & -- & yes &  \\
TurboVGGT & 0.2100 & 2026-05 & 2605.14315 & own & -- & -- &  \\
PAS3R & 0.2110 & 2026-03 & 2603.21436 & own & -- & -- &  \\
Any4D & 0.2120 & 2025-12 & median of 3 & re-eval. & 1 & 1 & d \\
CUT3R & 0.2130 & 2025-01 & 2501.12387 & own & -- & yes &  \\
One4D & 0.2130 & 2025-11 & 2511.18922 & own & -- & -- &  \\
Ray-Aware Pointer Memory & 0.2130 & 2026-05 & 2605.05749 & own & -- & -- &  \\
STream3R & 0.2130 & 2025-08 & 2508.10893 & own & -- & yes &  \\
Complet4R & 0.2150 & 2026-03 & 2603.27300 & own & -- & -- &  \\
MapAnything & 0.2165 & 2025-09 & median of 6 & re-eval. & 3 & 4 & e \\
BAT3R & 0.2190 & 2026-10 & 2610.16330 & own & -- & yes &  \\
StreamVGGT & 0.2190 & 2025-07 & 2507.11539 & own & -- & yes &  \\
WinT3R & 0.2250 & 2025-09 & median of 3 & re-eval. & 0 & 3 &  \\
OmniStream & 0.2270 & 2026-03 & 2603.12265 & own & -- & -- &  \\
RetrieveVGGT & 0.2270 & 2026-05 & 2605.09644 & own & -- & yes &  \\
MUT3R & 0.2280 & 2025-12 & 2512.03939 & own & -- & yes &  \\
MASt3R-SLAM & 0.2330 & 2024-12 & 2609.04201 & re-eval. & -- & yes &  \\
VGG-T3 & 0.2340 & 2026-02 & 2602.23361 & own & -- & -- &  \\
Evict3R (B = 0.9) & 0.2350 & 2025-09 & 2509.17650 & own & -- & yes &  \\
InfiniteVGGT & 0.2370 & 2026-01 & median of 3 & re-eval. & 1 & 2 &  \\
Cambrian-P & 0.2390 & 2026-05 & 2605.22819 & own & -- & -- &  \\
ViPE & 0.2410 & 2025-08 & 2610.14698 & re-eval. & -- & yes &  \\
MUSt3R & 0.2420 & 2025-03 & 2609.04201 & re-eval. & -- & yes &  \\
Slot3R (VPC-A) & 0.2539 & 2026-10 & 2610.12282 & own & yes & yes &  \\
Mem4D & 0.2630 & 2025-08 & 2508.07908 & own & -- & yes &  \\
Fast3R & 0.2740 & 2025-01 & median of 3 & re-eval. & 1 & 3 &  \\
DUSt3R (online/streaming) & 0.2900 & 2023-12 & median of 3 & re-eval. & 0 & 3 & g \\
G2VLM & 0.3010 & 2025-11 & 2605.22819 & re-eval. & -- & -- &  \\
VGGT-SLAM & 0.3030 & 2025-05 & 2512.13680 & re-eval. & -- & yes &  \\
Spann3R & 0.3290 & 2024-08 & median of 8 & re-eval. & 2 & 8 &  \\
Point3R & 0.3510 & 2025-07 & 2507.02863 & own & -- & yes &  \\
Pow3R (512, no priors) & 0.5780 & 2025-03 & 2508.11379 & re-eval. & -- & yes & a \\
\end{longtable}}

{\scriptsize\setlength{\tabcolsep}{3pt}
\begin{longtable}{@{}lcp{0.30\linewidth}p{0.46\linewidth}@{}}
\caption{The 35 excluded sources and their exclusion bases, sorted by their lowest excluded value (88 values in all; results/lit\_excluded\_v2.csv).}\label{tab:s_lit_excluded}\\
\toprule
Source (arXiv) & lowest & exclusion basis & methods \\
\midrule
\endfirsthead
\toprule
Source (arXiv) & lowest & exclusion basis & methods \\
\midrule
\endhead
\bottomrule
\endfoot
2602.24290 & 0.012 & two-frame inputs, ground-truth intrinsics & UFO-4D \\
2411.13291 & 0.015 & 9 of the 14 clips & DATAP-SfM (COLMAP subset) \\
2605.12774 & 0.017 & 18 clips, scale-normalised (MegaSaM setup) & DROID-SLAM, MegaSaM, VGGT, ViPE, WildGS-SLAM, WildPose, pi3 \\
2412.04463 & 0.018 & 18 clips, scale-normalised (MegaSaM setup) & ACE-Zero, ACE-Zero (calibrated), CasualSAM, CasualSAM (calibrated), LEAP-VO (calibrated), MegaSaM, MegaSaM (calibrated), MonST3R, ParticleSfM, ParticleSfM (calibrated), RoDynRF, RoDynRF (calibrated) \\
2512.06868 & 0.019 & CasualSAM/MegaSaM setup, cm units & BA-Track, DynSLAM-3DPrior, WildGS-SLAM \\
2605.10525 & 0.020 & video-depth protocol; ATE undefined & DA3 (GemDepth), GemDepth, VGGT (GemDepth) \\
2501.13183 & 0.029 & inferred: ground-truth intrinsics via LEAP-VO (not stated) & LEAP-VO (MONA re-run), MONA + LEAP-VO \\
2504.14516 & 0.034 & ground-truth intrinsics as input & BA-Track, DytanVO, TartanVO \\
2503.21761 & 0.035 & ground-truth intrinsics as input (released evaluation code) & DPVO, LEAP-VO, Uni4D (GT intrinsics) \\
2606.21300 & 0.035 & all 23 clips (MegaSaM setup) & SCOPE + MegaSAM \\
2401.01887 & 0.037 & ground-truth intrinsics as input (released evaluation code) & LEAP-VO (own) \\
2511.18416 & 0.037 & ground-truth intrinsics as input (released evaluation code) & DPVO, LEAP-VO \\
2610.14698 & 0.037 & trained on the test clips & VGGT post-trained on Sintel test images (after) \\
2605.06270 & 0.039 & same measurement as VGGT-Omega (one model, one paper) & Spark3R+VGGT-Omega \\
2603.02573 & 0.045 & caption and text give different datasets & MegaSaM (Opt.), Track4World (FF.+Opt.) \\
2602.20157 & 0.048 & sparsely sampled frames & Flow3r \\
2508.11379 & 0.054 & ground-truth intrinsics or depth as input & G-CUT3R (224, +D), G-CUT3R (224, +K), G-CUT3R (224, +K+D), G-CUT3R (512, +D), G-CUT3R (512, +K), G-CUT3R (512, +K+D), Pow3R (+D), Pow3R (+K), Pow3R (+K+D) \\
2511.19971 & 0.076 & subset of dynamic clips & VGGT4D \\
2610.01286 & 0.076 & subset of dynamic clips & VGGT4D (copy) \\
2503.23282 & 0.078 & mean error, not RMSE (evaluation code) & AnyCam, AnyCam w/o ref. \\
2512.08358 & 0.087 & frames 30-50 only & TrackingWorld (DELTA) (Sintel frames 30-50) \\
2410.03825 & 0.089 & ground-truth intrinsics as input (released evaluation code) & DPVO, DROID-SLAM, LEAP-VO \\
2607.10840 & 0.108 & 16-frame clips & OmniX \\
2609.15018 & 0.110 & ground-truth intrinsics as input & G-ray, MapAnything (with GT calibration) \\
2609.09145 & 0.114 & no scale alignment & DA3, Point4D, VDPM \\
2502.13078 & 0.115 & ground-truth intrinsics as input & L4P (e2e FT), L4P (head FT) \\
2605.30060 & 0.128 & all training clips, clean pass & DA3, Pi3, Stream3R, StreamVGGT, VGGT, ViGeo (offline), ViGeo (online) \\
2503.21082 & 0.155 & 17-frame clips & DUSt3R (Sora3R re-run), MonST3R (Sora3R re-run), Sora3R \\
2512.20531 & 0.163 & ground-truth intrinsics as input & DPVO \\
2510.17568 & 0.178 & 10 frames per clip & PAGE-4D \\
2508.07769 & 0.210 & poses from synthesised frames & Dream4D \\
2603.14528 & 0.214 & frames plus events, frame skipping & Interp3R (skip = 1) \\
2610.01314 & 0.233 & not the paper's method & OpenD4RT \\
2609.04201 & 0.381 & no scale alignment & CUT3R (no scale alignment), Scal3R (no scale alignment) \\
2605.21131 & 0.394 & no scale alignment & DepthAnything3 (metric-scale), MapAnything (metric-scale) \\
\end{longtable}}

\section{Additional Sintel results}
\label{sec:s_sintel}

\paragraph{Methods added later.} Of the 11 models, DA3-Small, DA3-Base and MapAnything were added after the others, which were chosen before any held-out result of Sec.~\mref{sec:cause} was computed. Published Sintel ATE of the five methods added after the main analysis (Tab.~\mref{tab:sintel}): MonST3R~\cite{monst3r} 0.108, UniQuery4R~\cite{uniquery4r} 0.031 (internal backbone), LASER~\cite{laser} 0.061 (no loop closure), AnyCam~\cite{anycam} 0.078 (a mean; its evaluation configuration), SpatialTrackerV2~\cite{spatrackerv2} 0.054 (full pipeline; re-evaluated 0.126~\cite{d4rt}, 0.133~\cite{track4world}).

\paragraph{Per-clip scores.}
\Cref{tab:s_perclip_ate,tab:s_perclip_anchored} give ATE and PA-ATE for every model and clip (tools/build\_supp\_tables.py).
The mean over 14 clips is dominated by cave\_2 and temple\_3, on which every model has a large error.

\begin{table*}[t]\centering\footnotesize\setlength{\tabcolsep}{3pt}
\begin{tabular}{@{}lcccccccccccc@{}}
\toprule
Clip & $\pi^3$ & $\pi^3$X & VGGT & DA3-S & DA3-B & DA3-L & DA3-G & MapAny. & CUT3R & TTT3R & StrVGGT & Line \\
\midrule
alley\_2 & 0.007 & 0.013 & 0.009 & 0.036 & 0.032 & 0.010 & 0.006 & 0.032 & 0.030 & 0.025 & 0.033 & 0.039 \\
ambush\_4 & 0.045 & 0.059 & 0.168 & 0.145 & 0.086 & 0.062 & 0.063 & 0.109 & 0.162 & 0.158 & 0.139 & 0.110 \\
ambush\_5 & 0.038 & 0.023 & 0.043 & 0.104 & 0.086 & 0.044 & 0.039 & 0.083 & 0.073 & 0.073 & 0.095 & 0.032 \\
ambush\_6 & 0.042 & 0.045 & 0.155 & 0.257 & 0.305 & 0.146 & 0.212 & 0.295 & 0.292 & 0.242 & 0.255 & 0.129 \\
cave\_2 & 0.257 & 0.922 & 1.195 & 0.554 & 0.335 & 0.892 & 0.704 & 1.195 & 0.940 & 1.071 & 0.968 & 0.270 \\
cave\_4 & 0.022 & 0.018 & 0.095 & 0.133 & 0.101 & 0.041 & 0.011 & 0.110 & 0.154 & 0.146 & 0.080 & 0.046 \\
market\_2 & 0.054 & 0.017 & 0.021 & 0.031 & 0.026 & 0.010 & 0.020 & 0.036 & 0.071 & 0.036 & 0.028 & $<$0.001 \\
market\_5 & 0.037 & 0.030 & 0.129 & 0.546 & 0.264 & 0.114 & 0.033 & 0.484 & 0.285 & 0.291 & 0.541 & 0.103 \\
market\_6 & 0.020 & 0.010 & 0.041 & 0.084 & 0.054 & 0.031 & 0.015 & 0.078 & 0.082 & 0.105 & 0.097 & 0.265 \\
shaman\_3 & 0.005 & 0.002 & 0.007 & 0.014 & 0.013 & 0.010 & 0.003 & 0.015 & 0.016 & 0.009 & 0.008 & 0.001 \\
sleeping\_1 & 0.007 & 0.004 & 0.015 & 0.011 & 0.007 & 0.011 & 0.004 & 0.103 & 0.016 & 0.014 & 0.029 & 0.021 \\
sleeping\_2 & 0.004 & 0.004 & 0.006 & 0.019 & 0.010 & 0.004 & 0.002 & 0.013 & 0.014 & 0.011 & 0.012 & 0.008 \\
temple\_2 & 0.048 & 0.033 & 0.053 & 0.148 & 0.082 & 0.025 & 0.024 & 0.105 & 0.114 & 0.078 & 0.228 & 0.222 \\
temple\_3 & 0.558 & 0.399 & 0.463 & 0.445 & 0.468 & 0.585 & 0.436 & 0.500 & 0.708 & 0.656 & 0.569 & 0.034 \\
\midrule
mean & 0.082 & 0.113 & 0.171 & 0.180 & 0.133 & 0.142 & 0.112 & 0.226 & 0.211 & 0.208 & 0.220 & 0.092 \\
median & 0.037 & 0.021 & 0.048 & 0.118 & 0.084 & 0.036 & 0.022 & 0.104 & 0.098 & 0.092 & 0.096 & 0.043 \\
\bottomrule
\end{tabular}
\caption{Per-clip ATE on the 14 Sintel clips (official protocol). Line: the line along the optical axis (all six axis directions give the same ATE).}
\label{tab:s_perclip_ate}
\end{table*}
\begin{table*}[t]\centering\footnotesize\setlength{\tabcolsep}{3pt}
\begin{tabular}{@{}lcccccccccccc@{}}
\toprule
Clip & $\pi^3$ & $\pi^3$X & VGGT & DA3-S & DA3-B & DA3-L & DA3-G & MapAny. & CUT3R & TTT3R & StrVGGT & Line \\
\midrule
alley\_2 & 0.013 & 0.014 & 0.011 & 0.051 & 0.039 & 0.018 & 0.006 & 0.034 & 0.040 & 0.034 & 0.041 & 0.479 \\
ambush\_4 & 0.047 & 0.059 & 0.169 & 0.168 & 0.116 & 0.067 & 0.071 & 0.129 & 0.177 & 0.175 & 0.159 & 0.177 \\
ambush\_5 & 0.073 & 0.084 & 0.069 & 0.109 & 0.096 & 0.093 & 0.062 & 0.086 & 0.080 & 0.073 & 0.110 & 0.253 \\
ambush\_6 & 0.048 & 0.050 & 0.185 & 0.288 & 0.315 & 0.155 & 0.300 & 0.303 & 0.310 & 0.289 & 0.258 & 0.138 \\
cave\_2 & 0.570 & 1.223 & 1.223 & 0.948 & 0.867 & 1.222 & 1.031 & 1.223 & 1.208 & 1.223 & 1.223 & 1.223 \\
cave\_4 & 0.028 & 0.025 & 0.096 & 0.200 & 0.157 & 0.045 & 0.017 & 0.164 & 0.207 & 0.186 & 0.120 & 0.250 \\
market\_2 & 0.063 & 0.018 & 0.023 & 0.031 & 0.029 & 0.018 & 0.040 & 0.042 & 0.072 & 0.037 & 0.060 & 0.074 \\
market\_5 & 0.074 & 0.044 & 0.155 & 0.571 & 0.314 & 0.269 & 0.173 & 0.525 & 0.337 & 0.371 & 0.566 & 2.149 \\
market\_6 & 0.031 & 0.020 & 0.082 & 0.174 & 0.104 & 0.088 & 0.036 & 0.188 & 0.164 & 0.200 & 0.165 & 1.053 \\
shaman\_3 & 0.005 & 0.003 & 0.007 & 0.033 & 0.013 & 0.010 & 0.005 & 0.016 & 0.018 & 0.012 & 0.009 & 0.064 \\
sleeping\_1 & 0.009 & 0.010 & 0.016 & 0.012 & 0.012 & 0.013 & 0.006 & 0.112 & 0.030 & 0.022 & 0.033 & 0.112 \\
sleeping\_2 & 0.005 & 0.004 & 0.009 & 0.050 & 0.022 & 0.006 & 0.007 & 0.017 & 0.016 & 0.015 & 0.016 & 0.190 \\
temple\_2 & 0.083 & 0.035 & 0.061 & 0.246 & 0.101 & 0.036 & 0.057 & 0.112 & 0.206 & 0.118 & 0.241 & 0.628 \\
temple\_3 & 0.788 & 0.807 & 0.749 & 0.488 & 0.513 & 0.807 & 0.464 & 0.525 & 0.802 & 0.723 & 0.589 & 0.770 \\
\midrule
mean & 0.131 & 0.171 & 0.204 & 0.241 & 0.193 & 0.203 & 0.163 & 0.248 & 0.262 & 0.249 & 0.256 & 0.540 \\
median & 0.047 & 0.030 & 0.076 & 0.171 & 0.102 & 0.056 & 0.048 & 0.120 & 0.171 & 0.147 & 0.139 & 0.251 \\
\bottomrule
\end{tabular}
\caption{Per-clip PA-ATE on the 14 Sintel clips (official protocol). Line: constant orientation, along the one camera-axis direction (of six) with the lowest mean, as in Tab.~\mref{tab:sintel}.}
\label{tab:s_perclip_anchored}
\end{table*}

\begin{figure*}[t]
\centering
\includegraphics[width=\linewidth]{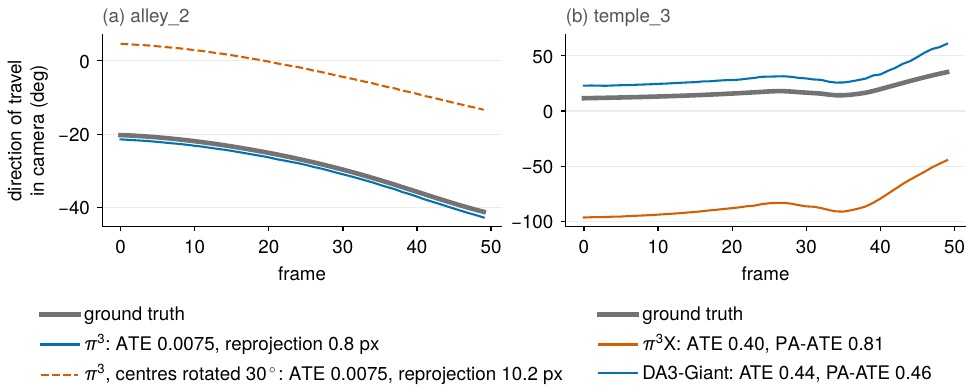}
\caption{Direction of travel (the constant-velocity fit of the camera centres) expressed in each frame's camera axes, as an azimuth in the camera's horizontal plane (0$^\circ$: along the optical axis). (a) alley\_2: $\pi^3$'s prediction and the same prediction with its centres rotated by 30$^\circ$ against its orientations, which has the same ATE. (b) temple\_3: $\pi^3$X, best of the eleven models by ATE, and DA3-Giant, best by PA-ATE.}
\label{fig:s_heading_examples}
\end{figure*}

\paragraph{Two real cases.}
\Cref{fig:s_heading_examples} shows the direction of travel, the constant-velocity fit of the centres, in each frame's camera axes. On alley\_2, rotating $\pi^3$'s centres by 30$^\circ$ against its orientations shifts this direction and leaves ATE at 0.0075, while the reprojection error of static points rises from 0.8 to 10.2\,px. On temple\_3, $\pi^3$X has the best ATE of the eleven models although its heading is 104$^\circ$ off. PA-ATE scores it as a static camera (0.81, the clip's RMS radius) and ranks first DA3-Giant, whose heading is 13$^\circ$ off. The heading check of PA-ATE also caught a convention error: with Bonn's marker-to-camera calibration, the median heading error of 8 of the 11 models falls from about 120$^\circ$ to 1--9$^\circ$.

\paragraph{Robustness of the headline.}
Leaving out one clip at a time, the line still beats 8 to 11 of the eleven models (\cref{tab:s_loo}).
Over 20,000 resamples of the 14 clips with replacement, the line ranks first among itself and the eleven models in 32\% of resamples, in the top two in 59\% and in the top four in 77\% (tools/rigor\_bootstrap\_hard.py; results/rigor/line\_rank\_bootstrap.csv).
This bounds how stable the second place on the full sample is. Leave-one-out, the geometric mean and the median over clips (8 and 7 of 11 models beaten) and the mean without cave\_2 and temple\_3 (6 of 11) answer different questions and do not remove this uncertainty.
In an exploratory analysis of the seven clips with $R^2_{\text{line}}$ below the overall median of 0.988 (their median 0.951), the line still ranks second by ATE, beating 10 models, and half straightening lowers the ATE of 10 of 11 models. Under PA-ATE the best of the six axis lines ranks 9th of 12 there by arithmetic means, while under geometric means every model is below every image-free reference (at most 0.203 against 0.212), and straightening helps 9 models (results/rigor/hard\_subset.csv; tools/hard\_subset\_gmean.py).
The regime is therefore not confined to the straightest clips. It ends where the motion is strongly nonlinear: on TUM, ScanNet, Bonn and Wild-SLAM every model beats the line and straightening raises every model's ATE (Tab.~\mref{tab:deletion}).

\begin{table}[t]\centering\footnotesize\setlength{\tabcolsep}{4pt}
\begin{tabular}{@{}lccccc@{}}
\toprule
& line beats & \multicolumn{2}{c}{ATE} & \multicolumn{2}{c}{PA-ATE} \\ \cmidrule(lr){3-4}\cmidrule(l){5-6}
Clip left out & (ATE) & gain & rev. & gain & rev. \\
\midrule
none (all 14) & 10 & 11 & 22 & 8 & 10 \\
alley\_2 & 10 & 11 & 22 & 8 & 10 \\
ambush\_4 & 10 & 11 & 22 & 8 & 11 \\
ambush\_5 & 10 & 10 & 21 & 8 & 10 \\
ambush\_6 & 10 & 11 & 23 & 8 & 10 \\
cave\_2 & 8 & 9 & 16 & 8 & 13 \\
cave\_4 & 10 & 11 & 21 & 8 & 10 \\
market\_2 & 10 & 10 & 22 & 8 & 10 \\
market\_5 & 10 & 11 & 21 & 8 & 8 \\
market\_6 & 11 & 11 & 27 & 9 & 14 \\
shaman\_3 & 10 & 11 & 22 & 8 & 10 \\
sleeping\_1 & 10 & 11 & 22 & 8 & 10 \\
sleeping\_2 & 10 & 11 & 22 & 8 & 10 \\
temple\_2 & 11 & 11 & 23 & 9 & 14 \\
temple\_3 & 8 & 10 & 16 & 7 & 11 \\
\bottomrule
\end{tabular}
\caption{Leaving out one Sintel clip at a time (means over the remaining 13). Line beats: models with a higher mean ATE than the line. Gain: models improved by half straightening. Rev.: of 55 pairwise orders, those reversed when only the weaker model is straightened.}
\label{tab:s_loo}
\end{table}

\paragraph{Straightening under other alignments and aggregates.}
With a rigid SE(3) alignment, which fits no scale and suits metric-scale models, half straightening still lowers the Sintel ATE of 6 of 11 models and the TUM ATE of none. On 9- and 15-s windows of Bonn, under Sim(3), it lowers the mean ATE of 3 and 1 of the 11 models (results/straighten\_test/straighten\_summary.csv).
\Cref{tab:s_fix} compares the arithmetic mean, the median and the geometric mean over clips for ATE and PA-ATE, \cref{tab:s_compare_full} gives all six scores of Tab.~\mref{tab:compare} on every test, including ranking reliability, and \cref{tab:s_compare_median} repeats it with medians.
\begin{table*}[t]
\centering\small
\setlength{\tabcolsep}{4pt}
\begin{tabular*}{\linewidth}{@{\extracolsep{\fill}}lccccccccccc@{}}
\toprule
& & \multicolumn{5}{c}{arithmetic mean over clips} & \multicolumn{5}{c}{geometric mean over clips} \\
\cmidrule(lr){3-7}\cmidrule(l){8-12}
Score & rot. & $>$ref & gain & rev. & split & $\tau_{\text{ref}}$ & $>$ref & gain & rev. & split & $\tau_{\text{ref}}$ \\
\midrule
ATE (positions) & $\times$ & 1 & 11 & 22 & .46 & .54 & 3 & 7 & 14 & .74 & .80 \\
rotation error & $\checkmark$ & -- & 2 & 0 & .19 & .54 & -- & 4 & 0 & .44 & .76 \\
RPE$_t$, 1 frame & $\checkmark$ & 3 & 11 & 18 & .54 & .69 & 3 & 11 & 23 & .81 & .69 \\
RPE$_t$, 10 frames & $\checkmark$ & 11 & 6 & 9 & .38 & .65 & 11 & 7 & 10 & .71 & .83 \\
DTE~\cite{lee2022ate} & $\checkmark$ & 11 & 8 & 9 & .50 & .65 & 11 & 7 & 8 & .66 & .83 \\
\textbf{PA-ATE} & $\pmb{\checkmark}$ & \textbf{11} & \textbf{8} & \textbf{10} & \textbf{.47} & \textbf{.69} & \textbf{11} & \textbf{7} & \textbf{8} & \textbf{.74} & \textbf{.83} \\
\bottomrule
\end{tabular*}
\caption{All six scores of Tab.~\mref{tab:compare} on the Sintel tests (11 models), means over the 14 clips. Rotation error: under ATE's alignment. Rot.: detects true centres rotated by 20$^\circ$ against the orientations. $>$ref: models beating every image-free reference (lines along the six camera axes and, except for DTE, a static camera); --: undefined. Gain: models improved by half straightening. Rev.: of 55 orders, those reversed when only the weaker model is straightened. Split: mean Kendall $\tau$ between the rankings on two halves of the clips (all 1,716 splits). $\tau_{\text{ref}}$: $\tau$ with the mean ATE rank on TUM and full Wild-SLAM, a reference from other data, not a true ranking. On TUM and Wild-SLAM no model gains under ATE, RPE$_t$ over 10 frames, DTE or PA-ATE.}
\label{tab:s_compare_full}
\end{table*}

A line along a model's mean optical axis, combined with that model's own orientations, ranks last under PA-ATE, RPE$_t$ over 10 frames and DTE, but second under ATE, and lowers the consecutive-frame RPE$_t$ of 7 models (results/metric\_probe/axisline\_reported\_summary.csv).
For a general prediction, one similarity jointly fits its constant-velocity component and its residual, so neither component's error need vanish.

\begin{table}[t]
\centering\small
\setlength{\tabcolsep}{5pt}
\begin{tabular}{@{}llccc@{}}
\toprule
Score & Over clips & gaining & reversals & $\tau_{\text{ref}}$ \\
\midrule
ATE            & arithmetic mean & 11 & 22 & 0.54 \\
ATE            & median & 4 & 8  & 0.80 \\
ATE            & geometric mean & 7 & 14 & 0.80 \\
PA-ATE         & arithmetic mean & 8 & 10 & 0.69 \\
PA-ATE         & median & 3 & 1  & 0.91 \\
PA-ATE         & geometric mean & 7 & 8  & 0.83 \\
\bottomrule
\end{tabular}
\caption{Sintel, eleven models, straightening at $\alpha=0.5$. Gaining: of the 11 models, those whose aggregate score improves. Reversals: of the 55 pairwise orders, those that flip when only the weaker model is straightened. $\tau_{\text{ref}}$: Kendall $\tau$ with the mean rank on TUM and full Wild-SLAM (clip-bootstrap 95\% intervals, in row order, 0.24--0.87, 0.57--0.91, 0.65--0.94, 0.35--0.91, 0.50--0.94, 0.69--0.94; paired interval of the difference under geometric means $-0.07$ to 0.19; tools/tauref\_gain\_ci.py, tools/tauref\_gmean\_ci.py).}
\label{tab:s_fix}
\end{table}

\begin{table}[t]\centering\small\setlength{\tabcolsep}{4pt}
\begin{tabular}{@{}lccccc@{}}
\toprule
Score & $>$ref & gain & rev. & split & $\tau_{\text{ref}}$ \\
\midrule
ATE (positions) & 4 & 4 & 8 & .61 & .80 \\
rotation error & -- & 3 & 0 & .06 & .65 \\
RPE$_t$, 1 frame & 3 & 11 & 18 & .67 & .69 \\
RPE$_t$, 10 frames & 11 & 3 & 4 & .68 & .91 \\
DTE & 11 & 5 & 3 & .55 & .69 \\
PA-ATE & 11 & 3 & 1 & .59 & .91 \\
\bottomrule
\end{tabular}
\caption{\Cref{tab:s_compare_full} with medians over clips (tools/comparison\_table\_median.py). The conclusions are those of the arithmetic and geometric means: the three orientation-aware scores other than the rotation error and consecutive-frame RPE$_t$ rank every image-free reference below every model, and fewer orders reverse under them than under ATE.}
\label{tab:s_compare_median}
\end{table}

\paragraph{Undefined entries of Tab.~\mref{tab:compare}.} The rotation error under a position alignment is undefined for the image-free references, whose positions do not determine the alignment's rotation, and DTE is undefined for a static camera, so its $>$ref column uses the six lines only.

\paragraph{Geometric means over clips.} Per-clip errors on Sintel span two orders of magnitude, so arithmetic means are dominated by a few clips. Split-half Kendall $\tau$ over all 1,716 splits, with the arithmetic mean, the median and the geometric mean over clips, is 0.46, 0.61 and 0.74 for ATE, 0.47, 0.59 and 0.74 for PA-ATE, and 0.38, 0.68 and 0.71 for RPE$_t$ over 10 frames. For reprojection error it is 0.45 with the arithmetic and 0.69 with the geometric mean. Under geometric means only 3 of the 11 models have a lower ATE than the line, and all 11 a lower PA-ATE than every image-free reference (tools/geomean\_checks.py). Under PA-ATE, \cref{eq:threshold} applied to the anchored errors predicts the effect of half straightening for 99\% of the Sintel pairs, 94\% of the TUM pairs and 97\% of the Wild-SLAM pairs, and every miss is a gain that it does not require (tools/snr\_rule\_anchored.py). The 10 saturated pairs, scored as a static camera, have $\mathrm{r}(\mathbf{E})=\mathrm{r}(\mathbf{G})$, so $c=1$ and $\mathrm{SNR}=1$: \cref{eq:threshold} holds with equality, and the score is unchanged, as observed. Under geometric means, half straightening lowers the PA-ATE of 7 of the 11 models on Sintel, as many as under ATE.

{\clubpenalty=10000 \widowpenalty=10000
\paragraph{Confident pairs.}
Pairs ordered in at least 97.5\% of 20,000 clip resamples (the bootstrap support of Sec.~\mref{sec:benchchange}): 26 under PA-ATE, none reversed by straightening the weaker model, and 30 under ATE, 2 reversed. At a one-sided 95\% criterion the counts are 30 (none reversed) and 32 (4 reversed) (tools/conf\_pairs\_threshold.py).\par}

\section{PA-ATE: properties and stress tests}
\label{sec:s_anchor}

\paragraph{Rotation error alone.} It orders the models with little reliability on Sintel: the Kendall $\tau$ between the rankings on two halves of the clips is 0.19, or 0.44 under geometric means.

\paragraph{Consecutive-frame RPE$_t$.} It is dominated by frame-to-frame jitter: a static camera beats 8 of the 11 models on Sintel, and half straightening lowers it for all 11 models on Sintel and 10 on TUM.

\paragraph{What PA-ATE does not change.}\label{sec:notchange}
RPE$_t$ over 10 frames, DTE and PA-ATE gain from half straightening alike, for 6 to 8 of the 11 models on the Sintel tests of Tab.~\mref{tab:compare}. Scoring orientation makes the score sensitive to inconsistent heading, not to information the clips lack.
Half straightening still lowers the PA-ATE of 8 of the 11 models, and \cref{eq:threshold}, applied to the anchored errors, predicts this for 99\% of the Sintel pairs (median SNR 1.0): it follows from the residual SNR, not from the alignment.
A motion prior can stay competitive: fully straightened, DA3-Giant still ranks third of twelve under PA-ATE and near the top under RPE$_t$ and DTE (\cref{sec:s_impl}, Straightened DA3-Giant), and a line along the optical axis on forward-driving KITTI beats 2 to 10 of the 11 models, depending on the score and window length, though every model has the lower rotation error (\cref{sec:s_kitti_dir}).

\begin{figure}[t]
\centering
\includegraphics[width=0.55\linewidth]{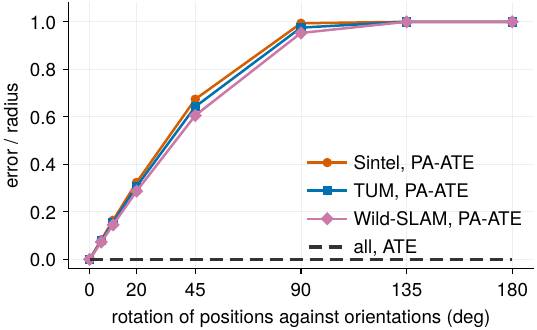}
\caption{Known-answer test on ground-truth trajectories: camera centres rotated about their centroid against the camera orientations, about a fixed axis. Median over clips and the three axes, in units of the trajectory's RMS radius.}
\label{fig:s_heading}
\end{figure}

\paragraph{Known answers.}
\Cref{fig:s_heading} rotates the true centres about their centroid against the true orientations. ATE stays at zero. PA-ATE grows with the angle (0.30, 0.63 and 0.97 of the RMS radius at 20, 45 and 90$^\circ$, medians over clips and axes) and saturates once the fitted scale reaches zero, at an angle that depends on the trajectory and the rotation axis and is 90$^\circ$ for motion perpendicular to that axis (Proposition~\ref{prop:pose_anchored_properties}).
Smooth extra bending raises both ATE and PA-ATE (results/metric\_probe/known\_answer.csv).
On the 32 test clips of Sintel, TUM and Wild-SLAM, 10$^\circ$ of per-frame orientation noise moves the anchor by at most 2.9$^\circ$ (median 1.0$^\circ$) and changes PA-ATE by a median of 1.4\% of the trajectory's RMS radius and at most 5\%. Proposition~\ref{prop:pose_anchored_properties}(f) bounds the change by $2\sin(\theta/2)$ radii for an anchor error $\theta$, and every case satisfies it (results/metric\_probe/known\_answer.csv).

\paragraph{Identifiability and clipping on real predictions.}
The anchor is unique when $\sigma_2+d\sigma_3>0$ for the singular values of $\frac1n\sum_i\mathbf{R}_i\hat{\mathbf{R}}_i^{\top}$ (Proposition~\ref{prop:pose_anchored_properties}). On all 407 model--clip pairs of Sintel, TUM, Bonn and Wild-SLAM this margin is at least 1.49 of a possible 2 (on the 1034 ScanNet pairs at least 0.63, median 2.00; results/scannet/extra\_checks.json), and the score is invariant to a common change of world frame to $10^{-14}$ (results/metric\_probe/anchor\_boundaries\_summary.csv). When the margin is zero the anchor is not unique and, by the proposition, the score can depend on which optimum is chosen. An evaluator should then flag the prediction and report the score as undefined instead of breaking the tie arbitrarily. No prediction in our runs comes close to this case.
The fitted scale is clipped at zero in 11 of the 407 pairs (10 of 154 on Sintel) and in 4 of the 1034 ScanNet pairs. Such a prediction is scored as a static camera. The 10 clipped Sintel pairs are among the 26 Sintel pairs with heading errors above 45$^\circ$. Leaving out the 10 clipped pairs, or all 26, the largest mean PA-ATE of a model is 0.268 or 0.243, against at least 0.540 for an image-free reference, and Kendall $\tau$ with the ATE order is 0.85 or 0.82 (0.85 on all pairs), so no small set of severe orientation failures decides these results.

\paragraph{Sequence length.} An error in the anchor rotation acts on the whole trajectory, so a longer path could inflate PA-ATE relative to ATE. It does not here: the median ratio of PA-ATE to ATE over the 11 models is 1.34, 1.22 and 1.12 on 9-s, 15-s and full Wild-SLAM sequences (median 32\,s) and 1.08, 1.10 and 1.05 on Bonn, and the scale is clipped at zero in 4, 1 and 1 of the Bonn pairs and in none of Wild-SLAM (tools/anchored\_length.py; results/package1/anchored\_length.csv).
\paragraph{The line's axis.} Under PA-ATE with constant orientation, the six camera-axis lines score 0.540 ($+x$), 0.543 ($+z$, forward), 0.585, 0.666, 0.686 and 0.695 on Sintel. \Cref{tab:sintel} reports the lowest, and every one lies above every model (at most 0.262).
{\clubpenalty=10000 \widowpenalty=10000
\paragraph{Orientation errors.}
The anchor is global, so time-varying orientation errors are left to the companion rotation error: on alley\_2, orientations alternating by $\pm30^\circ$ about the truth, with true positions, leave the PA-ATE at $7\times10^{-4}$ but give a rotation error of 30$^\circ$ (tools/anchor\_boundaries.py).
A slow drift of the predicted orientations is charged to the position score, as an inconsistency between heading and orientation should be: with 15$^\circ$ of added yaw drift, 6 of 11 models still beat every image-free reference on Sintel and all 11 on TUM and Wild-SLAM.
\Cref{tab:s_stress} gives the stability of the model ranking under further perturbations, for the closed-form chordal anchor we use and, as an ablation of this choice, a robust geodesic $L_1$ anchor. Under per-frame noise, a constant bias and drift the two are within 0.04 of each other, either way. The $L_1$ anchor is more stable when some frames have gross orientation errors (outliers and bursts: $\tau$ of 0.98 to 1.00 against 0.91 to 0.98), at the cost of an iterative solution. It can be substituted where such failures are expected.\par}

\begin{table}[t]\centering\footnotesize\setlength{\tabcolsep}{4pt}
\begin{tabular}{@{}lcccccc@{}}
\toprule
 & \multicolumn{2}{c}{Sintel} & \multicolumn{2}{c}{TUM} & \multicolumn{2}{c}{Wild-SLAM} \\
\cmidrule(lr){2-3} \cmidrule(lr){4-5} \cmidrule(lr){6-7}
Added orientation error & chordal & $L_1$ & chordal & $L_1$ & chordal & $L_1$ \\
\midrule
none & 1.00 & 1.00 & 1.00 & 1.00 & 1.00 & 1.00 \\
per-frame noise 5$^\circ$ & 0.98 & 0.99 & 1.00 & 1.00 & 0.99 & 0.98 \\
constant bias 5$^\circ$ & 0.96 & 0.96 & 0.89 & 0.89 & 0.89 & 0.85 \\
drift 5$^\circ$ & 0.96 & 0.96 & 0.93 & 0.93 & 0.96 & 0.93 \\
drift 15$^\circ$ & 0.93 & 0.93 & 0.85 & 0.85 & 0.85 & 0.89 \\
10\% outlier frames & 0.98 & 0.99 & 0.95 & 1.00 & 0.95 & 0.99 \\
burst of 20\% frames & 0.95 & 0.99 & 0.91 & 1.00 & 0.94 & 0.98 \\
\bottomrule
\end{tabular}
\caption{Stability of the PA-ATE ranking of the 11 models when the predicted orientations are perturbed: Kendall $\tau$ with the unperturbed ranking, mean over 5 seeds. Each perturbation multiplies the predicted orientations on the right (in the camera frame). Random rotations are the exponential map of a zero-mean isotropic Gaussian rotation vector, scaled to the stated RMS norm: per-frame noise, an independent rotation per frame of RMS norm 5$^\circ$; constant bias, 5$^\circ$ about the camera $y$-axis for every frame; drift, a rotation about the camera $y$-axis growing linearly from 0 to 5$^\circ$ or 15$^\circ$ over the clip; outlier frames, 10\% of the frames, chosen at random, each rotated by a random rotation whose RMS norm is drawn uniformly from 30$^\circ$ to 90$^\circ$ per frame; burst, one contiguous block of 20\% of the frames at a random position, all rotated by one random rotation of RMS norm 20$^\circ$. Chordal: the closed-form anchor of \cref{eq:anchor}, used throughout. $L_1$: a robust geodesic $L_1$ rotation average (iteratively reweighted) as the anchor (results/package1/stress\_summary.csv).}
\label{tab:s_stress}
\end{table}

\paragraph{Further alternatives.}
\Cref{tab:s_alt} compares four further scores on the tests of Tab.~\mref{tab:compare}. All four detect the rotated trajectory and rank every image-free reference below every model. A joint fit of positions and orientations needs a weight between position and rotation errors, and its behaviour depends on it. With the weight at 0.1, 1 and 10 times the squared RMS radius, 3, 11 and 11 models beat every image-free reference, and straightening the weaker model reverses 21, 14 and 11 of 55 orders (means over clips; results/package1/alt\_joint\_weight.csv). As the weight grows, the fitted rotation tends to the chordal mean of \cref{eq:anchor}, so PA-ATE is the joint fit's limit and needs no weight. The full-pose APE combines position and rotation units and orders the models about as unreliably as the rotation error (split-half $\tau$ 0.23 against 0.19). The first-frame alignment rests on a single pose. AUC@30 compares relative directions only and, like RPE$_t$ between consecutive frames, rewards straightening for all eleven models.

\begin{table*}[t]
\centering\small
\setlength{\tabcolsep}{4pt}
\begin{tabular*}{\linewidth}{@{\extracolsep{\fill}}lccccccccccc@{}}
\toprule
& & \multicolumn{5}{c}{arithmetic mean over clips} & \multicolumn{5}{c}{geometric mean over clips} \\
\cmidrule(lr){3-7}\cmidrule(l){8-12}
Score & rot. & $>$ref & gain & rev. & split & $\tau_{\text{ref}}$ & $>$ref & gain & rev. & split & $\tau_{\text{ref}}$ \\
\midrule
joint fit (positions + orientations) & $\checkmark$ & 11 & 8 & 14 & .46 & .65 & 11 & 7 & 9 & .72 & .87 \\
APE, full pose (evo) & $\checkmark$ & 11 & 8 & 3 & .23 & .65 & 11 & 4 & 0 & .46 & .76 \\
first-frame alignment & $\checkmark$ & 11 & 7 & 5 & .38 & .57 & 11 & 7 & 9 & .63 & .87 \\
$1-$AUC@30 (RRA/RTA) & $\checkmark$ & 11 & 11 & 11 & .60 & .76 & 11 & 11 & 15 & .73 & .76 \\
\textbf{PA-ATE (Tab.~\mref{tab:compare})} & $\pmb{\checkmark}$ & \textbf{11} & \textbf{8} & \textbf{10} & \textbf{.47} & \textbf{.69} & \textbf{11} & \textbf{7} & \textbf{8} & \textbf{.74} & \textbf{.83} \\
\bottomrule
\end{tabular*}
\caption{Further scores under the tests and column definitions of \cref{tab:s_compare_full} (Sintel, 11 models; medians in results/package1/alt\_scores\_table.csv). Joint fit: one similarity fitted to positions and orientations together, with the orientation term weighted by the squared RMS radius~\cite{salas2015alignment}; APE, full pose: the error of the whole pose after the position alignment, which mixes position and rotation units; first-frame alignment: the alignment fixed by the first pose, with a least-squares scale; AUC@30: area under the accuracy curve of the larger of the relative rotation and translation-direction errors over all frame pairs, up to 30$^\circ$~\cite{posediffusion}, reported as $1-$AUC so that lower is better (results/package1/alt\_scores\_table.csv).}
\label{tab:s_alt}
\end{table*}

\paragraph{Timing.}
The method was fixed before the run (tools/timing\_all\_scores.py), and every timed function is the released implementation, checked first against the implementations that produced the results (PA-ATE with its check that the anchor is unique). One CPU thread (Intel Xeon, 2.1\,GHz, as the virtual machine reports it; Python 3.13, NumPy 2.5, SciPy 1.18), the 154 Sintel model--clip pairs (20--50 frames, median 50), an otherwise idle machine. Five independent processes run sequentially. Each makes 30 passes over all pairs, interleaving the scores in a freshly shuffled order on every repeat. Per process and score we keep the fastest repeat and report the median over the five processes with their range (results/timing/timing\_runs.csv, timing\_all.csv, timing\_env.json). DTE runs in the first process only, with 5 repeats. Times per clip:
\begin{itemize}\itemsep0pt
\item Each position score alone: ATE 34\,\textmu s (34--35), PA-ATE 37\,\textmu s (37--40), RPE$_t$ over 1 and 10 frames 40 and 38\,\textmu s, and DTE $1.3\times10^5$\,\textmu s (our implementation, whose geometric and rotation medians are iterative).
\item With its rotation companion: ATE with the rotation error under its alignment 66\,\textmu s (65--72), PA-ATE with its rotation error 61\,\textmu s (60--65), RPE$_t$ with RPE$_r$ 68 and 66\,\textmu s over 1 and 10 frames.
\item The protocol's report (ATE, and RPE$_t$ and RPE$_r$ between consecutive frames) 106\,\textmu s, and 147\,\textmu s with PA-ATE added.
\end{itemize}
{\clubpenalty=10000 \widowpenalty=10000
On smooth synthetic trajectories of 50, 200, 1000 and 2000 frames (same protocol with three processes and 10 repeats; results/timing/timing\_length.csv), PA-ATE takes 44, 59, 149 and 300\,\textmu s and ATE 39, 54, 124 and 189\,\textmu s.
For every reported score, every process lies within 12\% of the median over processes. Within a process, the median repeat is 5 to 42\% slower than the fastest. Absolute times depend on the machine, so only times from the same run are compared.
\par}

\paragraph{Independent implementation.}
A second implementation, written by an LLM assistant from the written definitions alone, agrees with ours to $10^{-13}$ on 349 of the 352 model--clip pairs of Sintel, TUM and Wild-SLAM (3 were not supplied; results/second\_implementation).

\section{Static-point reprojection}
\label{sec:s_reproj}

\paragraph{Conventions.} With $\mathbf{X}_i$ a static point in camera $i$, $\hat{\mathbf{u}}_{ij}=\hat{\mathbf{R}}_i^{\top}(\mathbf{p}_j-\mathbf{p}_i)$ the predicted displacement in the predicted camera $i$, and $\lambda$ the scale of Sec.~\mref{sec:consequence}, the full prediction places the point in camera $j$ at $\hat{\mathbf{R}}_j^{\top}\hat{\mathbf{R}}_i(\mathbf{X}_i-\lambda\hat{\mathbf{u}}_{ij})$, and the translation-only version at $\mathbf{R}_j^{\top}\mathbf{R}_i(\mathbf{X}_i-\lambda\hat{\mathbf{u}}_{ij})$, with the true relative rotation. Both use only relative quantities, so they are invariant to a common change of the prediction's world frame. No alignment is applied to the centres. The rotated condition replaces the predicted centres by $\mathbf{Q}_{30}(\mathbf{p}_i-\bar{\mathbf{p}})+\bar{\mathbf{p}}$, with $\mathbf{Q}_{30}$ a 30$^\circ$ rotation about the $y$-axis of the prediction's world frame, and keeps $\hat{\mathbf{R}}_i$, the intrinsics, the points and $\lambda$'s definition (tools/build\_reprojection\_full.py). Given the true rotations, the line's translation alone still reprojects worst: 12.5\,px against 2.8 to 11.8\,px on Sintel, 8.3 against at most 4.2\,px on TUM and 30 against at most 13\,px on ScanNet. Clip by clip, the rotation worsens the reprojection for 85\% of the Sintel and 93\% of the TUM model--clip pairs, by a median factor of 1.53 and 1.22.

\paragraph{Moving points.} On Sintel a point is moving if its ground-truth flow differs from the camera-induced flow by more than 1\,px, as in rigidity masks for Sintel~\cite{wulff2017rigid}, or if it is occluded. On TUM, a point is removed if it falls inside the person masks of an instance segmenter~\cite{maskrcnn}, dilated by 15\,px. ScanNet's scenes are static.

\begin{figure}[t]
\centering
\includegraphics{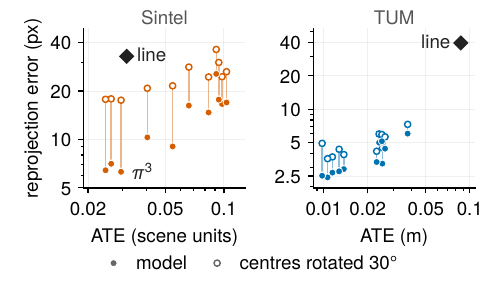}
\caption{ATE against static-point reprojection error (geometric means over clips). Filled: each model; open: its prediction with the centres rotated by 30$^\circ$ against its orientations (same ATE). The line has a low ATE on Sintel but reprojects worst of all unmodified predictions on both (arithmetic means in \cref{fig:s_reproj_mean}).}
\label{fig:s_reproj}
\end{figure}

\begin{figure}[t]\centering\includegraphics{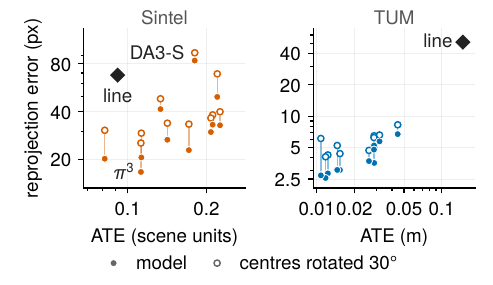}
\caption{As \cref{fig:s_reproj}, with arithmetic means over clips on both axes. The line ranks 2nd of 12 by ATE on Sintel. Only DA3-Small has a larger mean reprojection error (84\,px), set by points it places behind the camera on cave\_2. Rotating centres against orientations leaves ATE unchanged and raises the reprojection error of every model.}
\label{fig:s_reproj_mean}
\end{figure}

\paragraph{Sensitivity.}
\Cref{tab:s_reproj_sens} repeats the three checks of Tab.~\mref{tab:reproj} over 26 combinations of frame gap (1, 5, 10 on Sintel; 5, 15 on TUM), static-point filter and scale estimator (results/reprojection\_full/summary\_all.csv).
The line ranks 10th or 11th by mean reprojection error on Sintel and 12th on TUM. Rotation worsens the reprojection error of 10 or 11 models on Sintel (10 only for the depth-only filter at a 10-frame gap) and of all 11 on TUM. Full straightening improves it for 1 to 6 models on Sintel, against 10 by ATE, and for none on TUM.
\Cref{fig:s_reproj} plots ATE against the static-point reprojection error of Tab.~\mref{tab:reproj}, and \cref{fig:s_reproj_mean} repeats it with arithmetic means over clips on both axes. Under them, rotation raises the reprojection error by 5 to 20\,px on Sintel and 1 to 3\,px on TUM. With medians over clips, the line also reprojects worst on both datasets (Sintel 49\,px against 5 to 18\,px; translation only 12.7 against 2.2 to 11.5\,px on Sintel and 13.5 against at most 4.6\,px on TUM), ranks 5th of 12 by ATE, and full straightening lowers the Sintel reprojection error of 1 model. With medians, the agreement of ATE and PA-ATE with the reprojection ranking on Sintel, the line included, is 0.58 and 0.73 (paired interval of the difference including zero). With arithmetic means, the two agree with the reprojection ranking among the models alike (0.53 and 0.53 on Sintel, 0.82 and 0.89 on TUM; paired intervals including zero), and with the line included PA-ATE agrees better on Sintel (0.58 against 0.33, paired interval of the difference 0.03 to 0.45; results/reprojection\_full/tau\_bootstrap.csv).

\begin{table}[t]\centering\footnotesize\setlength{\tabcolsep}{2.5pt}
\begin{tabular}{@{}llcl ccc cc@{}}
\toprule
& & & & \multicolumn{2}{c}{line rank} & rotated & \multicolumn{2}{c}{straightened better} \\ \cmidrule(lr){5-6}\cmidrule(lr){7-7}\cmidrule(l){8-9}
Data & filter & gap & scale & ATE & reproj. & reproj.\ worse & ATE & reproj. \\
\midrule
Sintel & depthonly\_d2 & 1 & median ratio & 2 & 11 & 11 & 10 & 3 \\
Sintel & depthonly\_d2 & 5 & median ratio & 2 & 10 & 11 & 10 & 4 \\
Sintel & depthonly\_d2 & 10 & median ratio & 2 & 11 & 10 & 10 & 6 \\
Sintel & f1\_d2 & 1 & median ratio & 2 & 11 & 11 & 10 & 2 \\
Sintel & f1\_d2 & 1 & umeyama & 2 & 11 & 11 & 10 & 6 \\
Sintel & f1\_d2 & 5 & median ratio & 2 & 11 & 11 & 10 & 3 \\
Sintel & f1\_d2 & 5 & umeyama & 2 & 11 & 11 & 10 & 1 \\
Sintel & f1\_d2 & 10 & median ratio & 2 & 10 & 11 & 10 & 3 \\
Sintel & f1\_d2 & 10 & umeyama & 2 & 10 & 11 & 10 & 1 \\
Sintel & f1\_d5 & 1 & median ratio & 2 & 11 & 11 & 10 & 2 \\
Sintel & f1\_d5 & 5 & median ratio & 2 & 11 & 11 & 10 & 3 \\
Sintel & f1\_d5 & 10 & median ratio & 2 & 10 & 11 & 10 & 3 \\
Sintel & f2\_d2 & 1 & median ratio & 2 & 11 & 11 & 10 & 2 \\
Sintel & f2\_d2 & 5 & median ratio & 2 & 11 & 11 & 10 & 3 \\
Sintel & f2\_d2 & 10 & median ratio & 2 & 10 & 11 & 10 & 3 \\
Sintel & f2\_d5 & 1 & median ratio & 2 & 11 & 11 & 10 & 2 \\
Sintel & f2\_d5 & 5 & median ratio & 2 & 11 & 11 & 10 & 3 \\
Sintel & f2\_d5 & 10 & median ratio & 2 & 10 & 11 & 10 & 3 \\
TUM & depthonly\_d3 & 5 & median ratio & 12 & 12 & 11 & 0 & 0 \\
TUM & depthonly\_d3 & 15 & median ratio & 12 & 12 & 11 & 0 & 0 \\
TUM & p\_d3 & 5 & median ratio & 12 & 12 & 11 & 0 & 0 \\
TUM & p\_d3 & 5 & umeyama & 12 & 12 & 11 & 0 & 0 \\
TUM & p\_d3 & 15 & median ratio & 12 & 12 & 11 & 0 & 0 \\
TUM & p\_d3 & 15 & umeyama & 12 & 12 & 11 & 0 & 0 \\
TUM & p\_d5 & 5 & median ratio & 12 & 12 & 11 & 0 & 0 \\
TUM & p\_d5 & 15 & median ratio & 12 & 12 & 11 & 0 & 0 \\
\bottomrule
\end{tabular}
\caption{Reprojection results over all 26 settings (means over clips, full predicted poses). Filters: f1/f2, flow threshold 1 or 2\,px; d2/d5 (Sintel) and d3/d5 (TUM), depth tolerance in \%; p, TUM person masks; depthonly, depth check only. Scale: median ratio of displacements or the Umeyama scale. Rotated: models whose reprojection error rises when the centres are rotated by 30$^\circ$ against the orientations. Straightened: models improved by full straightening ($\alpha=1$).}
\label{tab:s_reproj_sens}
\end{table}


\paragraph{Coverage.}
Frame pairs with fewer than 20 static points are skipped: 0.7\% of the primary pairs on Sintel and 1.3\% on TUM. On TUM, pairs spanning a frame without a depth image within 30\,ms (frame drops of the sensor) are not scored either: 81 of the 680 pairs 5 frames apart, the same for every prediction and the line. The scale is one per clip and frame gap, and straightening here uses frame indices, whose constant-velocity fit differs from the timestamp-based one by under 1\% on TUM and not at all on Sintel and ScanNet. We use each frame's intrinsics, since cave\_2 zooms from 689 to 1529\,px. With them, the provided Sintel flow agrees with the flow induced by the provided intrinsics, poses and depth to a median below 0.04\,px on every clip (results/reprojection\_full/sintel\_flow\_consistency.csv).
The selected points keep 51\% of the grid points with valid depth on Sintel and 35\% on TUM.

\paragraph{Behind-camera penalty.} By the arithmetic mean over clips, the line's Sintel reprojection error is 68\,px against 17 to 50\,px for ten of the models and 84\,px for DA3-Small, whose mean is set by points placed behind the camera on cave\_2.
Placing a static point behind the camera is a geometric error of the prediction. The penalty only decides how it is converted to pixels, and the primary analysis uses the pre-specified $10^4$\,px (notebooks/reprojection\_penalty.ipynb; results/reprojection\_penalty). The recomputation agrees with the main run within numerical precision on all 990 rows. The angle between the true and the predicted viewing ray of each point is bounded and defined behind the camera, so it needs no penalty. By its mean over clips the line reprojects worst on Sintel in all eight combinations of gap (5 and 10 frames), filter and scale (3.98$^\circ$ against 0.67 to 2.42$^\circ$ at the primary setting), second worst at a gap of 1 frame, and worst on TUM in all eight combinations of gap (5 and 15 frames), filter and scale, under geometric means as well (tools/reprojection\_angular.py; results/reprojection\_angular). The pixel errors computed alongside equal the saved ones, with the same frame pairs, in every setting that was saved.
For the raw predictions, no prediction places a static point behind a camera on TUM. On Sintel, 1,407 of 4.2 million point evaluations (0.034\%) do, and they enter 11 of 6,828 frame-pair medians (0.16\%): 4 of DA3-Small and 5 of StreamVGGT, all on cave\_2, and 2 of the line on market\_5.
With an image-size reference penalty (the image diagonal, 1,113\,px; not an upper bound on the error, since projections can fall far outside the image) or with $10^3$\,px, DA3-Small's mean falls to 53\,px and the line ranks last of 12 by mean reprojection error on both datasets. The order of DA3-Small and the line under the arithmetic mean therefore depends on this penalty. Under geometric means it does not: the line is last under all three penalties (33\,px against at most 26, 23 and 23\,px).
Under all three penalties, rotation worsens the reprojection error of all 11 models, full and half straightening improve it for 3 and 6 models on Sintel under arithmetic means and for 0 and 3 under geometric means, and for none on TUM. With medians over clips nothing changes (results/package1/geomean\_reproj.json).

{\clubpenalty=10000 \widowpenalty=10000
\paragraph{Straightening, per pair.}
Half straightening is often genuine denoising: it lowers the translation-only reprojection error of 8 of the 11 models on Sintel. Of the 108 Sintel model--clip pairs whose ATE half straightening lowers, 53 reproject worse. Under PA-ATE, the counts are 94 and 46 (results/reprojection\_full/straighten\_quadrants.csv). These pairs share models and clips and are not independent samples.\par}

\paragraph{Reliability.} Reprojection ranks the models reproducibly on TUM and ScanNet, with a mean Kendall $\tau$ between the rankings on two halves of the clips (split-half $\tau$) of 0.89 and 0.97, and less so on Sintel (0.69 under geometric, 0.45 under arithmetic means).

\paragraph{Does a lower score mean better geometry?} Within each clip we correlate each score with the reprojection error over the 11 models' predictions (\cref{tab:s_score_geometry}; tools/score\_geometry.py). On the unmodified predictions both scores follow the geometry about equally. Adding the same predictions with their centres rotated against their orientations, and their straightened versions, separates them: PA-ATE keeps its correlation and ATE loses much of it, with intervals of the difference above zero on Sintel and ScanNet. The two other scores that use the orientations, RPE$_t$ over 10 frames and DTE, keep it as well, so the gain comes from scoring the positions relative to the orientations. On these pairs 5 frames apart, 10-frame RPE$_t$, which scores nearby frame pairs directly, correlates as closely as PA-ATE or more (0.92 against 0.82 on ScanNet with rotated and straightened predictions). PA-ATE gives the same check as one global error in ATE's form, without a choice of gap.

\begin{figure}[h]\centering\includegraphics{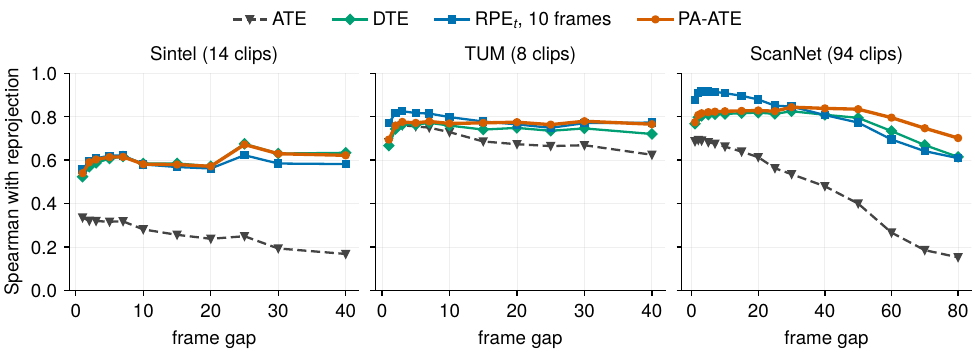}
\caption{Within-clip Spearman correlation between each score and the static-point reprojection error of the frame pairs at each gap, over the 11 models' predictions with their rotated and straightened versions (as in \cref{tab:s_score_geometry}), mean over clips. Gaps covered by fewer than half of a dataset's clips are not drawn. On TUM and ScanNet one frame gap is 0.1\,s.}
\label{fig:s_allgaps}
\end{figure}

\paragraph{Reprojection at every frame gap.} Pairs 5 frames apart measure short-range geometry. We repeated the reprojection for every frame gap and over all co-visible pairs, with the same points, scale and conditions, and at the gaps evaluated before every value equals the saved one (TUM and Sintel from the official archives, tools/reprojection\_allgaps.py; ScanNet, notebooks/scannet\_longgap.ipynb). \Cref{fig:s_allgaps} shows which score follows it at which gap. On ScanNet, 10-frame RPE$_t$ is closer up to 2.5\,s (by 0.10 at 5 frames, clip-bootstrap interval 0.08 to 0.12), the two tie at 3\,s, and from 4\,s on PA-ATE is closer, by 0.03 to 0.11 with intervals above zero (0.80 against 0.69 at 6\,s). The clips covered change with the gap. On the 53 clips covered up to 8\,s, PA-ATE is closer from 5\,s on (by 0.06 to 0.10, intervals above zero; 4\,s: $+$0.02, $-$0.01 to $+$0.04), and RPE$_t$ over as many frames as the gap follows the reprojection as closely as PA-ATE at 4 to 8\,s (differences $-$0.03 to 0.00, intervals including zero), so PA-ATE matches the gap-matched RPE$_t$ without a choice of gap (tools/reprojection\_allgaps\_fixed.py). PA-ATE is also closer from 5\,s on than 30-frame RPE$_t$ and DTE (at 6\,s by 0.05, interval 0.03 to 0.07, and 0.06, interval 0.04 to 0.08), while ATE falls from 0.68 at 5 frames to 0.15. On the unmodified predictions alone the two tie at long gaps under full reprojection (0.67 against 0.68 at 6\,s). With the true rotations (translation only), PA-ATE is closer from 2.5\,s on also for the unmodified predictions alone, by 0.02 to 0.09. On Sintel the two tie at every gap except 25 frames, where PA-ATE is closer (by 0.05, interval 0.01 to 0.10), and PA-ATE stays ahead at 30 and 40 frames (by 0.05 and 0.04) with intervals that include zero. On TUM 10-frame RPE$_t$ is closer up to 7 frames and the two tie at the longer gaps drawn. Proposition~\ref{prop:allpairs} concerns displacement errors, so this crossover is an empirical finding: RPE$_t$ at one gap scores the pairs at that gap, and PA-ATE, which equals the all-pairs error up to the rotation errors, pools every gap (tools/reprojection\_allgaps\_corr.py; results/reprojection\_allgaps).
\begin{table}[h]
\centering\small\setlength{\tabcolsep}{2.5pt}
\begin{tabular*}{\linewidth}{@{\extracolsep{\fill}}llccccc@{}}
\toprule
& & & RPE$_t$, & & & PA-ATE$-$ATE, \\
& predictions & ATE & 10 fr. & DTE & PA-ATE & 95\% \\
\midrule
Sintel & unmodified & 0.59 & 0.71 & 0.63 & 0.64 & $-0.01$ to $+0.11$ \\
 & + rotated & 0.34 & 0.65 & 0.64 & 0.63 & $+0.15$ to $+0.44$ \\
 & + rotated, straightened & 0.32 & 0.62 & 0.61 & 0.61 & $+0.16$ to $+0.45$ \\
TUM & unmodified & 0.84 & 0.87 & 0.86 & 0.86 & $-0.01$ to $+0.07$ \\
 & + rotated & 0.62 & 0.78 & 0.76 & 0.75 & $-0.06$ to $+0.35$ \\
 & + rotated, straightened & 0.76 & 0.82 & 0.76 & 0.77 & $-0.07$ to $+0.11$ \\
ScanNet & unmodified & 0.83 & 0.87 & 0.77 & 0.81 & $-0.03$ to $+0.00$ \\
 & + rotated & 0.40 & 0.89 & 0.85 & 0.85 & $+0.40$ to $+0.50$ \\
 & + rotated, straightened & 0.68 & 0.92 & 0.81 & 0.82 & $+0.10$ to $+0.17$ \\
\bottomrule
\end{tabular*}
\caption{Within-clip Spearman correlation between each score and the static-point reprojection error (pairs 5 frames apart) over the 11 models' predictions, mean over clips (Sintel 14, TUM 8, ScanNet 94). RPE$_t$ over 10 frames and DTE are the other scores of Tab.~\mref{tab:compare} that rank every image-free reference below every model. Last column: clip-bootstrap percentile interval of PA-ATE minus ATE. Clips, not predictions, are resampled, since the rotated and straightened versions of a model are not independent samples. Rotated: centres rotated by 30$^\circ$ against the orientations. Straightened: half and full straightening (tools/score\_geometry.py; results/reprojection\_full/score\_geometry.json).}
\label{tab:s_score_geometry}
\end{table}

\paragraph{Choices between learned models.} This exploratory analysis was added post hoc, after the main analysis, and was not registered. For every clip and every pair of the 11 models (raw predictions, no image-free reference), let $a$ be the model ATE ranks better and $b$ the one PA-ATE ranks better, and $\Delta=\log(e_a/e_b)$ with $e$ the reprojection error of the primary setting of Sec.~\mref{sec:consequence} (gap 5; ScanNet: its registered setting), so $\Delta>0$ favours PA-ATE and $\Delta=0$ when both pick the same model. \Cref{tab:s_choice} reports $100(e^{\bar\Delta}-1)$ for the mean over pairs within a clip and then over clips with equal weight (interval: clip bootstrap), and, pooled over all pairs where the two disagree with each pair weighted equally, the signed and absolute effect (weighting the clips with disagreements equally instead gives $-$2.5\% on TUM). Translation-only reprojection (conventions above) uses the true relative rotation and keeps the predicted displacement expressed in the predicted camera frame, so it still depends on the predicted orientation of camera $i$. Leaving out any one model family (DA3, $\pi^3$, streaming, VGGT, MapAnything) keeps the sign of the all-pairs effect under full reprojection on TUM and ScanNet, and on Sintel except without the streaming models ($-$0.02\%). The largest single-clip contributions are temple\_3 ($+$) and cave\_2 ($-$) on Sintel (tools/choice\_cost.py; results/reprojection\_full/choice\_cost.json). At 5 frames, where the local scores are the closer proxy, ATE's choices reproject 0.6\% better on ScanNet. With the reprojection error over all co-visible frame pairs instead (as in the paragraph on every frame gap), the all-pairs effect is $+$1.0\% on Sintel ($-$0.9 to $+$3.0), $+$0.6\% on TUM ($-$0.5 to $+$1.6) and $-$0.2\% on ScanNet ($-$1.0 to $+$0.6), and $-$0.2\% on ScanNet also with depth agreement at every frame. On ScanNet the interval excludes zero only at gaps of 1 to 5 frames, where the local scores are the closer proxy (\cref{fig:s_allgaps}). Leaving out one model family moves the all-pairs ScanNet effect between $-$1.1\% (without DA3) and $+$0.1\% (tools/choice\_cost\_allgaps.py; results/reprojection\_allgaps/choice\_cost\_gaps.csv).
\begin{table}[h]
\centering\small\setlength{\tabcolsep}{3pt}
\begin{tabular*}{\linewidth}{@{\extracolsep{\fill}}lccc@{}}
\toprule
 & Sintel & TUM & ScanNet \\
\midrule
model pairs (clip $\times$ pair) & 770 & 440 & 5170 \\
same choice (\%) & 88.2 & 92.7 & 91.4 \\
\multicolumn{4}{@{}l}{\emph{All pairs, full reprojection (\%)}} \\
effect & $+$0.8 & $+$0.2 & $-$0.6 \\
95\% interval & $-$1.2 to $+$2.8 & $-$0.6 to $+$1.0 & $-$1.2 to $-$0.04 \\
\multicolumn{4}{@{}l}{\emph{All pairs, translation-only (\%)}} \\
effect & $-$0.9 & $-$0.4 & $-$0.4 \\
95\% interval & $-$2.3 to $+$0.5 & $-$1.1 to $+$0.6 & $-$0.6 to $-$0.2 \\
\multicolumn{4}{@{}l}{\emph{Disagreements only, full reprojection}} \\
disagreements & 91 & 32 & 443 \\
PA-ATE better & 51 & 19 & 198 \\
signed effect (\%) & $+$6.7 & $+$2.2 & $-$6.4 \\
size, from mean $|\Delta|$ (\%) & 57.0 & 21.6 & 38.2 \\
\multicolumn{4}{@{}l}{\emph{Disagreements only, translation-only}} \\
PA-ATE better & 34 & 12 & 187 \\
signed effect (\%) & $-$7.1 & $-$4.8 & $-$4.5 \\
\bottomrule
\end{tabular*}
\caption{Reprojection error of the model each score picks, PA-ATE against ATE. Positive values favour PA-ATE. Exploratory, post hoc.}
\label{tab:s_choice}
\end{table}

\section{Direction of the line on KITTI}
\label{sec:s_kitti_dir}
\Cref{tab:kitti} gives the scores of constant-velocity lines along four camera-frame directions on the KITTI windows of Tab.~\mref{tab:survey}.
\begin{table}[h]
\centering\small
\setlength{\tabcolsep}{2pt}
\begin{tabular*}{\linewidth}{@{\extracolsep{\fill}}lcccc@{}}
\toprule
& \multicolumn{2}{c}{KITTI, 2\,s} & \multicolumn{2}{c}{KITTI, 10\,s} \\
\cmidrule(lr){2-3}\cmidrule(l){4-5}
Image-free line & ATE & PA-ATE & ATE & PA-ATE \\
\midrule
forward (optical axis)          & 0.204 & 0.278 & 2.96 & 3.26 \\
backward, sideways and up    & 0.204 & 5.73  & 2.96 & 30.2 \\
\bottomrule
\end{tabular*}
\caption{Constant-velocity lines with constant orientation, along four camera-frame directions; mean over the 22 KITTI windows of each length, in metres. For the three wrong directions the fitted scale is zero, so the score is the trajectory's RMS radius.}
\label{tab:kitti}
\end{table}

\paragraph{The forward line under every score.} \Cref{tab:kitti_scores} counts the models that beat the forward line under each score of Tab.~\mref{tab:compare}. Under every position score, ATE, PA-ATE, RPE$_t$ over 10 frames and DTE, the line beats 2 to 5 of the 11 models at 2 and 10\,s and 7 to 10 at 30 and 60\,s, where the frames are sparser. Its standing is therefore a property of the forward-driving windows, not of PA-ATE. On straight drives a constant velocity is close to the true translation. Its constant orientation is not close to the true rotations, and every model has the lower rotation error in every window length (tools/kitti\_line\_scores.py; results/kitti\_line\_scores.csv).
\begin{table}[h]
\centering\small
\setlength{\tabcolsep}{4pt}
\begin{tabular*}{\linewidth}{@{\extracolsep{\fill}}lcccc@{}}
\toprule
& \multicolumn{4}{c}{KITTI window} \\
\cmidrule(l){2-5}
Score & 2\,s & 10\,s & 30\,s & 60\,s \\
\midrule
ATE & 7 & 8 & 4 & 2 \\
RPE$_t$, 10 frames & 9 & 6 & 2 & 1 \\
DTE & 6 & 8 & 4 & 2 \\
PA-ATE & 7 & 7 & 3 & 2 \\
rotation error under the anchor & 11 & 11 & 11 & 11 \\
\midrule
line's rotation error ($^\circ$) & 3.9 & 10.3 & 29.8 & 52.8 \\
models' rotation error, median ($^\circ$) & 0.6 & 1.5 & 6.5 & 13.6 \\
\bottomrule
\end{tabular*}
\caption{Of the 11 models, those whose mean score over the KITTI windows (22 at 2 and 10\,s, 20 at 30 and 60\,s) is below that of the forward line, a constant-velocity line along the optical axis with constant orientation.}
\label{tab:kitti_scores}
\end{table}

\section{Longer KITTI windows}
\label{sec:s_kitti}
In a test registered before the runs (notes/preregistration\_v5\_kitti\_long.md), KITTI windows of 30 and 60\,s at a fixed budget of 150 frames, that is, with 2 and 4 times the frame spacing of the 10-s windows, lowered the median residual SNR from 2.09 at 10\,s to 0.95, because the models' errors grew faster than the nonlinear motion. With a fixed frame budget, as in the protocol, longer windows are necessarily sampled more sparsely. The test answers the question for that setting.
Half straightening then helps 9 and 10 of 11 models, and under PA-ATE only 3 and 2 models beat the forward line along the optical axis, against 7 at 2 and 10\,s (notes/kitti\_long\_result.md).
With a shrinkage gain chosen on the other KITTI sequences, shrinking the nonlinear component lowers the held-out PA-ATE, pooled over models and windows, by 16 to 38\% at the four window lengths, with sequence-level bootstrap intervals that exclude zero (results/metric\_probe/calibrated\_shrink\_seqcv.csv).
\paragraph{Registered test on KITTI and 7-Scenes.} The 72 windows of the held-out row of Tab.~\mref{tab:survey} (2 and 10\,s on the eleven KITTI sequences with ground truth, at 10\,Hz; 2 and 5\,s on the first sequence of each 7-Scenes scene, subsampled to 10\,Hz) were drawn from the ground truth, and the per-pair prediction of \cref{eq:threshold} with $c=0$ (the threshold $\sqrt3$) with accuracy at least 0.85 was registered, before any model was run on them (notes/preregistration\_v2\_protocol\_and\_survey.md). The registration also made benchmark-level predictions from the ground truth alone: at least 6 of the 11 models would gain on KITTI at both lengths, and at most 4 on 7-Scenes at 5\,s. The KITTI prediction missed (5 and 5 gain), and the 7-Scenes one held (none). The model-specific $S$ and $N$ give the observed counts exactly, as described next. After the models had run, and when only VGGT's results had been inspected, we added two stricter criteria: accuracy at least 0.80 on each setting, and identification of the gaining models. Both are met. A separately registered predictor that replaces the model's residual error $N$ by the clip's median scene depth times a per-model angular precision calibrated on the development data, so that it needs only the ground truth (notes/preregistration\_v4\_parallax.md), classified 195 of 242 KITTI pairs at 2\,s and 137 of 154 7-Scenes pairs at 5\,s, above its registered 0.80, but only 134 of 242 KITTI pairs at 10\,s and 106 of 154 7-Scenes pairs at 2\,s (results/parallax). The criterion therefore needs the model's own residual error, not a ground-truth proxy for it. The per-setting accuracies are in \cref{tab:s_rule_dev}. The numbers of models whose mean ATE falls under half straightening, with bootstrap 95\% intervals that resample the source sequences with all their windows, are 5 [4,7] and 5 [3,7] on KITTI at 2 and 10\,s and 0 [0,3] and 0 [0,0] on 7-Scenes at 2 and 5\,s (tools/gain\_count\_seqboot.py), each equal to the number predicted from the measured $S$ and $N$, and the predicted models are the observed ones in all 44 model--setting cases. A model is predicted to gain when the mean over its windows of the fixed-alignment ATE after half straightening, $\big(\mathrm{ATE}^2-N^2+\tfrac14(S^2+N^2)\big)^{1/2}$ per window (\cref{eq:threshold} with $c=0$), is below its mean ATE (results/fix\_analysis/survey\_pairs.csv; tools/verify\_numbers.py). The median $R^2_{\text{line}}$ of the windows is 0.999, 0.993, 0.950 and 0.935. On these windows, 7 of 11 models have positive skill over the line at 2\,s and 9 at 10\,s on KITTI, and all 11 at both lengths on 7-Scenes, with the skill $1-\sum_iw_ie_i^2/\sum_iw_i\ell_i^2$ of a model with ATE $e_i$ against the line's $\ell_i$ on window $i$, $w_i=n_i/r_i^2$, where $n_i$ is the number of frames and $r_i$ the RMS radius of the true centres in window $i$~\cite{murphy1988}.

\section{Three benchmarks registered in advance}
\label{sec:s_newbench}
\paragraph{Registration and data.} The protocols, the models, the pass mark and the predictions of Sec.~\mref{sec:newbench} were written down before any model ran on these data (notes/preregistration\_v8\_newbench.md), after only the ground truth had been read. \emph{Lightspeed} (DynPose-100K~\cite{dynpose100k}): the 36 clips at all frames of frames-24fps (30 to 150 frames), with poses.pkl read as world-to-camera, as the dataset's documentation states and its own visualisation script implements. Read the other way, the line's ATE would be 0.309 instead of 0.099, and the centre tracks would be 8.8 times rougher (median over clips of the RMS third difference over the RMS first difference; tools/lightspeed\_gt\_check.py). \emph{DyCheck}~\cite{dycheck}: MegaSaM's copy~\cite{megasam}, camera-0 frames of each scene's dense/images with the cameras of dense/dycheck/camera, which equal the centres and rotations of the COLMAP model that MegaSaM evaluates (to $10^{-15}$ on haru-sit), every 10th frame from the first, as Easi3R~\cite{easi3r}; all 12 scenes of MegaSaM's list. \emph{ADT}~\cite{adt}: Easi3R's 8 TAPVid-3D minival files~\cite{tapvid3d}, every 5th of the 300 frames, with their world-to-camera extrinsics. Models: the eight offline models of Tab.~\mref{tab:sintel} and VGGT-$\Omega$~\cite{vggtomega} (the checkpoint its authors designate for benchmarking, at the setting of its reproduction notes), official code and the wrappers of the other runs (notebooks/newbench\_offline.ipynb). The streaming models were not run. No pair failed. \Cref{tab:s_newbench} gives the scores (tools/newbench\_analysis.py; results/newbench\_offline).

\begin{table}[h]
\centering\small
\setlength{\tabcolsep}{3pt}
\begin{tabular*}{\linewidth}{@{\extracolsep{\fill}}lccccccc@{}}
\toprule
& \multicolumn{3}{c}{Lightspeed (36 clips)} & \multicolumn{2}{c}{DyCheck (12)} & \multicolumn{2}{c}{ADT (8)} \\
\cmidrule(lr){2-4}\cmidrule(lr){5-6}\cmidrule(l){7-8}
Model & ATE & half & PA-ATE$^*$ & ATE & PA-ATE & ATE & PA-ATE \\
\midrule
VGGT-$\Omega$ & 0.028 & 0.056 & 0.011 & 0.005 & 0.005 & 0.010 & 0.011 \\
$\pi^3$X & 0.040 & 0.060 & 0.015 & 0.009 & 0.010 & 0.010 & 0.011 \\
$\pi^3$ & 0.044 & 0.058 & 0.018 & 0.010 & 0.012 & 0.014 & 0.015 \\
DA3-Giant & 0.050 & 0.067 & 0.012 & 0.005 & 0.006 & 0.008 & 0.008 \\
DA3-Large & 0.144 & 0.112 & 0.044 & 0.009 & 0.010 & 0.055 & 0.065 \\
MapAnything & 0.164 & 0.118 & 0.094 & 0.024 & 0.026 & 0.043 & 0.047 \\
VGGT & 0.223 & 0.171 & 0.074 & 0.012 & 0.014 & 0.064 & 0.069 \\
DA3-Base & 0.229 & 0.168 & 0.109 & 0.021 & 0.023 & 0.139 & 0.157 \\
DA3-Small & 0.236 & 0.154 & 0.161 & 0.027 & 0.030 & 0.098 & 0.125 \\
\midrule
Line / best image-free ref. & 0.099 & -- & 0.096 & 0.135 & 0.138 & 0.155 & 0.273 \\
\bottomrule
\end{tabular*}
\caption{The three benchmarks registered before any model ran (\cref{sec:s_newbench}): mean ATE, for Lightspeed also after half straightening, and PA-ATE for the eight offline models and VGGT-$\Omega$. Last row: the image-free line under ATE and the best of the image-free references (six camera-axis lines, static camera) under PA-ATE. $^*$Lightspeed PA-ATE over the 11 clips whose ground-truth orientations agree with the models (\cref{sec:s_lightspeed_gt}); there the line's ATE is 0.038. Lightspeed and DyCheck in their units, ADT in metres (tools/build\_newbench\_supp.py).}
\label{tab:s_newbench}
\end{table}
 The registration quoted DyCheck's median $R^2_{\text{line}}$ as 0.17, from 11 scenes at every 10th frame; the 12 scenes evaluated give 0.14 (Tab.~\mref{tab:deletion}).

\paragraph{Outcomes.} (N1) The criterion with the measured $c$ gives the outcome after refitting for 99.7, 100 and 90.3\% of the Lightspeed, DyCheck and ADT pairs, $c=0$ for 95.1, 99.1 and 91.7\%, and with the refit term for all 504 pairs, also at $\alpha=0.25$, 0.75 and 1 (registered pass mark: 85\% for both forms). (N2) On Lightspeed 5 of the 9 models gain from half straightening and the line beats 5 of them, as registered. On DyCheck no model gains and every model beats the line, as registered. On ADT every model beats the line, but two gain (DA3-Small and DA3-Base, the two with the largest ATE), where the registration said none: the registered statement was made from the ground truth alone, whereas the criterion, applied per pair with each model's error, gives both. (N3) Kendall $\tau$ between the ATE and PA-ATE orders of the 9 models, both over the same clips, is 0.94, 0.94 and 1.00 (clip-bootstrap intervals 0.72 to 1, 0.89 to 1 and 0.89 to 1). On Lightspeed this uses all 36 clips, whose stored orientations enter PA-ATE. Over the 11 consistent clips alone it is also 0.94 (0.83 to 1). The median residual SNR is 1.3, 14 and 6.7.

\paragraph{Published values.} With the line at 0.099, of the 11 methods in Tab.~2 of DynPose-100K~\cite{dynpose100k} and in SpatialTrackerV2~\cite{spatrackerv2}, which uses the same values and protocol, only the DynPose-100K pipeline (0.072) is below it, counting each method's best variant: DROID-SLAM 0.198, DUSt3R 0.412, MonST3R 0.149, LEAP-VO 0.206, COLMAP 0.323 (with masks), ParticleSfM 0.185, CUT3R 0.274, VGGT 0.226, MegaSaM 0.105 and SpatialTrackerV2 0.134. A third table~\cite{dynapo}, whose units and subset are not stated, reports MegaSaM at 0.073 and bundle-adjusted variants of several methods below 0.099. On DyCheck under MegaSaM's protocol (all frames, ground truth divided by the first-to-last camera distance) the line is 0.97 on the 11 scenes whose cameras we read first, against 0.020 to 0.690 published, and on ADT 0.155 against 0.039 to 0.084 in Easi3R's table: every published method beats it there (tools/line\_dycheck.py, tools/line\_other\_benchmarks.py).

\paragraph{Lightspeed ground-truth orientations.}\label{sec:s_lightspeed_gt} On Lightspeed every model has an anchored rotation error of 8 to 14$^\circ$, against at most 2.4$^\circ$ on DyCheck and ADT. The stored rotations are proper (orthonormal to $10^{-7}$, determinant $+1$), and of the 24 proper signed permutations of the camera axes, with or without transposing, the reading used gives the smallest relative-rotation error against $\pi^3$ (1.9$^\circ$ over 5 frames, next best 3.0$^\circ$; tools/lightspeed\_gt\_check.py, results/lightspeed\_arbiter/gt\_check.json). Relative rotations over a few frames separate the clips into classes (notes/lightspeed\_gt\_classes.csv). In 11 clips the ground truth agrees with all nine models (at 3 frames 0.16 to 1.4$^\circ$, the level at which the models agree with one another). In 11 clips the ground-truth relative poses, rotation and translation, are the mirror image of the models' under a reflection of the camera frame (vertical in 6, horizontal in 3, diagonal in 2), and mirroring them back lowers the anchored rotation error of seven models from about 15$^\circ$ to 1 to 3$^\circ$ (gt\_check.json). In 3 clips the reflection that reconciles the relative rotations with the models does not reconcile the relative translations. In 6 more the rotations are mirrored and the motion is too small to decide the translations, and 5 clips are nearly static. Each type of reflection occurs with a single roll of the clip's first ground-truth rotation about the optical axis (vertical with 180$^\circ$, horizontal with 0$^\circ$, diagonal with $\pm$90$^\circ$; gt\_check.json). A time offset, an Euler-angle order, a fixed axis convention or rolling shutter cannot produce these per-clip sign flips. To decide between the ground truth and the models without any network, we estimated relative rotations from the frames alone, with SIFT matches, a five-point RANSAC essential matrix and cheirality (OpenCV; per clip, the focal length from a grid that maximises the inliers; pairs 5 and 10 frames apart with at least 50 inliers; notebooks/lightspeed\_arbiter.ipynb, tools/lightspeed\_arbiter\_analysis.py). On the 9 consistent clips with such pairs, these rotations lie a median of 1.0$^\circ$ from both the ground truth and the four best models. On the 19 clips with mirrored rotations, they lie 4.9$^\circ$ from the ground truth and 1.6$^\circ$ from the models, the ground truth is further away by more than 0.5$^\circ$ on 15 clips and the models on 1, and regressing the ground truth's rotation components on the image-based ones gives slopes near $-1$ on the reflected axes (median $-0.98$ for roll), against near $+1$ for the models. A reflection of the frames and a reflection of the stored poses produce the same disagreement, so no measurement on these data can say which of the two carries it. For evaluation the consequence is the same either way: the stored poses and the frames disagree. Stored orientations do not enter ATE, and with the released centres our VGGT value (0.223) reproduces the published one (0.226). On the 11 consistent clips, PA-ATE ranks the best image-free reference (0.096) below 7 of the 9 models. The other two, DA3-Base and DA3-Small, have the largest ATE there (0.080 and 0.106, against 0.038 for the line), and the four best models have anchored rotation errors of 0.9 to 1.5$^\circ$. We therefore report orientation-aware scores on Lightspeed only for these 11 clips.

\section{ScanNet}
\label{sec:s_scannet}
ScanNet~\cite{dai2017scannet} is the third pose benchmark of the papers that use the Sintel and TUM protocol~\cite{monst3r,cut3r}. We obtained the official ScanNet v2 release under its terms of use and follow CUT3R's preprocessing~\cite{cut3r}: the first 90 frames at stride 3 of each of the 100 test scenes (scene0707\_00 to scene0806\_00). Six scenes (0711, 0737, 0750, 0761, 0791, 0796) are dropped because one of their 90 ground-truth poses is not finite, the only exclusion rule we registered. All 11 models ran on all 94 clips with their official code (1034 model--clip pairs). Our CUT3R scores (ATE 0.097, RPE$_t$ 0.023, RPE$_r$ 0.598) are close to the published 0.099, 0.022 and 0.600~\cite{cut3r}. \Cref{tab:s_scannet} lists every registered score. The four later methods of Tab.~\mref{tab:sintel} ran afterwards on the same 94 clips in the configurations of that table (mean ATE: UniQuery4R 0.034, LASER 0.031, SpatialTrackerV2 0.053, AnyCam 0.106). LASER's value equals the one its paper reports~\cite{laser}, and the other three papers report no ScanNet pose. AnyCam, which optimises at test time, gave 0.1057 in each of two independent runs. MonST3R, run last on the same clips (two halves of 47 on two GPUs of the same type), scores 0.057 (published 0.068~\cite{monst3r}; 0.077 in CUT3R's re-run~\cite{cut3r}). These 470 pairs and the 154 Sintel and TUM pairs of the five methods enter the added-methods row of \cref{tab:survey}.

The analysis was registered before any ScanNet data were accessed (notes/preregistration\_v7\_scannet.md), with an addendum, written before any prediction existed, that requires an accuracy of at least 0.85 from both forms of \cref{eq:threshold}. Both pass: the criterion with the measured $c$ gives the outcome for 97.3\% of the pairs and $c=0$ for 95.5\%. Gains are rare on ScanNet (56 of 1034 pairs, 5\%), so the test is mainly one of false certification: the criterion certifies 28 gains and no loss, and the gains it misses are typically smaller (median ATE decrease 13.3\%, against 23.2\% for those it certifies). On this set, always predicting no gain is right for 94.6\% of the pairs, and the straightness and error thresholds of \cref{tab:survey} for 93.5\% and 93.6\%. The held-out windows of KITTI and 7-Scenes, where gains are common, are the test that separates the criterion from these baselines.
ScanNet's hand-held indoor paths are far from straight (median $R^2_{\text{line}}$ 0.67; median residual SNR 5.7), and it behaves like TUM: the line ranks last of the 12 under ATE (0.232) and under PA-ATE, half straightening raises the mean ATE of all 11 models (median ratio 2.25; also under geometric means), and ATE and PA-ATE order the models identically (Kendall $\tau$ 1.00, clip-bootstrap 95\% interval 0.93--1.00). DTE also ranks the best of the six axis lines (0.274) below every model (at most 0.102). The registered reprojection test uses ScanNet's depth camera and its intrinsics (its colour and depth extrinsics are the identity in every scene), a $40\times30$ grid and the registered depth-agreement filter (3\% at the target frame), with the settings fixed in a second addendum before the run. The ground-truth poses reproject the selected points to within $2\times10^{-4}$\,px, and a pair 5 frames apart keeps a median of 783 points (minimum 273, so no such pair is skipped). Under geometric means over clips, the line places points worst (81\,px against 3 to 11\,px; 30 against at most 13\,px with the true rotations), rotating the centres by 30$^\circ$ raises the error of all 11 models by 7 to 11\,px, full or half straightening lowers it for none, and reprojection orders the models like ATE and PA-ATE (Kendall $\tau$ 0.93 for both; split-half $\tau$ of reprojection 0.97). Arithmetic means, 15-frame gaps and the filter applied at every intermediate frame give the same counts (results/scannet\_reproj/summary.json).
\begin{table*}[t]
\centering\small
\setlength{\tabcolsep}{4pt}
\begin{tabular*}{\linewidth}{@{\extracolsep{\fill}}lcccccccccc@{}}
\toprule
& \multicolumn{2}{c}{TUM (8 clips)} & \multicolumn{8}{c}{ScanNet (94 clips)} \\
\cmidrule(lr){2-3}\cmidrule(l){4-11}
Method & ATE$\downarrow$ & PA-ATE$\downarrow$ & ATE$\downarrow$ & median$\downarrow$ & PA-ATE$\downarrow$ & rot.$\downarrow$ & RPE$_t\downarrow$ & RPE$_r\downarrow$ & DTE$\downarrow$ & cl. \\
\midrule
$\pi^3$ & 0.015 & 0.016 & 0.030 & 0.027 & 0.038 & 1.15 & 0.0124 & 0.345 & 0.038 & 0 \\
Line (no images) & 0.148 & 0.162 & 0.232 & 0.210 & 0.364 & 27.84 & 0.0306 & 1.869 & 0.274 & -- \\
DA3-Giant & 0.012 & 0.014 & 0.032 & 0.029 & 0.041 & 1.19 & 0.0128 & 0.362 & 0.042 & 0 \\
$\pi^3$X & 0.011 & 0.013 & 0.030 & 0.027 & 0.038 & 1.12 & 0.0124 & 0.345 & 0.037 & 0 \\
DA3-Base & 0.029 & 0.032 & 0.059 & 0.047 & 0.076 & 3.31 & 0.0233 & 1.152 & 0.076 & 5 \\
DA3-Large & 0.015 & 0.017 & 0.035 & 0.031 & 0.044 & 1.53 & 0.0162 & 0.644 & 0.040 & 1 \\
VGGT & 0.012 & 0.013 & 0.035 & 0.030 & 0.043 & 1.15 & 0.0152 & 0.384 & 0.043 & 0 \\
DA3-Small & 0.032 & 0.036 & 0.073 & 0.058 & 0.092 & 3.83 & 0.0339 & 1.722 & 0.086 & 10 \\
TTT3R & 0.029 & 0.033 & 0.064 & 0.052 & 0.084 & 2.04 & 0.0207 & 0.570 & 0.072 & 9 \\
CUT3R & 0.044 & 0.051 & 0.097 & 0.076 & 0.123 & 3.37 & 0.0230 & 0.598 & 0.102 & 25 \\
StreamVGGT & 0.026 & 0.031 & 0.048 & 0.043 & 0.058 & 1.40 & 0.0191 & 0.487 & 0.052 & 1 \\
MapAnything & 0.029 & 0.031 & 0.053 & 0.047 & 0.066 & 2.27 & 0.0250 & 0.775 & 0.058 & 5 \\
\bottomrule
\end{tabular*}
\caption{TUM and ScanNet, official protocol; means over clips unless marked. PA-ATE: pose-anchored ATE; rot.: its rotation error (degrees); RPE between consecutive frames (RPE$_r$ in degrees); DTE~\cite{lee2022ate}. For the line: constant orientation and, for PA-ATE, RPE and DTE, the camera-axis direction, of six, with the lowest mean PA-ATE (for DTE, its own lowest). Its rotation error, under the orientation anchor, is that of a constant orientation, and straightening leaves it unchanged. Cl.: ScanNet clips on which half straightening lowers the ATE. ScanNet: first 90 frames at stride 3 of the 100 test scenes, 6 with invalid ground-truth poses dropped.}
\label{tab:s_scannet}
\end{table*}

\section{Beyond camera poses: the alignment orbit of a template}
\label{sec:s_orbit}
This section states the two properties of ATE for any score that minimises an error over an alignment, marks where they stop holding, and collects input-free templates across tasks.

\paragraph{Setting.} A score of this kind aligns a prediction $\mathbf p\in\mathbb R^m$ to the truth $\mathbf y$ over a set $\mathcal G$ of admissible transforms and reports $E_{\mathcal G}(\mathbf p,\mathbf y)=\inf_{g\in\mathcal G}\|\mathbf y-g\cdot\mathbf p\|/\sqrt m$. For ATE, $\mathbf p$ stacks the camera centres and $\mathcal G=\mathrm{Sim}(3)$. For video depth evaluated with a per-sequence scale and shift, $\mathbf p$ stacks the depths of all valid pixels of a sequence and $\mathcal G=\{\mathbf p\mapsto s\mathbf p+b\mathbf 1\}$.

\paragraph{Invariance.} If $\mathcal G$ is a group, $E_{\mathcal G}(h\cdot\mathbf p,\mathbf y)=E_{\mathcal G}(\mathbf p,\mathbf y)$ for every $h\in\mathcal G$, because $g\mapsto gh$ maps $\mathcal G$ onto itself. Whatever a transform in $\mathcal G$ changes in a prediction is therefore never scored. Rotating ATE's positions against their orientations is the instance of \cref{sec:supplies}. The prediction's other parts, such as the orientations, matter only if the score reads them.

\paragraph{Proposition 6 (orbit of a template).} \emph{Let $\mathbf t$ be a prediction that does not depend on the input. If the closure of its orbit $\mathcal G\cdot\mathbf t$ is an affine subspace $\mathbf a+\mathcal V$, then $E_{\mathcal G}(\mathbf t,\mathbf y)^2=\|(\mathbf I-\mathbf P_{\mathcal V})(\mathbf y-\mathbf a)\|^2/m$, with $\mathbf P_{\mathcal V}$ the orthogonal projection onto $\mathcal V$.}
\emph{Proof.} The distance to a set equals the distance to its closure, and the point of $\mathbf a+\mathcal V$ nearest to $\mathbf y$ is $\mathbf a+\mathbf P_{\mathcal V}(\mathbf y-\mathbf a)$. $\square$

The constant-velocity line is an instance: $\mathrm{Sim}(3)$ maps a moving line onto every moving line, whose closure is the space of paths affine in time, and Proposition~\ref{prop:constant_velocity_ate} follows. So is a non-constant depth template $\mathbf t$ under scale and shift, whose orbit closure is $\mathrm{span}\{\mathbf t,\mathbf 1\}$ when $s$ may take either sign (with $s>0$ only, the least-squares slope must also be positive). The condition fails for a rotation alone, whose orbit of $\mathbf t=(1,0)$ is the unit circle: for $\mathbf y=\mathbf 0$ the distance is 1 while the projection onto the span of the circle gives 0. It also fails for alignments that rematch correspondences, such as ICP. Where it holds, the template's score is the ground truth's distance to a known subspace, and $R^2_{\mathbf t}=1-\|(\mathbf I-\mathbf P_{\mathcal V})\mathbf y\|^2/\|(\mathbf I-\mathbf P_{\mathcal C})\mathbf y\|^2$, with $\mathcal C$ the constants, measures how much of a benchmark the template explains before any model runs. This is $R^2_{\text{line}}$ of \cref{sec:diagnosis} for poses. A constant template has $R^2=0$ by construction, so its relative distance $\|(\mathbf I-\mathbf P_{\mathcal V})\mathbf y\|/\|\mathbf y\|$ is the quantity to report.

\paragraph{Proposition 7 (the straightening test for any template subspace).} \emph{Let $\boldsymbol\Pi$ project along the samples (frames or pixels) onto a subspace $\mathcal V$ that contains the constants, write $\mathrm{cv}=\boldsymbol\Pi$ and $\mathrm r=\mathbf I-\boldsymbol\Pi$, and let the fixed alignment act on the values, $F(\mathbf X)=\mathbf X\mathbf L^\top+\mathbf 1\mathbf b^\top$. Then \cref{eq:threshold}, with $S$, $N$ and $c$ defined as in \cref{eq:snr} for this $\mathrm r$, is the exact condition for straightening $\mathbf P_\alpha=\mathrm{cv}(\mathbf P)+(1-\alpha)\,\mathrm r(\mathbf P)$ to lower the error at the fixed alignment.}
\emph{Proof.} The proof of Proposition~\ref{prop:straightening_threshold} uses $\mathbf H$ only through $\mathbf H\mathbf 1=\mathbf 1$ and its action on the sample index, so it holds verbatim with $\boldsymbol\Pi$ in place of $\mathbf H$. $\square$

Scale and shift of a depth map satisfy the condition ($\mathbf L=s$, $\mathbf b=b$). The test then speaks about the fixed alignment only. When $\mathcal G$ contains the straightening itself, re-aligning undoes it: moving a depth map towards its mean, $\bar{\mathbf p}\mathbf 1+(1-\alpha)(\mathbf p-\bar{\mathbf p}\mathbf 1)$, is a scale and shift of $\mathbf p$, so the score after re-alignment does not change. For ATE the straightened path is not a similarity transform of the prediction, and the refit term of \cref{sec:cause} is needed precisely because re-alignment moves the score.

\paragraph{Proposition 8 (anchoring cannot flatter).} \emph{If an anchored score chooses $g_A\in\mathcal G$ from information the task must recover and reports the same error at $g_A$, it is never below $E_{\mathcal G}$.} \emph{Proof.} $g_A$ is one of the transforms over which $E_{\mathcal G}$ takes the infimum. $\square$
PA-ATE $\ge$ ATE is this statement with the rotation taken from the orientations. Sharing one transform across all frames likewise never lowers the loss that the alignment minimises, compared with one transform per frame. This holds for the minimised loss only: when the fit minimises one loss and the benchmark reports another (AbsRel or $\delta$ after an $L_1$ fit, say), no order is guaranteed.

\paragraph{Evidence across tasks.} \Cref{tab:s_orbit_ladder}, summarised in Tab.~\mref{tab:templates}, collects input-free templates scored under each benchmark's own alignment, where the alignment is strong and the target varies little beyond it. Counts depend on which methods a table lists, so they compare benchmarks only roughly. Where much of the target lies outside the orbit, templates do not compete, as for a mean pose (1 of 7 methods in SMPLy's Tab.~4~\cite{leroy2020smply}) or a mean hand (none of four in FreiHAND's Tab.~3~\cite{freihand}).

\begin{table}[t]
\centering\small
\setlength{\tabcolsep}{3pt}
\begin{tabular}{@{}lllll@{}}
\toprule
Benchmark & Alignment & Input-free template & Score & Beats \\
\midrule
Sintel camera poses (14 clips) & Sim(3) of the positions & constant-velocity line & ATE 0.092 & 10 of 11 \\
Lightspeed camera poses & Sim(3) of the positions & constant-velocity line & ATE 0.099 & 5 of 9 \\
SSP-3D body shape$^*$ & centroid and RMS radius & SMPL mean shape & 18.6\,mm & 5 of 9 \\
NoW faces~\cite{nowleaderboard} & similarity, refined on the scan & FLAME 2020 template & 1.21\,mm & 8 of 21, 1 tie \\
NoW-Metric faces~\cite{mica} & rigid, refined on the scan & FLAME 2020 template & 1.49\,mm & 9 of 14 \\
KITTI video depth$^*$ & scale and shift per sequence & mean layout of the other sequences & 0.191, 71.2 & 1 of 5 \\
KITTI single-image depth$^*$ & scale and shift per image & mean layout of the other drives & 20.4, 67.5 & 1 of 8 \\
Bonn video depth$^*$ & scale and shift per sequence & mean layout of the other sequences & 0.326, 38.8 & 0 of 5 \\
Sintel video depth$^*$ & scale and shift per sequence & mean layout of the other sequences & 0.481, 32.8 & 0 of 5 \\
\bottomrule
\end{tabular}
\caption{Input-free templates scored under each benchmark's own alignment, grouped by task. Score: the benchmark's error (video depth: AbsRel and $\delta{<}1.25$ in \%). Beats: methods of the source table or leaderboard with a worse error (depth: worse on both metrics, in DepthCrafter's Tab.~1 for video, same code, and Marigold's Tab.~1 for single images). Camera poses: Tab.~\mref{tab:sintel} and Sec.~\mref{sec:newbench}. SSP-3D: methods of HuManiFlow's Tab.~4~\cite{humaniflow}. NoW: median errors, leaderboards accessed 11 November 2026. $^*$Computed by us from ground truth alone with the official metric code.}
\label{tab:s_orbit_ladder}
\end{table}

\paragraph{Body shape on SSP-3D.} SSP-3D has 311 images of 62 athletes with SMPL shapes~\cite{straps}. Its metric, PVE-T-SC, poses prediction and target in a T-pose with the target's gendered SMPL model, moves the prediction to the target's centroid and RMS radius, and averages the per-vertex distance. This normalisation is not a minimisation, so Proposition~6 gives no closed form, but a template's score still depends on the ground truth alone. The SMPL mean shape (all shape coefficients zero) is the same prediction for every image and uses no label. The gender enters through the metric, as for every method. It scores 18.6\,mm (95\% interval 15.3 to 22.1 when the athletes are resampled, unpaired and so conservative). Of the nine single-image methods in HuManiFlow's Tab.~4~\cite{humaniflow} (SPIN 22.2, PARE 21.7, HybrIK 22.9, STRAPS 15.9, 3D Multibodies 22.3, Sengupta et al.\ 15.2, ProHMR 22.2, HierProbHuman 13.6, HuManiFlow 13.5), it beats the five general pose-and-shape regressors and none of the four methods trained for shape (STRAPS and its three successors). Write $\mathbf g$ and $\mathbf d$ for the truth's and a method's departures from the mean shape. In squared error the method beats the mean exactly when $2\langle\mathbf g,\mathbf d\rangle>\|\mathbf d\|^2$, that is, when its departure follows the truth's by more than half its own size. This is the straightening test at full strength: replacing an estimate by the mean lowers the error when the estimate's departure is mostly noise. The regressors lack training data with varied body shapes~\cite{straps}, and the mean shape beats all five. The mean coefficients of the other athletes of the same gender, which uses the gender label, score 14.8\,mm and beat seven. We used the SMPL v1.1.0 male and female models (tools/build\_template\_shape.py; results/template\_shape).

\paragraph{Video depth.} Two evaluation codes are in use (\cref{tab:s_depth_templates}), both with one scale and shift per sequence. DepthCrafter's fits them by least squares in disparity and produced the NVDS, Marigold, Depth-Anything-V2, ChronoDepth and DepthCrafter rows that later tables, MonST3R's among them, reprint~\cite{depthcrafter,monst3r}. MonST3R's code, also used by CUT3R and TTT3R~\cite{cut3r,ttt3r}, fits an $L_1$ loss in depth. We ran both unchanged on three templates that never see an image and are the same for every frame: a constant, a row ramp whose disparity grows linearly down the image as for flat ground, and the mean layout, the median-normalised disparity averaged over the other test sequences of the dataset. DepthCrafter's fit is the orthogonal projection of Proposition~6, and the two agree to $10^{-9}$ on every sequence. On Sintel and Bonn every template explains at most 9\% of the disparity variance (median $R^2$) and beats no published row on both AbsRel and $\delta{<}1.25$ under either code. On KITTI the mean layout explains 75\% and beats NVDS under DepthCrafter's code (0.191 and 71.2 against 0.253 and 58.8), and NVDS and CasualSAM under MonST3R's. The label-free ramp explains 55\% and beats none. Single images get one scale and shift each in Marigold's protocol, fitted in depth for methods that predict depth and in disparity for those that predict disparity, such as MiDaS~\cite{marigold}. Scored with Marigold's own alignment and metric functions, which reproduce its evaluation script on the depth-space run, the mean layout of the other drives explains 77\% of an image's disparity variance on the 652 KITTI Eigen test images and, as disparity, scores 20.4 and 67.5, better than MiDaS (23.6, 63.0) on both metrics and worse than the other seven rows of Marigold's Tab.~1. As depth it scores 24.8 and 64.2. On saved Sintel predictions our use of MonST3R's code reproduces the official script's output to within 0.004 (tools/build\_template\_depth.py, tools/build\_template\_depth\_dc.py, tools/build\_template\_mono.py; results/template\_depth, results/template\_depth\_dc, results/template\_mono).

\begin{table}[t]
\centering\small
\setlength{\tabcolsep}{4pt}
\begin{tabular}{@{}llrrrrrrr@{}}
\toprule
& & & \multicolumn{3}{c}{DepthCrafter's code} & \multicolumn{3}{c}{MonST3R's code} \\
\cmidrule(lr){4-6}\cmidrule(l){7-9}
Dataset & Template & $R^2$ & AbsRel & $\delta{<}1.25$ & beats & AbsRel & $\delta{<}1.25$ & beats \\
\midrule
Sintel (23) & constant & 0.00 & 0.504 & 29.4 & 0 of 5 & 0.635 & 44.1 & 0 of 8 \\
 & row ramp & 0.07 & 0.484 & 32.9 & 0 of 5 & 1.592 & 24.3 & 0 of 8 \\
 & mean layout & 0.09 & 0.481 & 32.8 & 0 of 5 & 0.617 & 40.7 & 0 of 8 \\
\midrule
Bonn (5) & constant & 0.00 & 0.342 & 34.7 & 0 of 5 & 0.416 & 59.3 & 0 of 7 \\
 & row ramp & 0.06 & 0.319 & 40.6 & 0 of 5 & 1.584 & 19.9 & 0 of 7 \\
 & mean layout & 0.04 & 0.326 & 38.8 & 0 of 5 & 0.399 & 61.1 & 0 of 7 \\
\midrule
KITTI (13) & constant & 0.00 & 0.391 & 31.6 & 0 of 5 & 0.406 & 30.7 & 0 of 7 \\
 & row ramp & 0.55 & 0.310 & 56.2 & 0 of 5 & 0.304 & 50.3 & 0 of 7 \\
 & mean layout & 0.75 & 0.191 & 71.2 & 1 of 5 & 0.188 & 72.0 & 2 of 7 \\
\bottomrule
\end{tabular}
\caption{Image-free depth templates under the two video-depth evaluation codes in use, each with one scale and shift per sequence. $R^2$: median over sequences of the share of the true disparity's variance that the template's orbit explains, which is exactly DepthCrafter's least-squares fit (Proposition~6). Beats: published rows of the same code family that the template beats on both AbsRel and $\delta{<}1.25$ (DepthCrafter's Tab.~1, MonST3R's Tab.~2). MonST3R's $L_1$ fit runs 1000 Adam steps and does not always converge, as for the ramp on Sintel (tools/template\_tables.py).}
\label{tab:s_depth_templates}
\end{table}

\paragraph{The same choice in other tasks.} Other fields have restricted their alignments for the reason of Proposition~8. SI-SDR replaced the 512-tap distortion filter of BSS-eval with a single scale, because the filter let a badly distorted estimate score well~\cite{leroux2019sdr}. Procrustes-aligned joint errors hide where a body stands in the world~\cite{drazic2024procrustes}. Affine-aligned depth metrics are blind to affine errors, and the choice of alignment changes the ranking of depth models~\cite{benchdepth,wu2025deptheval}. Video depth fits one scale and shift per video rather than per frame, so that scale drift counts~\cite{depthcrafter,videodepthanything}. NoW-Metric drops the scale from NoW's alignment, so that a face's metric size counts~\cite{mica}. Each keeps in the alignment only what the task may ignore. Restricting the alignment does not by itself defeat a template: under NoW-Metric's rigid alignment the mean face still scores 1.49\,mm and beats 9 of 14 entries (\cref{tab:s_orbit_ladder}), a score set by the ground truth alone. Where the target varies little beyond the remaining alignment, reporting the template (recommendation 4 of the main paper) is the safeguard. PA-ATE applies the same rule to the camera's orientation. What the pose case adds is the template side, Propositions~6 and~7: an input-free template whose orbit closes to an affine subspace scores the ground truth's distance to that subspace, which flags a benchmark before any model runs.

\section{Implementation notes}
\label{sec:s_impl}
\paragraph{Models.} All eleven models were run with their official inference code and released weights. VGGT, $\pi^3$, $\pi^3$X, DA3-Large, DA3-Giant, CUT3R, TTT3R and StreamVGGT were fixed before any result on the test datasets was computed, and DA3-Small, DA3-Base and MapAnything were added afterwards.
\paragraph{Straightened DA3-Giant.} Straightened fully, DA3-Giant ranks third of twelve under PA-ATE, second under RPE$_t$ over 10 frames and fourth under DTE. When each of the 11 models has its positions replaced by their constant-velocity fit in the same way as DA3-Giant in \cref{sec:notchange}, its median rank among the twelve is seventh under PA-ATE, seventh under RPE$_t$ over 10 frames and fifth under DTE.

\paragraph{Implementing the line.} evo rejects an exactly straight prediction as degenerate, and the official evaluators then drop the clip from the mean. We therefore add isotropic noise of amplitude $10^{-6}$. The score then equals Eq.~(\mref{eq:line}) to within $5\times10^{-8}$ on every clip.

\paragraph{Official evaluator.} Our evaluation code reproduces the per-sequence ATE written by the official CUT3R/TTT3R evaluator to within $5\times10^{-7}$ on all 44 model--sequence pairs where both exist (results/official\_eval\_check.csv).
\paragraph{Reproducibility.} We re-ran all eleven models from scratch on Sintel and TUM with the current releases, recording every code commit and weight revision (results/repro\_*/versions\_*.json), on a GPU type different from the original runs (offline models on an A100, streaming models on an RTX PRO 6000). The per-clip ATE changed by a median of 0.2 to 2.6\% per model, by at most 12\% on single clips for the non-DA3 models, and by up to 68\% on single Sintel clips for the four Depth Anything 3 models (DA3-Giant: mean Sintel ATE 0.130 against 0.112). Re-running the DA3 models on the original GPU type reproduces their predictions to within 0.2\% of every clip's ATE, identically in two passes. Their output depends on the GPU type, not on chance. With either set of predictions the line is second on Sintel, beating 10 of the 11 models, and last on TUM (tools/repro\_compare.py, tools/repro\_da3\_compare.py).

\paragraph{Data and ethics.} All datasets are public research benchmarks used under their published terms, and ScanNet was obtained from its official release under its terms of use. We collected no new data and ran no study with human participants. Several real sequences (TUM, Bonn, Wild-SLAM, ADT and ScanNet) show people or private indoor spaces. The paper shows images only from Sintel, which is rendered, and the released package contains predicted camera poses, scores and code but no dataset images, and it withholds the ground truth.

\paragraph{SpatialTrackerV2.} We ran its full pipeline (front end, tracking and bundle adjustment) with the official inference settings in the environment its README specifies (Python 3.11, torch 2.4.1, pycolmap 3.11.1, pyceres 2.4; notebooks/sota\_models.ipynb), without modification. It scores a mean Sintel ATE of 0.115 (median 0.036, 9 wins of 14) and 0.018 on TUM. The front end alone, re-run in the same environment, gives 0.118 on Sintel and 0.012 on TUM. Its paper reports 0.054 on Sintel. Two re-evaluations under this protocol report 0.126~\cite{d4rt} and 0.133~\cite{track4world}.
\paragraph{LASER and AnyCam.} Both were run with their official code and weights, each in its own Python 3.11 environment (notebooks/sota\_models.ipynb). LASER's repository no longer has the branch that its README names for the paper's evaluation, so we ran its default branch (commit 9119e33) with the two configurations it ships: with loop closure (windows of 10 frames) it scores 0.085 on Sintel and 0.021 on TUM, and without (windows of 20) 0.064 and 0.017, close to the published 0.061. \Cref{tab:sintel} shows the better of the two, which is conservative for the line.
For AnyCam we installed torch 2.5.1 for CUDA 12.4, as its README specifies, and every other package as released before the code's last commit (7 May 2025, commit e609cc8), and ran the released seq8 model with the repository's evaluation configuration (anycam/configs/eval\_cfgs/fit\_video.yaml: global bundle adjustment at the end), and, as a check, through its PyTorch Hub entry point with default settings (global bundle adjustment every 2 windows) and without bundle adjustment. The repository's evaluation script reports the mean of the per-frame position errors after Sim(3) alignment, not their RMS (anycam/scripts/evaluate\_trajectories.py, line 252). Under that statistic our predictions score 0.078 with the evaluation configuration, the published value, 0.077 with the default settings and 0.104 without bundle adjustment (published 0.099). Under the protocol's RMS they score 0.093, 0.092 and 0.120 (tools/sota\_mean\_ape.py). AnyCam's training data, as listed in its paper~\cite{anycam}, do not include Sintel.
Of the five configurations that enter \cref{tab:survey} here (the PyTorch Hub run with default settings is only a check), half straightening lowers the Sintel ATE of four, each with a median residual SNR below $\sqrt3$ (0.84 to 1.68): AnyCam from 0.093 to 0.089 and from 0.120 to 0.094 without bundle adjustment, LASER with loop closure from 0.085 to 0.071, and SpatialTrackerV2 from 0.115 to 0.087. It raises the ATE of LASER without loop closure, whose median SNR is 1.83, from 0.064 to 0.066.
\paragraph{UniQuery4R.} Its repository releases inference code and weights but no evaluation script (commit 9511974). We ran the released model with its own image loader and default settings on the 14 Sintel and 8 TUM clips and scored its predicted extrinsics under the protocol (notebooks/sota\_models.ipynb). It scores a mean Sintel ATE of 0.033, against the published 0.031, and 0.014 on TUM. It beats the line on 11 of the 14 clips. Half straightening raises its Sintel ATE to 0.054 and lowers it on only 3 clips. Its median residual SNR is 4.25. With the measured $c$, \cref{eq:threshold} gives the outcome on all 22 of its Sintel and TUM clips.
\paragraph{MonST3R.} We ran the official repository (commit 574cc77) and released checkpoint with its pose-evaluation launcher at default settings (sliding-window scene graph of 5 frames, 300 iterations of global alignment, optical flow and SAM2~\cite{sam2} motion masks; notebooks/monst3r\_sintel\_tum.ipynb, notebooks/monst3r\_scannet\_a.ipynb and \_b.ipynb) on an RTX PRO 6000, and scored the trajectories it saves with our code. Its own per-clip evaluator fails with current numpy releases, so we log that error and continue, leaving inference unchanged. MonST3R scores a mean Sintel ATE of 0.118 (median 0.084; published 0.108~\cite{monst3r}, 0.111 in CUT3R's re-run~\cite{cut3r}) and 0.106 on TUM, within the published 0.063 to 0.108, which differ between its arXiv versions and later re-runs (results/lit\_pose\_table.csv). Its PA-ATE is 0.232 on Sintel and 0.141 on TUM. Half straightening lowers its Sintel ATE to 0.093 and lowers it on 11 of the 14 clips. Its median residual SNR is 0.90. With the measured $c$, \cref{eq:threshold} gives the outcome on all 22 of its Sintel and TUM clips. On ScanNet, where its median residual SNR is 4.6, half straightening raises its mean ATE from 0.057 to 0.113 and lowers it on 7 of the 94 clips. The criterion gives the outcome on 90 of them and certifies no loss.
\paragraph{DTE.} Our implementation follows Eqs.~10--19 of~\cite{lee2022ate}: both trajectories are centred on their geometric medians and scaled by their median distance from it, the rotation is the rotation median of $\mathbf{R}_i\hat{\mathbf{R}}_i^\top$, each position error is capped at $k=5$ times the true median distance and divided by that cap, and the score averages their mean and RMS, with $k$ as in~\cite{lee2022ate} (results/second\_implementation/run\_analysis.py).
\enlargethispage{\baselineskip}\paragraph{Bonn.} Its ground truth describes the motion-capture markers, which we convert to camera poses with the published calibration (trajaudit.core.bonn\_camera\_poses).
\paragraph{Predicted gains.} Tab.~\mref{tab:deletion} predicts a gain when half straightening lowers the mean fixed-alignment ATE from $S$, $N$ and $c$. The Sintel miss gains after refitting. Bootstrap 95\% intervals: 0--0 (TUM, ScanNet), 0--1 (Bonn, Wild-SLAM).

\end{document}